\documentclass{article}

\PassOptionsToPackage{numbers,sort&compress}{natbib}
\usepackage[preprint]{neurips_2026}

\usepackage[utf8]{inputenc} 
\usepackage[T1]{fontenc}    
\usepackage{hyperref}       
\usepackage{url}            
\usepackage{booktabs}       
\usepackage{amsfonts}       
\usepackage{nicefrac}       
\usepackage{microtype}      
\usepackage[normalem]{ulem}

\usepackage{multirow}
\usepackage{amsmath}
\usepackage{graphicx}
\usepackage{caption}
\usepackage{algorithm}
\usepackage{subcaption}
\usepackage{algpseudocode}
\usepackage[table]{xcolor}

\usepackage{wrapfig}

\usepackage{pifont}
\newcommand{\cmark}{\textcolor{green!85!black}{\ding{51}}}
\newcommand{\xmark}{\textcolor{red!85!black}{\ding{55}}}

\title{Beyond Rigid: Benchmarking Non-Rigid Video Editing}

\author{
Bingzheng Qu\textsuperscript{1} \quad
Xuefeng Bai\textsuperscript{1} \quad
Kehai Chen\textsuperscript{1,*} \quad
Min Zhang\textsuperscript{1}\\
\textsuperscript{1}Harbin Institute of Technology, Shenzhen, China\\
\textsuperscript{*}Corresponding author\\
\texttt{qbzz@stu.hit.edu.cn, \{baixuefeng, chenkehai, zhangmin2021\}@hit.edu.cn}
}

\begin{document}

\maketitle

\begin{abstract}
As video generation models are increasingly expected to manipulate physical dynamics, 
there is a growing need to move evaluation beyond appearance fidelity and semantic alignment. 
Non‑rigid video editing offers a uniquely revealing testbed, where distinct materials impose distinct physical constraints. 
In this paper, we introduce NRVBench, a diagnostic benchmark for non-rigid video editing, where the task is to modify deformable motion while preserving irrelevant regions and maintaining material-specific plausibility. 
NRVBench contains 180 curated videos across six physics-grounded categories, 2,340 fine-grained editing instructions, 360 multiple-choice questions, and pixel-accurate masks. We further propose NRVE-Acc, a structured VLM-based protocol that decomposes editing success into instruction following, material-aware deformation plausibility, and temporal coherence with motion cues. Experiments on representative inference-time video editing methods reveal a clear mismatch between conventional metrics and physics-aware perceptual editing success: methods that preserve appearance or achieve strong global alignment may still fail under non-rigid dynamics. 
We additionally introduce VM‑Edit, a simple region‑conditioned editing baseline that frees the foreground while locking the background, exposing the stability–plasticity trade‑off. 
\end{abstract}

\vspace{-5pt}
\section{Introduction}
\label{Introduction}
\vspace{-3pt}
Recent advances in text-to-video generation~\citep{yang2025cogvideox,deng2025autoregressive,Dalal2025One} and video editing~\citep{Ouyang2024CoDeF,Wang2025VideoDirector,Zhu2025FADE} have made it feasible to manipulate not only visual appearance but also the underlying physical dynamics of a scene~\citep{Xue2025PhyT2V}.
This shift creates an urgent need for benchmarks that assess physical plausibility, beyond semantic alignment, background preservation, and frame-level temporal smoothness—metrics that are inherently blind to whether an edited motion respects material-specific physics.
Non‑rigid video editing offers a revealing and well-scoped test scenario for precisely this requirement. 
To succeed, a method is expected to generate deformable motion that follows an instruction while preserving unedited regions and maintaining material-specific physical plausibility~\citep{meng2025worldsimulator,Yang2025VLIPP}. 
Editing a cloth fold, a waving paw, a flame, or a liquid surface requires different physical constraints, making non-rigid editing a natural stress test for physics-aware video generation.


However, existing video editing benchmarks and metrics offer limited diagnostic power for this setting~\citep{sun2025vebench,chen2025editboard,Li2025fivebench}. General video editing benchmarks primarily evaluate appearance-level or task-level transformations, while common metrics such as CLIP similarity~\citep{pmlr2021learning}, LPIPS~\citep{zhang2018lpips}, PSNR~\citep{huynh2008psnr}, and generic motion scores do not directly verify whether the edited deformation is physically plausible. 
Consequently, methods can achieve high scores even when they produce failures typical of non‑rigid editing, including rubber‑like limbs, ghosting, object popping, texture melting, and topology violations. 
This mismatch makes it difficult to compare methods fairly or to reliably measure progress: an edit may preserve the background and align with the text yet still contain physically implausible, temporally incoherent motion. 

\begin{figure}[!t]
  \centering
  \includegraphics[width=\linewidth]{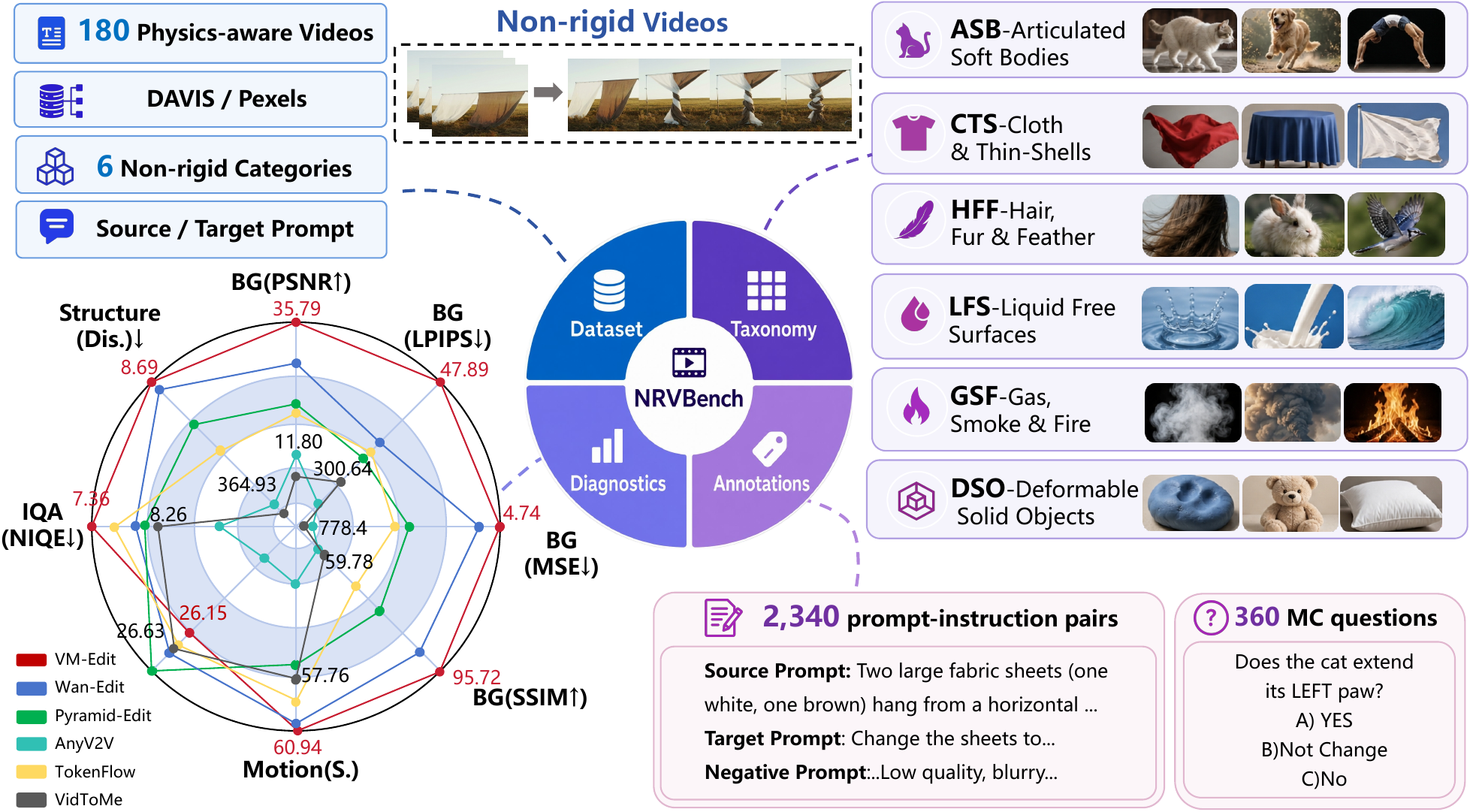}
  \caption{Overview of NRVBench. The benchmark covers six physics-grounded categories and supports fine-grained diagnosis of non-rigid video editing methods. }
  \label{fig:nrvbench-overview}
\end{figure}

To close this gap, we introduce NRVBench (Figure~\ref{fig:nrvbench-overview}), a dedicated diagnostic evaluation suite for physics-aware non-rigid video editing. 
NRVBench comprises 180 curated videos across six physics-grounded categories, covering articulated, thin-shell, fibrous, fluid, gaseous/fire, and deformable-solid dynamics. Each instance is equipped with fine-grained edit instructions, diagnostic multiple-choice questions, and pixel-accurate masks, yielding 2,340 instructions and 360 structured questions in total. 
In addition to the standard 60-frame benchmark, NRVBench includes a dedicated 150-frame long-video subset for analyzing non-rigid editing performance under extended temporal evolution and physical cycles.

Building on NRVBench, we propose NRVE-Acc, a structured VLM-based protocol for evaluating non-rigid video editing. NRVE-Acc decomposes editing success into instruction following, material-specific deformation plausibility, and temporal coherence, using category-conditioned diagnostic questions and motion cues rather than a single global similarity score. 
To assess judge‑model dependency, we evaluate NRVE-Acc across multiple VLM judges and compare it with human annotations. We then benchmark representative video editing methods under a controlled setting with the same source videos, masks, and instructions. The results reveal a pronounced gap between conventional metrics and physics-aware perceptual success: methods may preserve appearance or semantic alignment while still failing under non-rigid dynamics. 
To complement the evaluation suite, we introduce VM-Edit, a region-conditioned video editing method. By freeing the foreground while locking the background, it reduces background flicker and temporal artifacts while probing the stability--plasticity trade-off.
Its promising performance suggests that region-conditioned editing is a promising route for improving controllability and physical plausibility in non-rigid video editing.

\vspace{-3pt}
\section{Related Work}
\label{Related Work}
\vspace{-3pt}

\subsection{Video editing benchmarks}
The rapid development of text-guided video editing has motivated several benchmarks for standardized evaluation~\citep{liu2025editeval}. TGVE~\citep{wu2023TGVE} and TGVE+~\citep{singer2024TGVE+} provide early protocols for controllable video editing, while V2VBench~\citep{sun2024v2vbench}, VEditBench~\citep{wu2024veditbench}, and FiVE~\citep{Li2025fivebench} further expand the coverage of editing tasks and fine-grained categories. However, these benchmarks mainly evaluate edits at the task, object, or semantic level, and do not explicitly assess whether edited videos preserve material-specific deformation realism. As a result, they provide limited diagnostic evidence for non-rigid editing, where success requires not only instruction alignment but also physically plausible deformation and temporal coherence. NRVBench addresses this gap by organizing videos into physics-grounded non-rigid categories and pairing them with fine-grained instructions and diagnostic questions.

\subsection{Video editing metrics and methods}
Existing video editing evaluation commonly relies on text-alignment metrics such as CLIP similarity~\citep{pmlr2021learning}, preservation metrics such as LPIPS~\citep{zhang2018lpips}, PSNR~\citep{huynh2008psnr}, MSE~\citep{wang2009mse}, and SSIM~\citep{wang2004ssim}, no-reference quality metrics such as NIQE, and motion-based temporal scores. These metrics are useful for measuring visual fidelity, background preservation, and global semantic consistency, but cannot verify physically plausible non-rigid deformation. Pixel-level metrics may penalize valid deformation, while semantic metrics may overlook material-inconsistent motion or temporal artifacts. Recent VLM-based evaluators offer more flexible judgment, but a single global score often lacks diagnostic interpretability~\citep{meng2025worldsimulator,Li2025fivebench}. We therefore propose NRVE-Acc, a structured VLM-based protocol that decomposes evaluation into instruction following, deformation plausibility, and temporal coherence.

Recent video editing methods have explored diverse mechanisms for improving controllability, temporal consistency, and edit fidelity~\citep{khachatryan2023text2videozero,QI2023fatezero,yoon2024dni,chang2025bytemorph}. 
Representative approaches include TokenFlow~\citep{geyer2024tokenflow}, AnyV2V~\citep{ku2024anyv2v}, VidToMe~\citep{li2024vidtome}, Wan-Edit~\citep{Li2025fivebench}, and Pyramid-Edit~\citep{Li2025fivebench}, which perform video editing through inference-time propagation, attention manipulation, or region-aware generation, as well as Tune-A-Video~\citep{wu2023tuneavideo}, MotionEditor~\citep{tu2024motioneditor,tu2025motionfollower}, and ReVideo~\citep{mou2024revideo}, which adapt models or motion representations for improved controllability. 
Despite this progress, non-rigid editing remains challenging under discontinuous, topology-changing, or material-specific dynamics. 
We therefore evaluate representative inference-time methods to enable controlled comparison under identical inputs and avoid unfairness introduced by method-specific training data, fine-tuning budgets, or benchmark adaptation, while NRVBench itself is method-agnostic.

\vspace{-3pt}
\section{NRVBench}
\label{NRVBench}

\subsection{Overview}
 NRVBench is designed to evaluate physics-aware non-rigid video editing. Given a source video involving physical dynamics, an editing instruction, and an editing region, the benchmark asks whether a model can generate the requested deformable motion while preserving unedited regions and maintaining material-specific physical plausibility. Unlike generic video editing benchmarks that mainly assess semantic alignment or appearance preservation, NRVBench targets deformation adherence: an edit is considered successful only when it follows the instruction, respects the underlying material regime, and remains temporally coherent. 

\begin{wraptable}{r}{0.7\textwidth}
\vspace{-8pt}
\centering
\scriptsize
\setlength{\tabcolsep}{2.6pt}
\renewcommand{\arraystretch}{1.1}
\caption{Comparison of controllable video editing benchmarks.}
\label{tab:bench_compare}
\resizebox{\linewidth}{!}{
\begin{tabular}{lcccccc}
\toprule
\textbf{Benchmark} 
& \textbf{\#Videos} 
& \textbf{\#Prompts} 
& \textbf{T-step} 
& \textbf{Mask} 
& \textbf{Physical-aware} 
& \textbf{Taxonomy} \\
\midrule
BalanceCC~\citep{feng2024balancecc} 
& 100 & 400 & \xmark & \xmark & \xmark & Task-level \\

V2VBench~\citep{sun2024v2vbench} 
& 50 & 150 & \xmark & \xmark & \xmark & Task-level \\

TGVE~\citep{wu2023TGVE} 
& 76 & 304 & \xmark & \xmark & \xmark & Task-level \\

TGVE+~\citep{singer2024TGVE+} 
& 76 & 1417 & \xmark & \xmark & \xmark & Task-level \\

FiVE~\citep{Li2025fivebench} 
& 100 & 420 & \xmark & \cmark & \xmark & Object-level \\

\rowcolor{gray!20}
\textbf{NRVBench} 
& \textbf{180} 
& \textbf{2340} 
& \textbf{\cmark} 
& \textbf{\cmark} 
& \textbf{\cmark} 
& \textbf{Physics-level} \\
\bottomrule
\end{tabular}
}
\vspace{-5pt}
\end{wraptable}

\paragraph{Benchmark Composition.} NRVBench consists of two complementary evaluation settings. The main benchmark contains 180 curated videos across six physics-grounded non-rigid categories: articulated soft bodies \textbf{(ASB)}, cloth and thin shells \textbf{(CTS)}, deformable solid objects \textbf{(DSO)}, gas/smoke/fire \textbf{(GSF)}, hair/fur/feather \textbf{(HFF)}, and liquid free surfaces \textbf{(LFS)}. Each video is paired with fine-grained editing instructions, source and target prompts, diagnostic multiple-choice questions, negative prompts, and pixel-accurate editing masks. In total, NRVBench provides 2,340 prompt-instruction pairs and 360 diagnostic questions, supporting both conventional metric evaluation and structured VLM-based assessment. As shown in Table~\ref{tab:bench_compare}, NRVBench is larger than most mainstream controllable video editing benchmarks in terms of source videos and prompt annotations, while also providing temporal-step annotations, masks, and physics-aware labels. NRVBench is intended as a controlled diagnostic benchmark rather than an exhaustive open-world dataset. This design prioritizes evaluation density and reliability while remaining complementary to larger-scale benchmarks. Appendix~\ref{appendix:datasets} provides details.

In addition to the standard 60-frame benchmark, we design three groups of 150-frame long-video analyses to further verify non-rigid editing performance under extended temporal evolution. The long-video setting covers three aspects: L1 motion, L2 fluid dynamics, and L3 elasticity. By observing complete deformation processes over longer durations, this setting provides additional evidence for whether models can maintain instruction-consistent motion, physically plausible responses, and material consistency throughout a full physical cycle.

\subsection{Benchmark Construction}
\begin{figure}[!t]
  \centering
  \includegraphics[width=\linewidth]{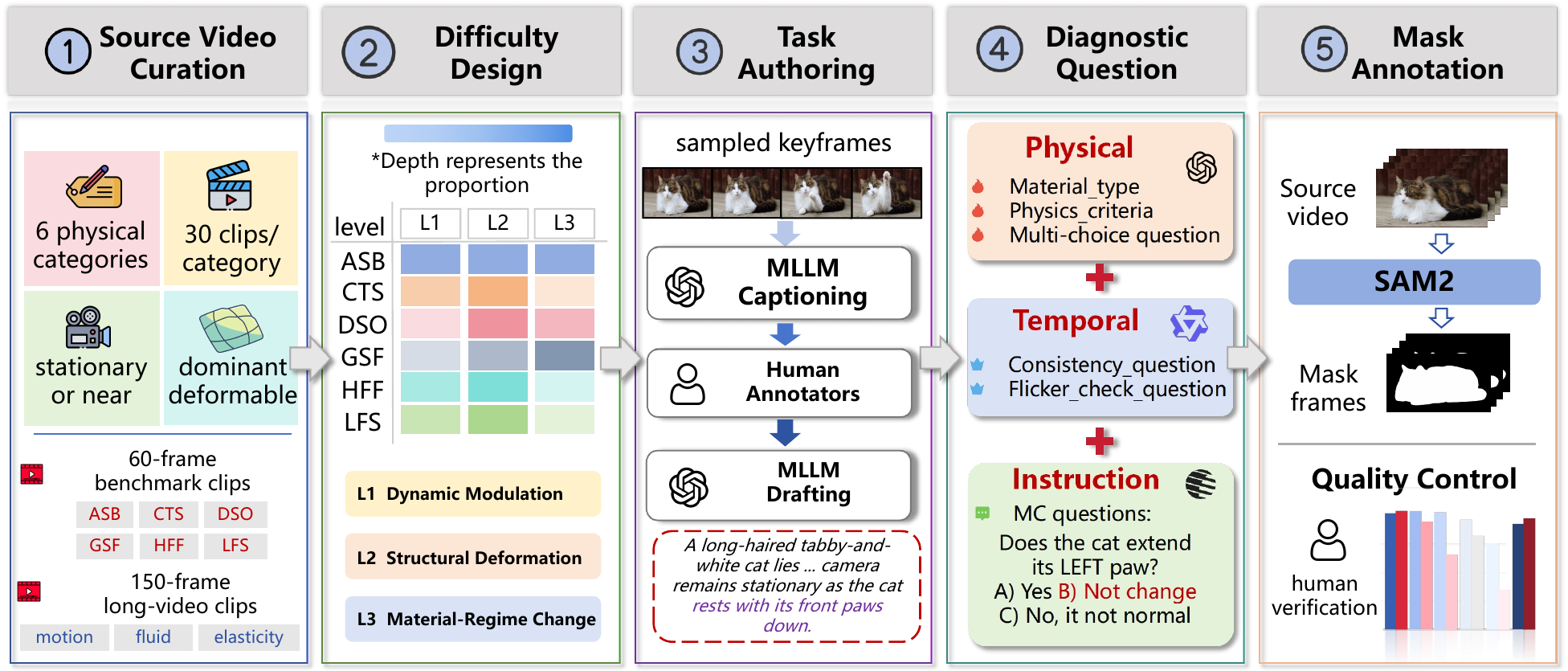}
  \caption{Five-stage pipeline for benchmark construction.}
  \label{fig:benchmark_pipeline}
\end{figure}

Inspired by PhyGenBench~\citep{meng2025worldsimulator}, we construct NRVBench through a five-stage pipeline designed to ensure physical groundedness, identity preservation, and evaluation falsifiability as shown in Figure~\ref{fig:benchmark_pipeline}.

\textbf{1) Source video curation.}
For each physical category, we curate 30 source clips with stationary or near-stationary cameras, a dominant deformable subject, and clearly observable non-rigid motion. Standard benchmark videos are resampled to 60 frames for dense annotation and compatibility with current editing models, while 150-frame clips are retained for long-video analysis.

\textbf{2) Difficulty design.}
To support fine-grained diagnosis, we organize editing tasks into three levels: L1 dynamics modulation, L2 coupled or structural deformation, and L3 topology or material-regime change. This produces a category-by-difficulty evaluation grid for analyzing model behavior.

\textbf{3) Editing task authoring.}
For each video, we construct an editing prompt tuple through a human--MLLM collaborative process. An MLLM first generates literal source captions from sampled keyframes. Human annotators then specify the desired non-rigid edit and difficulty level, after which the MLLM proposes candidate target prompts and instructions. Annotators finally revise them to ensure that the edit preserves identity, scene layout, and viewpoint, while requiring genuine deformation rather than appearance-only modification.

\textbf{4) Diagnostic question construction.}
We use GPT-4o to construct multiple-choice diagnostic questions for each instance to evaluate instruction following, material-specific deformation plausibility, and temporal coherence. These questions provide observable evidence for whether an edit succeeds, and are used by NRVE-Acc for structured VLM-based evaluation.

\textbf{5) Mask annotation and quality control.}
We generate per-frame editing-region masks with SAM2~\citep{ravi2025sam} and manually verify them frame by frame. The masks support local metric routing between edited and unedited regions and provide consistent conditioning for mask-based baselines. 

We provide details of the data construction, prompt generation templates, filtering rules, manual verification process, and difficulty design in Appendix~\ref{appendix:datasets}.
\vspace{-3pt}
\section{NRVE-Acc}
\label{NRVE-Acc}
\begin{figure}[!t]
  \centering
  \includegraphics[width=\linewidth]{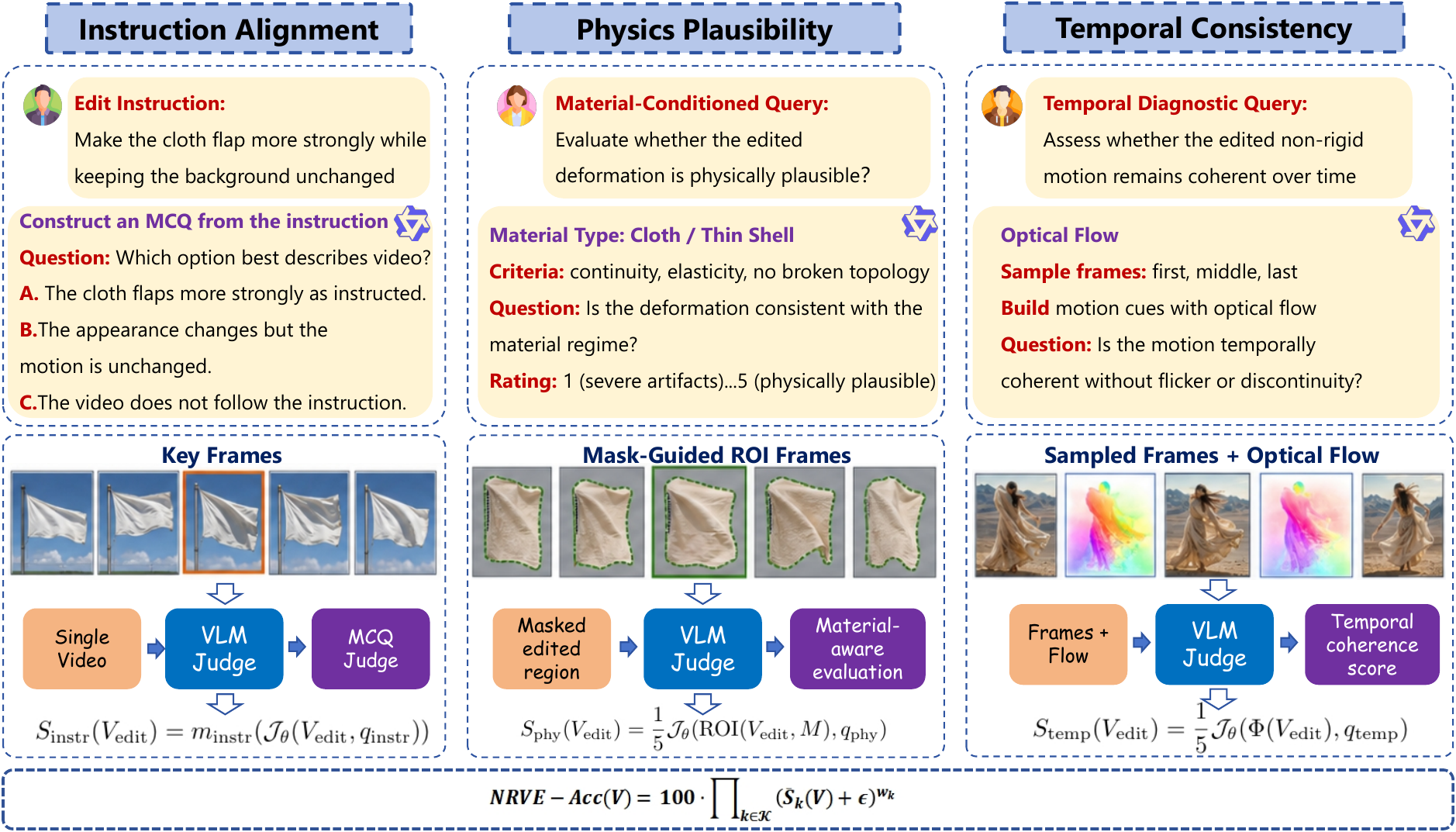}
  \caption{ An overview of the proposed NRVE-Acc. NRVE-Acc evaluates non-rigid video editing from three complementary dimensions: instruction alignment, physics plausibility, and temporal consistency. The sub-scores are aggregated with a geometric mean to obtain the NRVE-Acc score. 
}
  \label{fig:NRVE-Acc_overview}
\end{figure}

\vspace{-2pt}
NRVE-Acc is a structured VLM-based evaluation protocol for non-rigid video editing. Given an edited video $V$, NRVE-Acc queries a visual-language judge along three complementary dimensions: \textit{instruction alignment}, \textit{physics plausibility}, and \textit{temporal consistency}. These dimensions correspond to whether the edit follows the target instruction, whether the generated deformation is physically plausible under the corresponding material regime, and whether the motion remains coherent over time. Unlike a single holistic VLM score, NRVE-Acc decomposes the judgment into interpretable sub-scores and aggregates them with a geometric mean, so that failure in any one dimension lowers the final adherence score as shown in Figure~\ref{fig:NRVE-Acc_overview}. 

Since VLM-based evaluation can be sensitive to the choice of judge model, we further validate NRVE-Acc through extensive experiments with both open-source and closed-source VLMs, together with human-annotation agreement analysis. This design makes NRVE-Acc a judge-aware evaluation protocol rather than a metric tied to a single proprietary model.

Let $\mathcal{K}=\{\mathrm{instr}, \mathrm{phy}, \mathrm{temp}\}$ denote the three evaluation dimensions. For an edited video $V$, we obtain normalized scores $\bar{S}_k(V)\in[0,1]$ for each $k\in\mathcal{K}$. The final score is computed as
\begin{equation}
\mathrm{NRVE\text{-}Acc}(V)
=
100 \cdot
\prod_{k\in\mathcal{K}}
\left(\bar{S}_k(V)+\epsilon\right)^{w_k},
\label{eq:nrve_acc}
\end{equation}
where $w_k$ denotes the weight of each dimension and $\sum_k w_k=1$. We use uniform weights $w_k=1/3$ by default and set $\epsilon=10^{-6}$ for numerical stability. The geometric mean encourages balanced editing success: a high NRVE-Acc requires the video to be instruction-consistent, physically plausible, and temporally coherent.
\vspace{-2pt}
\subsection{Instruction Alignment Evaluation}
Instruction alignment measures whether the edited video satisfies the requested non-rigid change. For each instance, we derive a diagnostic multiple-choice question $q_{\mathrm{instr}}$ from the target instruction and query the VLM judge $\mathcal{J}_{\theta}$ on the edited video. The predicted answer is then mapped to a soft correctness score:
\begin{equation}
S_{\mathrm{instr}}(V_{\mathrm{edit}})=m_{\mathrm{instr}}\!\left(\mathcal{J}_{\theta}(V_{\mathrm{edit}},q_{\mathrm{instr}})\right), \quad m_{\mathrm{instr}}(c)\in\{1.0,0.5,0.0\}.
\end{equation}
Here, scores of $1.0$, $0.5$, and $0.0$ indicate correct, partial or ambiguous, and failed instruction satisfaction, respectively. This formulation turns instruction following into a falsifiable visual question rather than relying on generic text-video similarity.

\vspace{-2pt}
\subsection{Physics Plausibility Evaluation}
Physics plausibility measures whether the generated deformation is consistent with the target material regime. NRVE-Acc is not intended to replace simulation-based physical verification. Instead, it provides a material-aware perceptual diagnostic protocol that evaluates whether the edited deformation exhibits visually observable physical plausibility under the target material regime. For each instance, we construct a material-conditioned query $q_{\mathrm{phy}}=(q,\mu,\Gamma_{\mu})$, where $\mu$ denotes the physical category and $\Gamma_{\mu}$ specifies its evaluation criteria. The VLM judge scores the mask-guided edited region and normalizes the result as:
\begin{equation}
S_{\mathrm{phy}}(V_{\mathrm{edit}})=\frac{1}{5}\mathcal{J}_{\theta}\!\left(\mathrm{ROI}(V_{\mathrm{edit}},M),q_{\mathrm{phy}}\right), \quad \mathcal{J}_{\theta}(\cdot)\in\{1,2,3,4,5\}.
\end{equation}
Low scores indicate severe artifacts such as broken topology, unnatural stretching, or material-inconsistent motion, while high scores indicate natural and physically plausible deformation. By conditioning on the material category, this metric focuses on material-aware deformation cues rather than generic visual realism.

\vspace{-3pt}
\subsection{Temporal Consistency Evaluation}
\vspace{-1pt}
Temporal consistency measures whether the edited non-rigid motion remains coherent over time. 
Given an edited video $V_{\mathrm{edit}}=\{I_t\}_{t=0}^{T-1}$ with $T$ frames, where $I_t$ is the $t$-th frame, we sample the first, middle, and last frames, and construct a flow-based motion-evidence image from adjacent sampled frames, where $\Phi(V_{\mathrm{edit}})$ is the motion evidence. 
Here, $\mathrm{Flow}(I_a,I_b)$ denotes optical flow from $I_a$ to $I_b$, $\mathrm{Vis}(\cdot)$ denotes color-coded flow visualization:
\begin{equation}
\Phi(V_{\mathrm{edit}})=\mathrm{Concat}\!\left(\mathrm{Vis}(\mathrm{Flow}(I_0,I_m)),\mathrm{Vis}(\mathrm{Flow}(I_m,I_{T-1}))\right), \quad m=\left\lfloor \frac{T}{2} \right\rfloor,
\end{equation}
\vspace{-3pt}
\begin{equation}
S_{\mathrm{temp}}(V_{\mathrm{edit}})=\frac{1}{5}\mathcal{J}_{\theta}\!\left(\Phi(V_{\mathrm{edit}}),q_{\mathrm{temp}}\right), \quad \mathcal{J}_{\theta}(\cdot)\in\{1,2,3,4,5\}.
\end{equation}
By evaluating optical-flow evidence rather than individual frames, $S_{\mathrm{temp}}$ is sensitive to flicker, teleportation, abrupt geometry changes, and discontinuous deformation.

\begin{figure}[!t]
  \centering
  \includegraphics[width=0.9\linewidth]{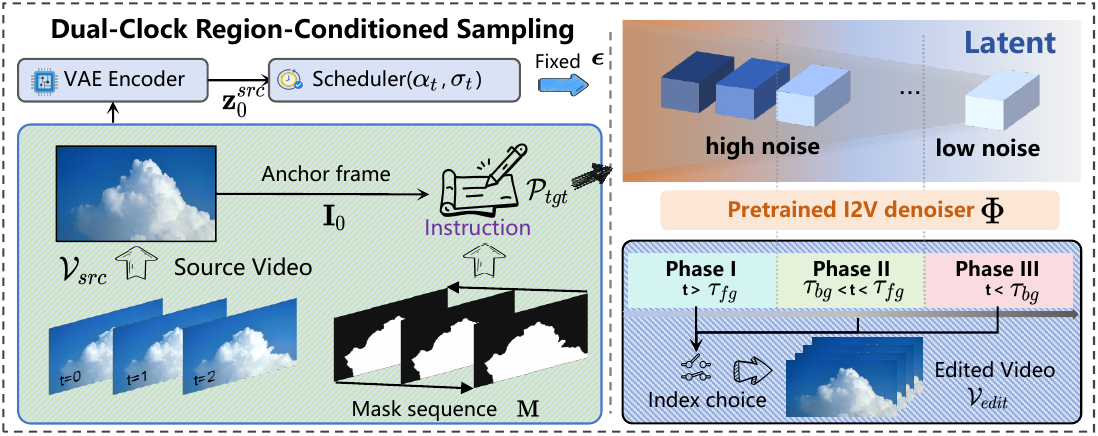}
  \caption{Overview of VM-Edit, a dual-clock region-conditioned sampling method.
}
  \label{fig:VM-Edit}
  \vspace{-3pt}
\end{figure}
\vspace{-3pt}

\subsection{VM-Edit}
\vspace{-2pt}
Inspired by region-conditioned diffusion sampling for controllable synthesis and editing, including TTM~\citep{singer2025ttm}, Blended Diffusion~\citep{avrahami2022blended}, and RePaint~\citep{lugmayr2022repaint}, we identify region-wise sampling as a promising way to resolve the stability--plasticity dilemma in non-rigid video editing. Specifically, we propose VM-Edit (Figure~\ref{fig:VM-Edit}), a structure-aware editing method that replaces manual motion guidance with a simple region-conditioned pipeline. Designed to resolve the stability-plasticity dilemma, VM-Edit employs a region-conditioned sampling pipeline that allocates freedom to the foreground while locking the background, reducing background flicker and temporal artifacts.

\paragraph{Dual-Clock Region-Conditioned Sampling.}
Let $\mathbf{z}_t$ denote the current latent at reverse diffusion step $t$ (larger $t$ indicates higher noise). Let $\Phi(\cdot)$ be one reverse-step update of the pre-trained denoiser/sampler, conditioned on $\mathcal{P}_{tgt}$ and the anchor frame $\mathbf{I}_0$:
\begin{equation}
\hat{\mathbf{z}}_{t-1} = \Phi(\mathbf{z}_t, t, \mathcal{P}_{tgt}, \mathbf{I}_0).
\label{eq:denoise_update}
\end{equation}
We enforce region-wise control by recomposing the next latent with a \emph{foreground edit} term and a \emph{background stabilization} term:
\begin{equation}
\mathbf{z}_{t-1} \leftarrow \mathbf{m}\odot \mathbf{z}^{fg}_{t-1} + (\mathbf{1}-\mathbf{m})\odot \mathbf{z}^{bg}_{t-1},
\label{eq:recompose}
\end{equation}
where $\odot$ is element-wise multiplication and $\mathbf{m}$ is downsampled to the latent resolution. To explicitly allocate plasticity between regions, we use two timestep indices $\tau_{fg}$ and $\tau_{bg}$ (with $\tau_{fg}>\tau_{bg}$). Following an SDEdit-style initialization at $t=\tau_{fg}$ , we enforce a plasticity gap during the interval $\tau_{fg} \ge t > \tau_{bg}$. In this phase, the foreground evolves freely via the diffusion model to accommodate non-rigid deformations, while the background is explicitly anchored to the source latent trajectory to suppress temporal flicker. Once $t \le \tau_{bg}$, both regions are refined jointly to ensure boundary coherence. To address the problem of manually selecting fixed foreground and background timesteps, we propose an adaptive dual-clock scheduler that estimates the required editing plasticity from each input instance. (Details are provided in Appendix~\ref{appendix:VM-Edit}).

\section{Experiment}
\label{Experiment}
\vspace{-2pt}
\paragraph{Experimental Setup.} 
 We compare representative training-free or inference-time video editing baselines, including TokenFlow~\cite{geyer2024tokenflow}, AnyV2V~\cite{ku2024anyv2v}, VidToMe~\cite{li2024vidtome}, Pyramid-Edit~\cite{Li2025fivebench}, and Wan-Edit~\cite{Li2025fivebench}, together with our VM-Edit method. For benchmark annotation, we use GPT-4o as a prompt-generation assistant to propose prompts, which are then manually revised and verified by human annotators. For NRVE-Acc, we use multiple VLM judges, including Qwen2.5-VL-7B-Instruct, Qwen2.5-VL-32B-Instruct, Qwen2.5-VL-72B-Instruct, Gemini-2.5 Pro, and Kimi-2.5, and report both component scores and cross-judge statistics. In addition to NRVE-Acc, we report common video editing metrics on NRVBench, including structure distance, PSNR, LPIPS, MSE, and SSIM, CLIP-S, NIQE, motion score, and FPS. We provide details in Appendix~\ref{appendix:experiments_setup}.

\paragraph{Human Evaluation.}
To verify the reliability of our metric, we conducted a stratified random sampling of ten source videos from each non-rigid target category, accompanied by five corresponding editing prompts and four MCQs for each video. Five human annotators answered the same diagnostic questions as the VLM judges under a blind, randomized setting, and their scores were mapped to the NRVE-Acc scoring space and averaged for comparison. The detailed quantitative agreement shows that NRVE-Acc maintains substantial agreement with human rankings.

\begin{figure}[htbp]
  \centering
  \includegraphics[
    width=\linewidth,
  ]{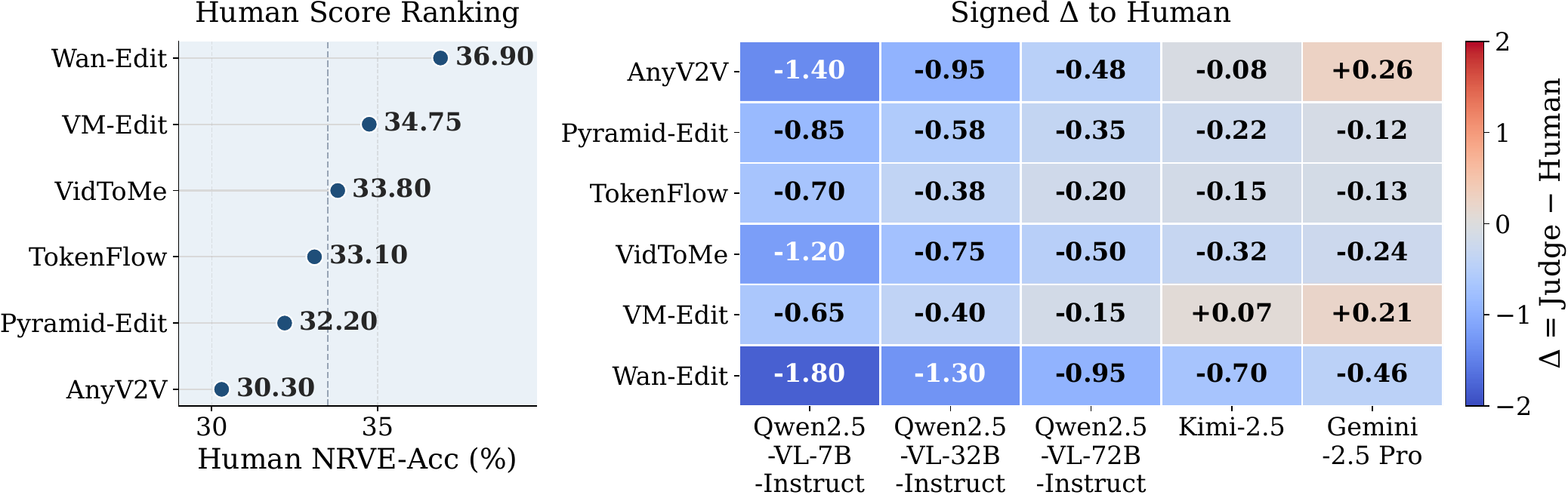}
  \caption{Human scores and signed deviations of VLM Judges. Columns are ordered by average judge performance from worse to better. Signed $\Delta>0$ indicates overestimation relative to Human.}
  \label{fig:human_evaluation}
\end{figure}

Different VLM judges exhibit different score calibration behaviors. 
Some closed-source judges assign consistently lower absolute NRVE-Acc scores, but their rankings remain consistent with human annotations as shown in Figure~\ref{fig:human_evaluation}. 
We therefore interpret judge-specific scores mainly through within-judge comparisons and report cross-judge statistics to quantify robustness. Additional details on human evaluation in Appendix~\ref{appendix:human_Evaluation}, mask quality, and VM-Edit are provided in Appendix~\ref{appendix:experiments_setup}. 

\paragraph{Overall Results.}

Table~\ref{tab:nrve_multijudge_results} shows that current inference-time video editing methods still struggle with physics-aware non-rigid editing. 
Wan-Edit achieves the highest cross-judge mean score of $31.51$, while VM-Edit ranks second with the lowest cross-judge standard deviation ($30.89$ mean, $3.58$ std), indicating stronger judge stability. 
VM-Edit also achieves the highest physics-plausibility component, with $S_{\mathrm{phy}}=71.31$, suggesting that region-conditioned editing can better preserve material-level deformation cues in some cases. 
The temporal component remains consistently low across methods, with $S_{\mathrm{temp}}=44.35$ for Wan-Edit and $42.80$ for VM-Edit.

\begin{table}[htbp]
    \centering
    \small
    \setlength{\tabcolsep}{3.2pt}
    \renewcommand{\arraystretch}{1.0}
    \caption{
    Quantitative comparison on NRVBench using NRVE-Acc.
    The final NRVE-Acc is computed per instance before averaging
    (not equal to the geometric mean of the displayed component averages).
    Mean and Std summarize cross-judge performance and sensitivity to judge calibration.
    }
    \label{tab:nrve_multijudge_results}
    \resizebox{\linewidth}{!}{
    \begin{tabular}{lcccccccccc}
        \toprule
        \multirow{2}{*}{Method} 
        & \multicolumn{3}{c}{Component Avg.} 
        & \multicolumn{5}{c}{VLM Judges} 
        & \multicolumn{2}{c}{Cross-judge} \\
        \cmidrule(lr){2-4}
        \cmidrule(lr){5-9}
        \cmidrule(lr){10-11}
        & $S_{\mathrm{phy}}\uparrow$ 
        & $S_{\mathrm{temp}}\uparrow$ 
        & $S_{\mathrm{instr}}\uparrow$ 
        & Qwen-7B 
        & Qwen-32B 
        & Qwen-72B 
        & Gemini 
        & Kimi 
        & Mean$\uparrow$ 
        & Std$\downarrow$ \\
        \midrule
        TokenFlow~\citep{geyer2024tokenflow}     
        & 68.16 
        & 41.69 
        & 62.33 
        & 33.05 
        & 34.77 
        & 29.85 
        & 22.95 
        & 28.88
        & 29.90 
        & 4.07 \\

        Pyramid-Edit~\citep{Li2025fivebench}  
        & 59.39 
        & 41.40 
        & \textbf{67.25} 
        & 32.20 
        & 38.02 
        & 29.87 
        & 17.66 
        & 24.42 
        & 28.43 
        & 6.93 \\

        Wan-Edit~\citep{Li2025fivebench}      
        & \uline{69.87} 
        & \textbf{44.35} 
        & \uline{64.25} 
        & 36.88 
        & 34.94 
        & 31.82 
        & 26.53
        & 27.40 
        & \textbf{31.51} 
        & \uline{4.06} \\

        VidToMe~\citep{li2024vidtome}       
        & 69.64 
        & 38.98 
        & 61.46 
        & 30.24 
        & 34.03 
        & 31.41 
        & 20.66 
        & 27.56 
        & 28.78 
        & 4.56 \\

        AnyV2V~\citep{ku2024anyv2v}        
        & 59.24 
        & 39.28 
        & 62.05 
        & 30.35 
        & 33.88 
        & 26.00 
        & 15.33 
        & 24.33 
        & 25.98 
        & 6.29 \\

        \rowcolor{gray!20}
        \textbf{VM-Edit} 
        & \textbf{71.31} 
        & \uline{42.80} 
        & 62.57 
        & 34.71 
        & 34.91 
        & 30.53 
        & 25.55 
        & 28.76 
        & \uline{30.89} 
        & \textbf{3.58} \\
        \bottomrule
    \end{tabular}
    }
\vspace{1pt}
\end{table}

\begin{table}[!t]
    \centering
    \fontsize{6.0}{6.4}\selectfont
    \setlength{\tabcolsep}{2.4pt}
    \renewcommand{\arraystretch}{}
    \caption{
    Comparison of inference-time methods using conventional metrics on NRVBench. 
    }
    \label{tab:benchmark_results}
    \resizebox{\linewidth}{!}{
    \begin{tabular}{@{}lcccccccccc@{}}
        \toprule
        \multirow{2}{*}{Method}
        & \multicolumn{1}{c}{Struct.}
        & \multicolumn{4}{c}{Background Preservation}
        & \multicolumn{2}{c}{Text Align.}
        & \multicolumn{1}{c}{IQA}
        & \multicolumn{1}{c}{Motion}
        & \multicolumn{1}{c}{Speed} \\
        \cmidrule(lr){2-2}
        \cmidrule(lr){3-6}
        \cmidrule(lr){7-8}
        \cmidrule(lr){9-9}
        \cmidrule(lr){10-10}
        \cmidrule(lr){11-11}
        & Dist.$\downarrow$
        & PSNR$\uparrow$
        & LPIPS$\downarrow$
        & MSE$\downarrow$
        & SSIM$\uparrow$
        & CLIP$\uparrow$
        & CLIP$_e\uparrow$
        & NIQE$\downarrow$
        & Mot.$\uparrow$
        & FPS$\uparrow$ \\
        \midrule

        \multicolumn{11}{c}{\textbf{Long-video subset} 
        $(15 \times 3 \times 150\ \mathrm{frames})$} \\
        \midrule
        Wan-Edit~\citep{Li2025fivebench}      
        & 18.04 & 32.36 & 96.76 & 7.99 & 95.66 
        & 27.52 & 24.41 & 10.70 & 60.06 & 2.03 \\
        Pyramid-Edit~\citep{Li2025fivebench}  
        & 84.06 & 22.41 & 128.14 & 77.72 & 87.78 
        & 27.68 & 24.99 & 9.58 & 57.95 & 1.47 \\
        TokenFlow~\citep{geyer2024tokenflow}     
        & 50.52 & 24.19 & 98.68 & 53.96 & 88.38 
        & 26.83 & 24.69 & 9.96 & \textbf{62.70} & 2.96 \\
        VidToMe~\citep{li2024vidtome}       
        & 231.82 & 13.80 & 214.59 & 491.33 & 75.65 
        & 25.00 & 22.85 & 10.37 & 55.78 & \textbf{4.59} \\
        AnyV2V~\citep{ku2024anyv2v}
        & 182.00 & 13.61 & 354.90 & 786.42 & 72.44 
        & 23.31 & 21.97 & 11.63 & 54.63 & 3.55 \\
        \rowcolor{gray!20}
        \textbf{VM-Edit}
        & \textbf{16.45} & \textbf{37.35} & \textbf{58.91} & \textbf{2.99} & \textbf{97.36}
        & \textbf{27.87} & \textbf{25.48} & \textbf{9.04} & 43.69 & 2.34 \\
        \midrule

        \multicolumn{11}{c}{\textbf{Standard NRVBench} 
        $(180 \times 60\ \mathrm{frames})$} \\
        \midrule
        Reference source 
        & 0 & $\infty$ & 0 & 0 & 100 
        & 26.66 & 23.83 & 7.37 & 90.83 & / \\
        Wan-Edit~\citep{Li2025fivebench}      
        & 17.66 & 29.37 & 111.92 & 17.35 & 92.15 
        & 26.63 & 23.24 & 8.28 & 60.65 & 2.67 \\
        Pyramid-Edit~\citep{Li2025fivebench}   
        & 83.30 & 21.73 & 147.40 & 95.24 & 84.50 
        & \textbf{26.65} & 22.47 & 8.26 & 53.89 & 2.78 \\
        TokenFlow~\citep{geyer2024tokenflow}     
        & 111.93 & 21.10 & 134.31 & 118.23 & 75.30 
        & 26.47 & \textbf{23.28} & 7.98 & 58.49 & 3.11 \\
        VidToMe~\citep{li2024vidtome}       
        & 364.93 & 11.80 & 300.64 & 778.40 & 59.78 
        & 26.60 & 22.89 & 8.34 & 57.76 & \textbf{3.89} \\
        AnyV2V~\citep{ku2024anyv2v}        
        & 329.34 & 13.83 & 353.94 & 646.59 & 58.48 
        & 23.32 & 20.75 & 9.47 & 49.14 & 2.76 \\
        \rowcolor{gray!20}
        \textbf{VM-Edit}
        & \textbf{8.69} & \textbf{35.79} & \textbf{47.89} & \textbf{4.74} & \textbf{95.72}
        & 26.15 & 23.19 & \textbf{7.36} & \textbf{60.94} & 2.21 \\
        \bottomrule
    \end{tabular}
    }
\vspace{-3pt}
\end{table}

\begin{figure}[h]
  \centering
  \includegraphics[width=\linewidth]{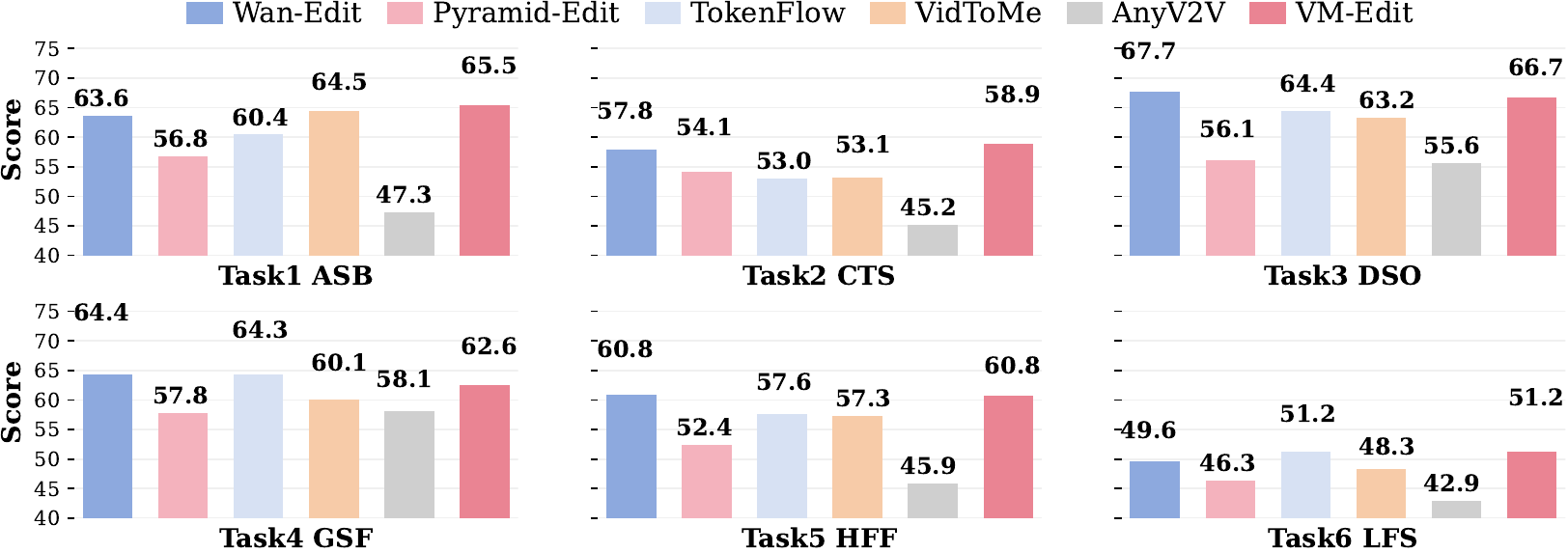}
\caption{
Comparison of six models across the six non-rigid categories. Scores are scaled by $10^2$.
}
  \label{fig:motion_fidelity}
\vspace{-3pt}
\end{figure}

On the standard NRVBench setting, Table~\ref{tab:benchmark_results} shows that VM-Edit achieves the strongest structure and background preservation, obtaining the best Dist. ($8.69$), PSNR ($35.79$), LPIPS ($47.89$), MSE ($4.74$), and SSIM ($95.72$).
Figure~\ref{fig:motion_fidelity} shows motion fidelity across the six categories.
Detailed results for additional metrics, per-category breakdowns, judge-specific analyses, qualitative examples, failure cases and limitations are provided in Appendix~\ref{appendix:experiments_setup}.

\paragraph{Diagnostic Analysis.}

\begin{wraptable}{r}{0.3\textwidth}
\centering
\scriptsize
\setlength{\tabcolsep}{3pt}
\renewcommand{\arraystretch}{0.85}
\vspace{-8pt}
\caption{Rank agreement.}
\label{tab:human_vlm_rank_correlation}
\begin{tabular*}{0.88\linewidth}{@{\extracolsep{\fill}}lcc}
\toprule
Judge & $\rho\uparrow$ & $\tau\uparrow$ \\
\midrule
Qwen-7B    & 0.658 & 0.594 \\
Qwen-32B   & 0.771 & 0.667 \\
Qwen-72B   & 0.886 & 0.733 \\
Kimi-2.5   & 0.914 & 0.800 \\
Gemini-2.5 & 0.943 & 0.867 \\
\midrule
Mean judge & 0.834 & 0.732 \\
\bottomrule
\end{tabular*}

\end{wraptable}
To quantify whether NRVE-Acc preserves human method preferences,
we compute
Spearman's $\rho$ and Kendall's $\tau$ between each VLM judge and human
evaluation across the six evaluated methods. 
As shown in Table~\ref{tab:human_vlm_rank_correlation}, most judges
show positive agreement with human rankings, with stronger agreement observed
for larger or closed-source judges.
Appendix~\ref{appendix:Diagnostic Analysis} provides complete per-category metric tables, negative-control analyses, inter-annotator agreement, per-dimension agreement, and VLM failure cases.

\begin{figure}[h]
  \centering
  \includegraphics[width=\linewidth]{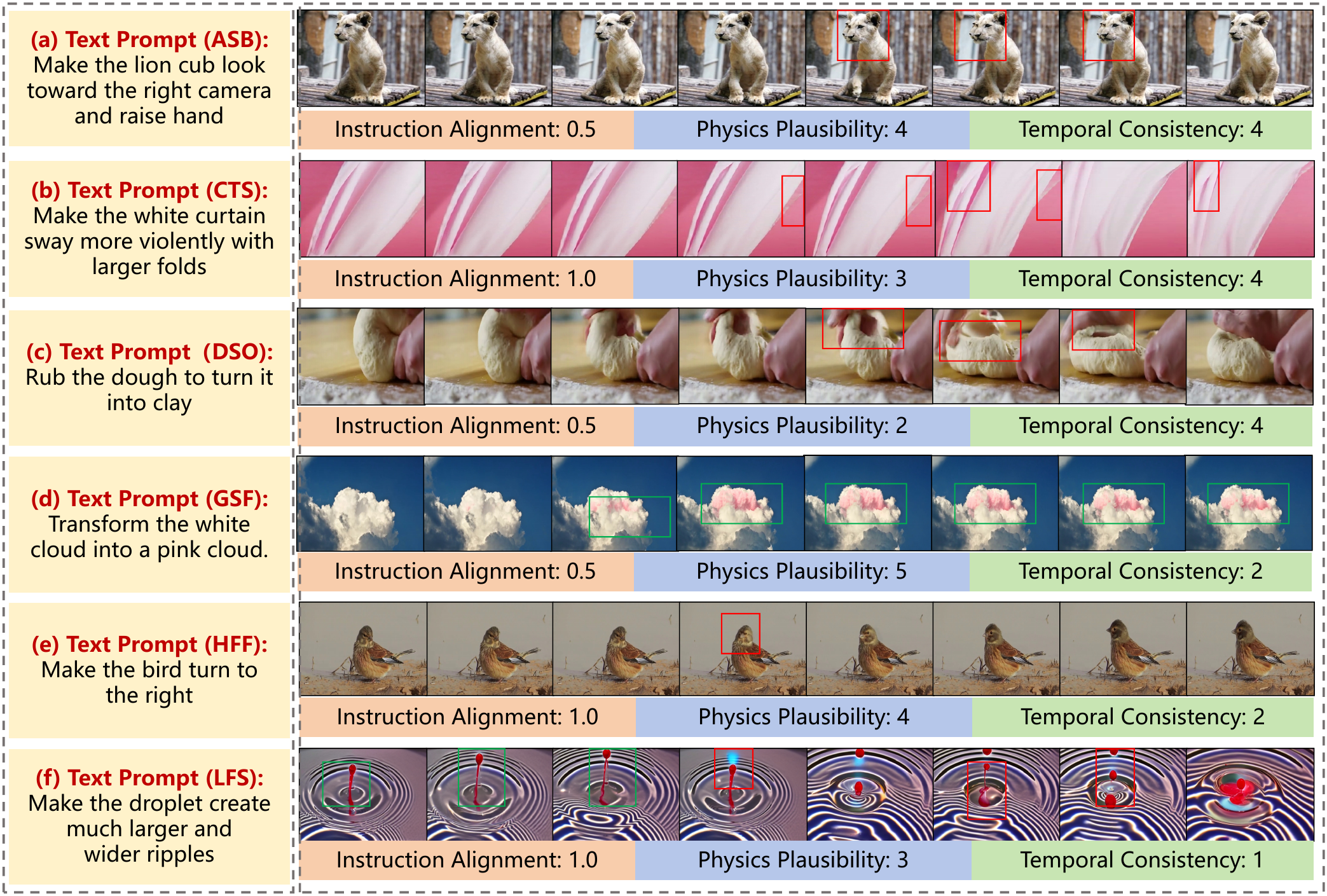}
\caption{
Case studies across the six NRVBench categories. Each row shows sampled frames from one edited video, together with the corresponding text prompt and NRVE-Acc component scores.
}
  \label{fig:case_study}
\end{figure}

\paragraph{Case study.}
Figure~\ref{fig:case_study} presents six representative NRVBench cases covering all physics-grounded non-rigid categories. The examples show that non-rigid video editing failures are often multi-dimensional and cannot be captured by a single appearance or text-alignment metric. They also demonstrate the diagnostic value of NRVE-Acc: by separating instruction alignment, material-specific physical plausibility, and temporal consistency, it identifies whether a failure arises from incomplete editing, implausible deformation, or unstable non-rigid motion.

\vspace{-2pt}
\section{Discussion}
\label{Limitation}

This benchmark reflects a broader shift in video generation and editing: models must increasingly manipulate dynamic scenes, not merely preserve appearance or follow text semantics~\citep{Xue2025PhyT2V,Yang2025VLIPP}.
While existing video editing benchmarks have advanced controllability, object-level editing, and semantic alignment~\citep{sun2025vebench,chen2025editboard,Li2025fivebench}, physical behavior is becoming an equally important evaluation axis. 
Non-rigid editing exposes this need clearly.
Future video editing models should therefore treat physical constraints not as incidental properties of generated videos, but as explicit constraints for controllable editing and evaluation. Beyond VLM-based diagnostics, future extensions may incorporate geometry-aware validators, such as 3D reconstruction, or topology consistency checks, to assess physical properties more directly. As recent work has introduced physical benchmarks for video understanding and generation~\citep{meng2025worldsimulator,bansal2025videophy,chow2025physbench,Li2025towardsvisual}, NRVBench fills a complementary gap in video editing.

\vspace{-2pt}
\section{Conclusion}
\label{conclusion}
We introduced NRVBench and NRVE-Acc for evaluating physics-aware non-rigid video editing.
Existing benchmarks and metrics mainly focus on appearance preservation or semantic alignment, whereas NRVBench emphasizes instruction adherence, material-aware deformation plausibility, and temporal coherence.
Experiments on representative inference-time video editors, together with our VM-Edit method, show that current methods still struggle to produce non-rigid video edits that are both visually coherent and perceptually plausible.
These findings highlight the need for evaluation protocols and video editing models that treat physical dynamics as a central objective rather than an incidental property.
We hope NRVBench provides a reproducible foundation for measuring, diagnosing, and advancing future progress in non-rigid video editing.

\bibliographystyle{plainnat}
\bibliography{references}

\clearpage
\appendix

\section{Datasets}
\label{appendix:datasets}

We provide the hosted NRVBench dataset URL and a validated Croissant metadata file with the required Responsible AI fields in the OpenReview submission. The metadata documents source datasets, provenance activities, intended use cases, limitations, biases, personal or sensitive information, social impact, and synthetic-data status. The evaluation code is available in an anonymous repository at \url{https://anonymous.4open.science/r/NRVBench-755D/}.

\begin{figure}[h]
  \centering
  \includegraphics[width=\linewidth]{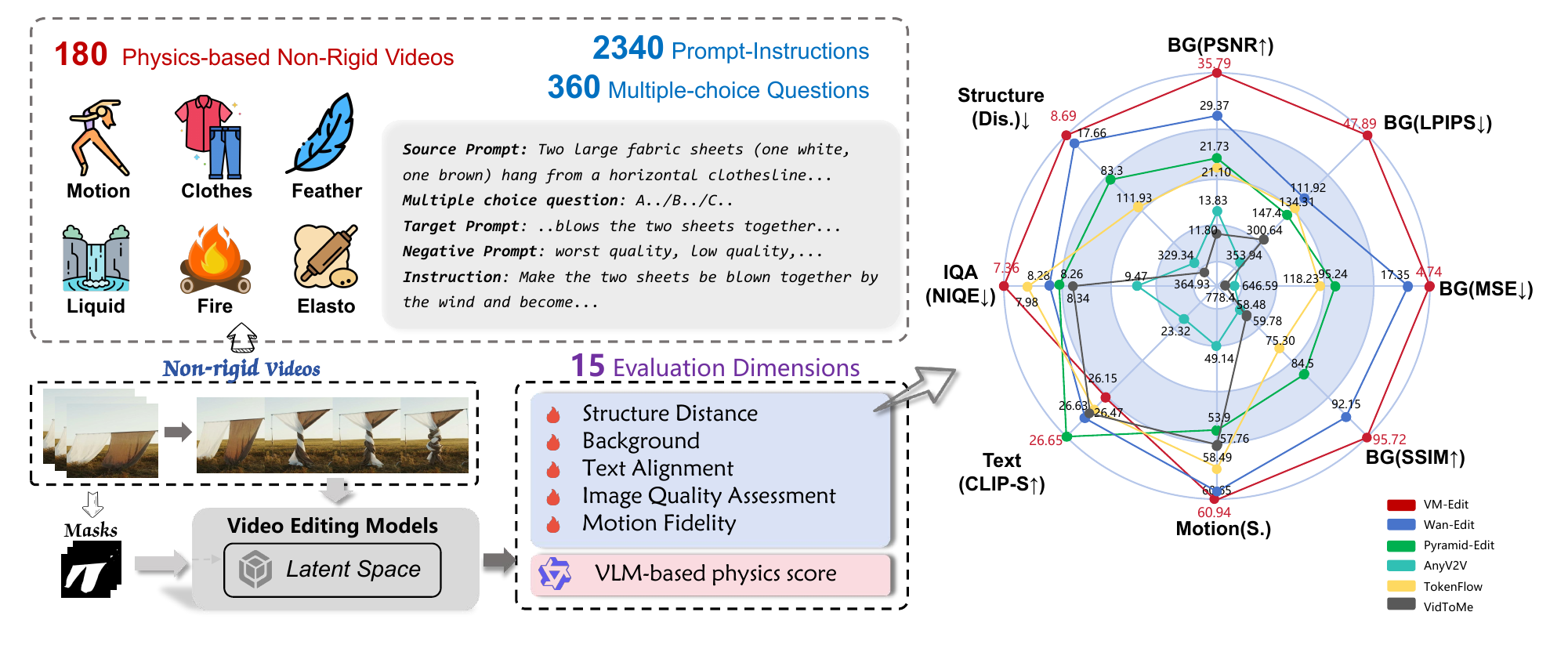}
\caption{Overview of NRVBench.}
  \label{appendix_fig:overreview}
\end{figure}

\subsection{NRVBench Construction}
\paragraph{Video Sources and Category Balance.}
NRVBench is constructed from source videos collected from DAVIS~\citep{davis2016} and Pexels\footnote{\url{https://www.pexels.com/videos/}}, with an overall source ratio of approximately $3{:}1$ in the standard subset. 
DAVIS serves as the primary source due to its high-quality object-centric videos and temporally coherent motion, while Pexels supplements the benchmark with more diverse real-world scenes and object appearances. 
The standard subset covers six non-rigid categories, with 30 videos per category and 60 frames per video, resulting in 180 source videos and 10,800 frames.

In addition, we construct a long-video subset to evaluate editing performance under extended temporal duration. 
This subset contains 15 videos from DAVIS, each with 150 frames. 
The long-video subset is further organized into three editing difficulty levels, allowing us to evaluate how methods behave under increasingly challenging temporal and deformation requirements. 
Together, the standard and long-video subsets contain 195 videos and 13,050 frames in total as shown in Figure~\ref{appendix_fig:overreview}.

\begin{figure}[!h]
  \centering
  \includegraphics[width=\linewidth]{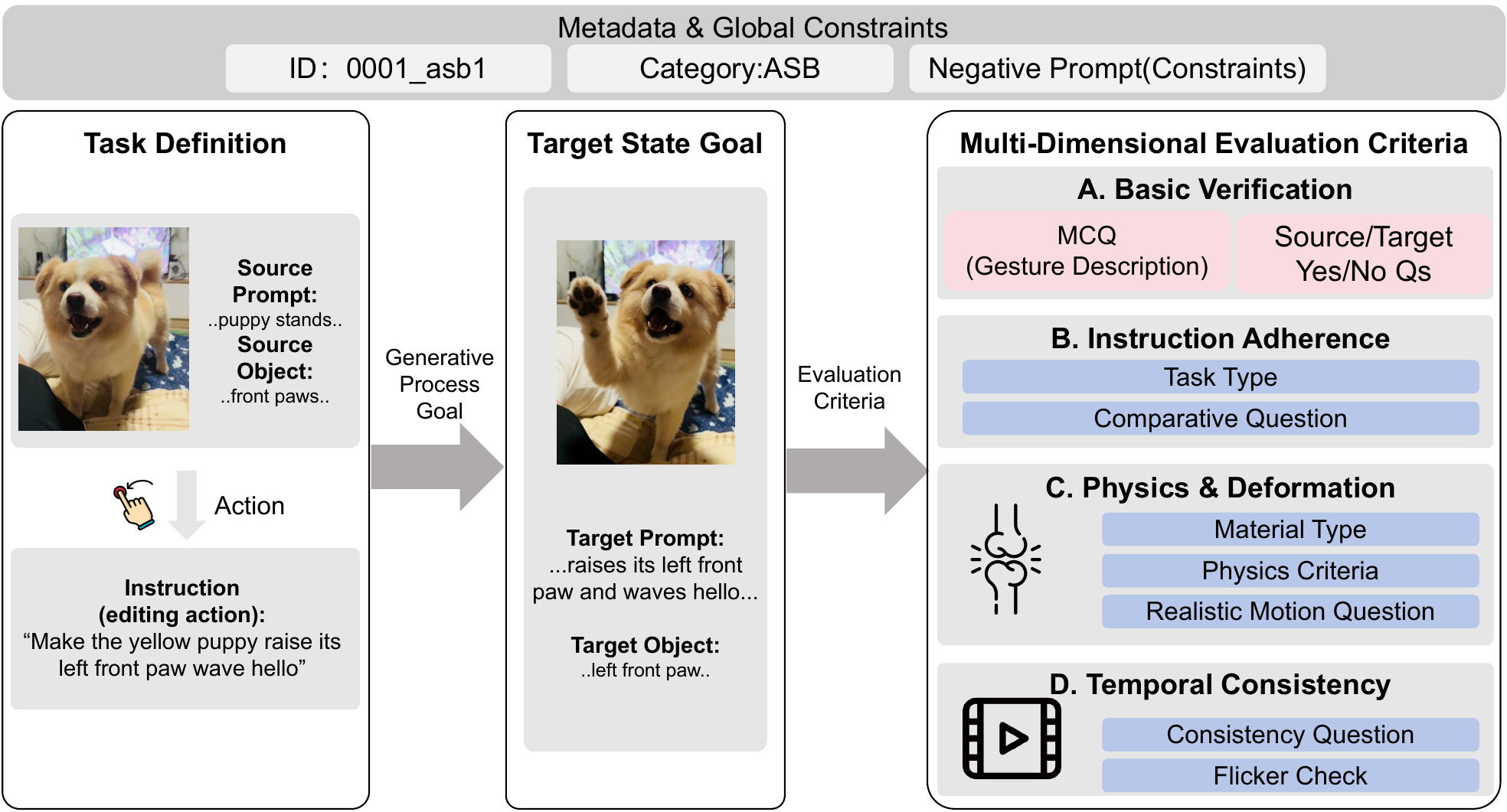}
\caption{Dataset Construction Process.}
  \label{appendix_fig:appendice_json}
\end{figure}

\paragraph{Prompt Generation Templates.}
For the verified video-caption pairs, we design specialized meta-prompts to synthesize target editing instructions across six non-rigid physical categories (e.g., Fluidity, Elasticity). GPT-4o is employed to transmute the original descriptive captions into dynamic target prompts by injecting category-specific deformation constraints while preserving the background context. Figure~\ref{appendix_fig:appendice_json} and table~\ref{appendix_table:dso_edit_entry}  illustrate the prompt framework and provide concrete examples of the mapping from source captions to physics-grounded editing instructions.

\paragraph{Filtering Rules.}
To ensure high visual fidelity, consistent temporal structure, and unambiguous evaluation targets, we apply a set of strict filtering rules to all candidate videos before annotation. First, we retain only videos with a spatial resolution of at least 720p, which provides sufficient detail for pixel-accurate masks and reduces evaluation noise for fine non-rigid structures. We discard samples containing prominent watermarks, large subtitles, or stickers that occlude the target object for a non-trivial portion of frames, as such overlays confound both editing and downstream assessment. We further remove videos with severe compression artifacts (e.g., strong blocking, ringing, or excessive quantization noise) that degrade boundary visibility or introduce spurious temporal inconsistencies. Second, we prefer single-shot segments to avoid abrupt viewpoint changes; clips with evident shot transitions are excluded or re-trimmed to a continuous single-shot interval. Finally, we enforce a single-primary-target constraint: each clip must contain one clearly identifiable main editing subject that remains trackable throughout the selected segment. We allow composite-material scenes (e.g., a person wearing clothes), but require annotators to assign exactly one primary evaluation category per clip (corresponding to the dominant non-rigid attribute under evaluation), while other materials are treated as contextual or secondary factors to prevent category ambiguity.

\begin{algorithm}[t]
\caption{Example construction of a target edit entry for the DSO category.}
\label{appendix_table:dso_edit_entry}
\small
\begin{algorithmic}[1]
\Require Source video $V_{\mathrm{src}}$, source caption $C_{\mathrm{src}}$
\Ensure Target edit entry $E_{\mathrm{edit}}$

\State \textbf{Define editing guidelines for DSO.}
\State \quad Content structure: object, non-rigid action, environment, and camera movement.
\State \quad Object constraints: preserve core identity while emphasizing flexible material properties.
\State \quad The object should remain solid but allow visible shape deformation.
\State \quad Action constraints: use bending, twisting, swaying, crumpling, or other physics-based deformation.
\State \quad Avoid fluid-like melting or purely rigid sliding.
\State \quad Camera and environment should follow the source video and keep the background unchanged.

\State \textbf{Process source input.}
\State \quad $V_{\mathrm{src\_id}} \gets \texttt{0074\_dso14}$
\State \quad $C_{\mathrm{src}} \gets$ ``A hand holds up a bunch of flat pasta noodles which dangle down loosely.''

\State \textbf{Generate target prompt.}
\State \quad Instruction: ``Turn the pasta noodles in the hand into strands of hair.''
\State \quad $C_{\mathrm{tgt}} \gets \mathrm{ApplyEdit}(C_{\mathrm{src}},$
\State \quad \quad ``pasta noodles'' $\rightarrow$ ``strands of hair''$)$
\State \quad Result: ``A hand holds up a bunch of long, dark strands of hair which dangle down loosely.''

\State \textbf{Construct diagnostic questions.}
\State \quad Identity check:
\State \quad \quad $Q_{\mathrm{src}}$: ``Is the hand holding pasta noodles?'' Answer: Yes.
\State \quad \quad $Q_{\mathrm{tgt}}$: ``Is the hand holding strands of hair?'' Answer: Yes.
\State \quad \quad $Q_{\mathrm{MC}}$: ``What is the hand holding?''
\State \quad \quad Options: A) Flat pasta noodles; B) Strands of hair.

\State \quad Physics plausibility check:
\State \quad \quad Criteria: \texttt{strand\_separation}, \texttt{texture\_fidelity}, \texttt{gravity\_compliance}.
\State \quad \quad $Q_{\mathrm{phy}}$: ``Does the material look like fine organic hair strands without merging into a solid block?''

\State \quad Temporal consistency check:
\State \quad \quad $Q_{\mathrm{temp}}$: ``Across frames, does the hair texture remain consistent without flickering back to pasta?''
\State \quad \quad $Q_{\mathrm{flicker}}$: ``Are individual strands temporally coherent without excessive jitter?''

\State \textbf{Compile edit entry.}
\State \quad $E_{\mathrm{edit}} \gets \mathrm{Compile}(V_{\mathrm{src\_id}}, C_{\mathrm{src}}, C_{\mathrm{tgt}},$
\State \quad \quad $\mathrm{Guidelines}, \mathrm{QA\_Metrics})$
\State \Return $E_{\mathrm{edit}}$
\end{algorithmic}
\end{algorithm}

\begin{figure}[!htbp]
  \centering
  \includegraphics[width=\linewidth]{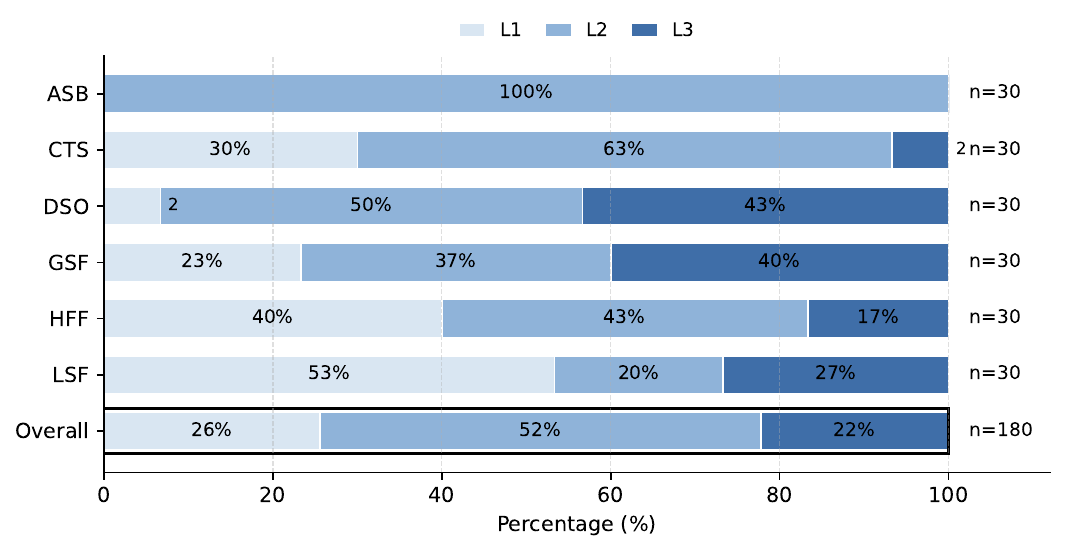}
\caption{Difficulty design.}
  \label{appendix_fig:Difficulty design}
\end{figure}

\paragraph{Difficulty design.}
To support fine-grained diagnosis, we organize editing tasks into three difficulty levels according to the type of non-rigid change required: L1 dynamics modulation, L2 coupled or structural deformation, and L3 topology or material-regime change as shown in Figure~\ref{appendix_fig:Difficulty design}. 
This produces a category-by-difficulty evaluation grid for analyzing model behavior under increasing physical and temporal complexity. 
The distribution is category-aware rather than uniform: most instances are centered on L2, while L1 provides basic dynamics-modulation checks and L3 serves as a stress test for topology-level or material-regime changes. 
ASB is assigned entirely to L2 because its edits are intended to evaluate coupled structural deformation under a fixed material regime; L1 would make the task under-constrained, whereas L3 would change the category semantics and confound the evaluation.

\section{Human Evaluation}
\label{appendix:human_Evaluation}
\subsection{Manual Verification Processes for benchmark construction.}
\begin{figure}[h]
  \centering
  \includegraphics[width=0.8\linewidth]{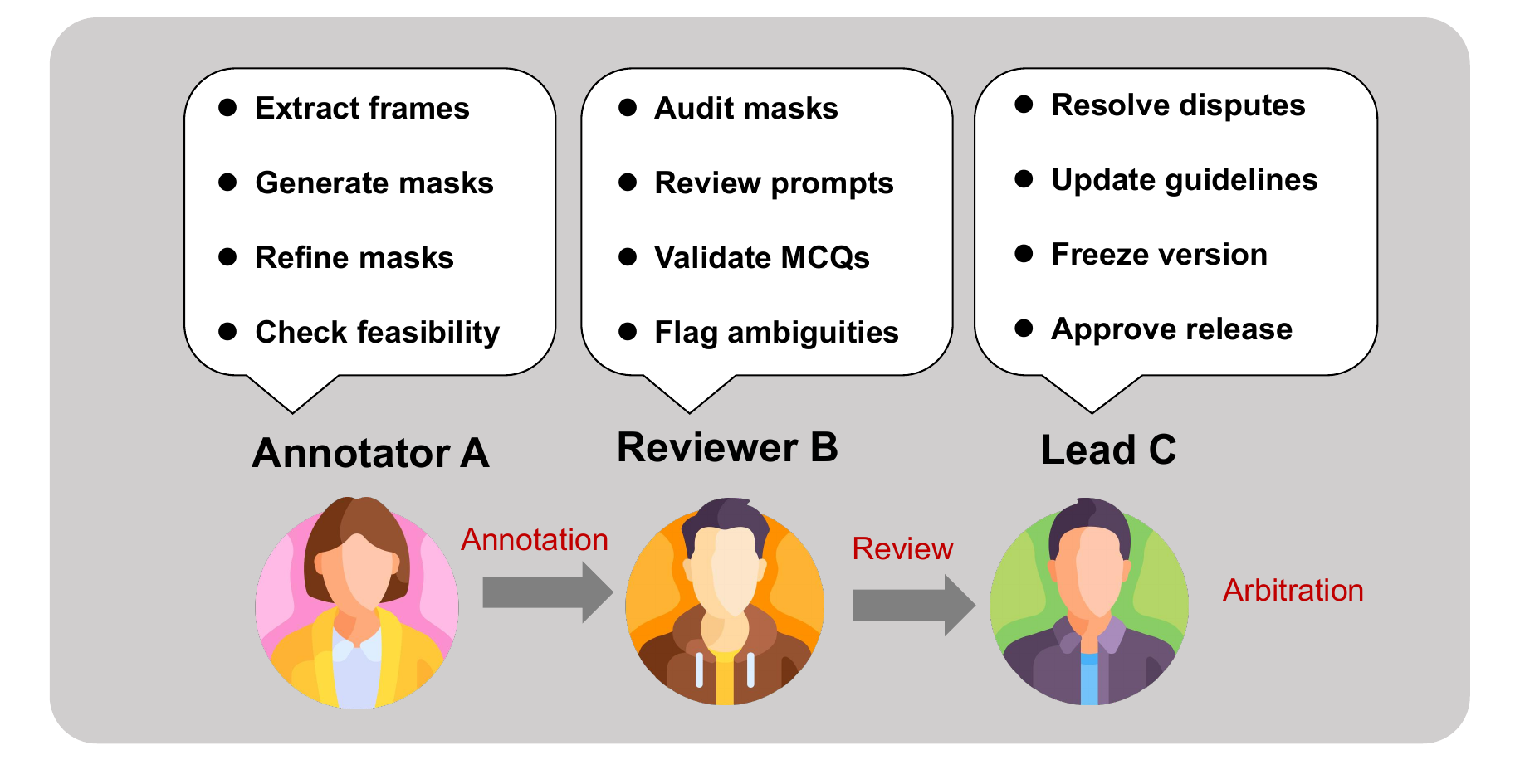}
\caption{Manual Verification Processes.}
  \label{appendix_fig:manual_verrification_process}
\end{figure}

We adopt a three-role manual verification workflow to guarantee pixel-level mask quality, category consistency, and unambiguous instruction/QA design. \textbf{Annotator A} performs primary annotation, including mask refinement, category confirmation, and instruction feasibility checking. Given an input video, we first use ffmpeg to extract and trim a standardized frame sequence, after which SAM2 is used to generate an initial per-frame mask that serves as the annotation starting point. Annotator A then refines masks to ensure accurate boundaries and temporally consistent object identity, and verifies that the associated editing instruction is feasible and visually verifiable for the specified target and its physical/material properties. \textbf{Reviewer B} conducts an independent pass to audit the consistency and potential ambiguity of mask/prompt/MCQs triplets, focusing on common failure modes such as boundary jitter, background leakage into the target mask, unclear target references, non-observable changes, and MCQs with non-mutually exclusive options or multiple plausible answers. When disagreements arise, \textbf{Lead} serves as an adjudicator: resolving conflicts, updating annotation guidelines to reduce future variance, and freezing the finalized version of the sample once it meets quality criteria, as shown in Figure~\ref{appendix_fig:manual_verrification_process}. For challenging categories, we adopt category-specific rules to improve reproducibility. For \textbf{HFF} (hair/feather-like structures), we annotate only the main hair/feather mass rather than individual strands, and explicitly follow a boundary policy centered on the dominant visible region. For \textbf{LFS/GSF} (gas-like phenomena), we prioritize the visible core body (e.g., the main liquid surface or dense smoke plume) and avoid including transparent or semi-transparent trailing wisps, which are inherently ambiguous and would reduce cross-annotator consistency and evaluation repeatability.

\subsection{Human Evaluation for NRVE-Acc}
To validate NRVE-Acc against human judgments, we constructed a stratified evaluation subset from NRVBench. We randomly sampled ten source videos from each of the six non-rigid categories, yielding 60 source videos in total covering L1/L2/L3 difficulty levels. Each video was paired with five editing prompts and four diagnostic questions, covering instruction alignment, material-specific plausibility, and temporal consistency. For each video--prompt pair, we evaluated the outputs of all compared editing methods, forming a balanced human-evaluation subset across categories and methods.

\begin{figure}[htbp]
  \centering
  \includegraphics[
    width=\linewidth,
  ]{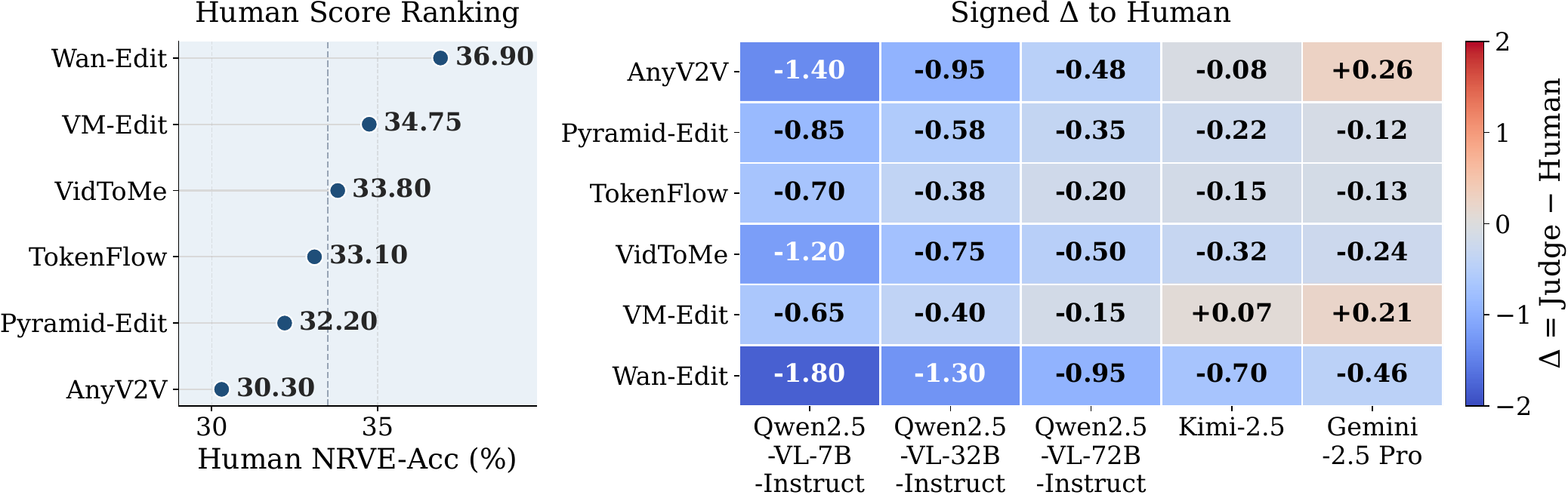}
  \caption{Human scores and signed deviations of VLM Judges. Columns are ordered by average judge performance from worse to better. Signed  > 0 indicates overestimation relative to Human.}
  \label{appendix_fig:human_evaluation}
\end{figure}

To quantify agreement, we report rank correlations between human scores and each VLM judge using Spearman's $\rho$ and Kendall's $\tau$ in Table~\ref{appendix_table:human_vlm_rank_correlation}. 
The results show that NRVE-Acc preserves human method preferences and provides a consistent diagnostic protocol for non-rigid video editing evaluation. 
Figure~\ref{appendix_fig:human_evaluation} further visualizes the human score ranking and the signed deviations of individual VLM judges from human scores. 
Although different judges exhibit different calibration behaviors, their relative rankings remain broadly aligned with human judgments, with stronger judges producing smaller signed deviations. 
We also provide the inter-annotator agreement in Table~\ref{appendix_table:human_inter_annotator}; the results indicate substantial agreement for instruction alignment and physics plausibility, and strong agreement for temporal consistency, suggesting that the human evaluation protocol is reliable despite the perceptual nature of non-rigid deformation assessment.

\begin{table}[htbp]
\centering
\caption{Inter-annotator agreement in human evaluation.}
\label{appendix_table:human_inter_annotator}
\begin{tabular}{lcc}
\toprule
Dimension & Agreement Metric & Score \\
\midrule
Instruction alignment & Fleiss' $\kappa$ & 0.75 \\
Physics plausibility & Krippendorff's $\alpha$ & 0.62 \\
Temporal consistency & Krippendorff's $\alpha$ & 0.81 \\
\bottomrule
\end{tabular}
\end{table}

\begin{table}[htbp]
\centering
\small
\renewcommand{\arraystretch}{0.9}
\caption{Spearman's $\rho$ and Kendall's $\tau$.}
\label{appendix_table:human_vlm_rank_correlation}
\begin{tabular*}{0.45\linewidth}{@{\extracolsep{\fill}}lcc}
\toprule
Judge & $\rho\uparrow$ & $\tau\uparrow$ \\
\midrule
Qwen-7B    & 0.658 & 0.594 \\
Qwen-32B   & 0.771 & 0.667 \\
Qwen-72B   & 0.886 & 0.733 \\
Kimi-2.5   & 0.914 & 0.800 \\
Gemini-2.5 & 0.943 & 0.867 \\
\midrule
Mean judge & 0.834 & 0.732 \\
\bottomrule
\end{tabular*}
\vspace{-4pt}
\end{table}

We recruited five human annotators with graduate-level background in computer vision, graphics, or video generation. None of the annotators participated in dataset construction, prompt writing, mask annotation, or model evaluation. Each edited sample was independently evaluated by five annotators. The evaluation was conducted under a blind and randomized setting: method names, VLM scores, conventional metric scores, and sample ordering were hidden from annotators. Annotators were shown the source video, edited video, editing instruction, and the same diagnostic questions used by the VLM judges.
Human responses were mapped to the same scoring space as NRVE-Acc. For instruction-alignment MCQs, correct, partially correct or ambiguous, and incorrect answers were mapped to $1.0$, $0.5$, and $0.0$, respectively. For physics plausibility and temporal consistency, annotators provided 1--5 ratings, which were normalized to $[0,1]$ following the same protocol as the VLM-based evaluation. We first averaged scores across annotators for each edited sample, then computed the final NRVE-Acc score using the same geometric-mean aggregation as in the automatic metric. Method-level human scores were obtained by averaging over all evaluated samples.

Annotators completed a short calibration session with representative examples and failure cases before the main study. 
They were compensated at RMB 100 per hour. 
The task involved only visual-quality and physical-plausibility judgments; no sensitive personal information or demographic data was collected from annotators. 
Because full human evaluation over all edited videos is costly, we use the stratified human-evaluation subset to validate whether VLM judges preserve human preferences. 
After confirming rank agreement and analyzing judge calibration behavior, we apply VLM judges to full-scale NRVE-Acc evaluation.

\section{Six Non-Rigid Target Categories}
\label{appendix:six_non_rigid_categories}
To cover diverse non-rigid editing regimes, NRVBench taxonomizes source videos and target editing tasks into six physics-grounded categories. 
The taxonomy is designed around the material regime of the edited region rather than only the object name. 
This design allows each category to be evaluated with category-specific physical criteria, while still sharing the same three NRVE-Acc dimensions: instruction alignment, physics plausibility, and temporal consistency. 
For each category, we specify the intended non-rigid behavior, valid editing scope, physical constraints, and common failure modes.

\begin{figure}[!htbp]
  \centering
  \includegraphics[width=\linewidth]{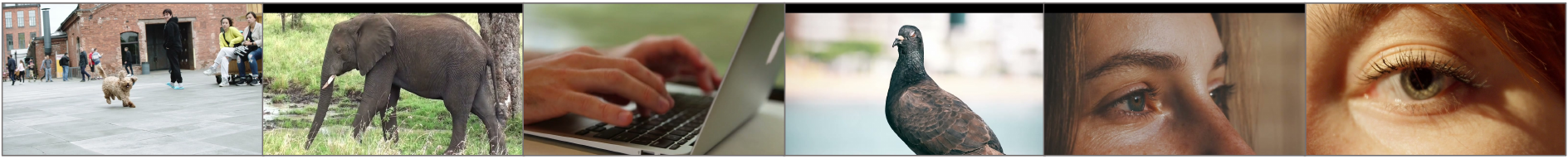}
\caption{Category ASB Examples.}
  \label{appendix_fig:asb}
\end{figure}

\paragraph{ASB: Articulated Soft Bodies.}
Articulated soft bodies include animals, humans, plush-like objects, and other deformable subjects whose motion is governed by both articulation and soft-tissue deformation. 
Typical edits involve changing limb direction, pose, gesture, facial orientation, or localized body motion, while preserving the subject identity and scene layout. 
The key physical criterion is joint-aware deformation: edited limbs or body parts should move according to plausible articulation constraints without excessive stretching, broken topology, or unnatural rubber-like deformation. 
Successful edits should maintain the relative connectivity of body parts and avoid artifacts such as duplicated limbs, ghosting, object popping, or identity drift. 
We use this category to test whether a model can modify articulated motion while preserving anatomically or structurally plausible soft-body behavior as shown in Figure~\ref{appendix_fig:asb}.

\begin{figure}[!htbp]
  \centering
  \includegraphics[width=\linewidth]{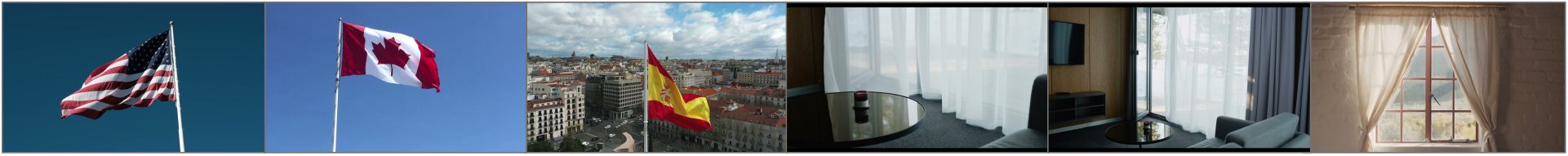}
\caption{Category CTS Examples.}
  \label{appendix_fig:CTS}
\end{figure}

\paragraph{CTS: Cloth and Thin Shells.}
Cloth and thin-shell objects include curtains, flags, sheets, clothing, paper-like surfaces, and other thin deformable surfaces. 
Typical edits involve increasing or decreasing waving strength, changing fold patterns, modifying draping behavior, or producing stronger flapping and bending motion. 
The key physical criterion is surface continuity: the edited surface should deform smoothly while maintaining thin-shell structure. 
A valid edit should preserve the sheet-like topology, maintain coherent folds, and avoid tearing, sudden surface discontinuities, or texture jitter. 
This category is particularly sensitive to temporal artifacts because individual frames may look plausible while folds move inconsistently across time. 
We use CTS to evaluate whether video editors can generate continuous non-rigid surface deformation rather than only changing the appearance of the cloth as shown in Figure~\ref{appendix_fig:CTS}.

\begin{figure}[!htbp]
  \centering
  \includegraphics[width=\linewidth]{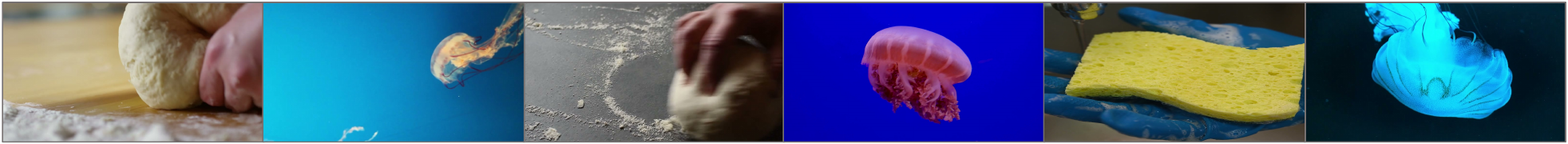}
\caption{Category DSO Examples.}
  \label{appendix_fig:DSO}
\end{figure}

\paragraph{DSO: Deformable Solid Objects.}
Deformable solid objects include dough, clay, rubber-like materials, soft toys, sponges, and other solid objects that can bend, compress, twist, or recover shape under deformation. 
Typical edits involve rubbing, pressing, squeezing, twisting, stretching, or transforming the apparent material while keeping the object solid. 
The key physical criterion is shape integrity under deformation: the object should remain volumetric and materially coherent while exhibiting plausible local deformation. 
Valid edits should preserve object continuity, avoid fluid-like melting unless explicitly required by the instruction, and maintain consistent material identity across frames. 
Common failures include surface-level texture changes without actual deformation, collapsed geometry, unrealistic volume loss, or material-regime mismatch. 
We use DSO to test whether models can edit deformation mechanics rather than only modifying texture or color as shown in Figure~\ref{appendix_fig:DSO}.

\begin{figure}[!htbp]
  \centering
  \includegraphics[width=\linewidth]{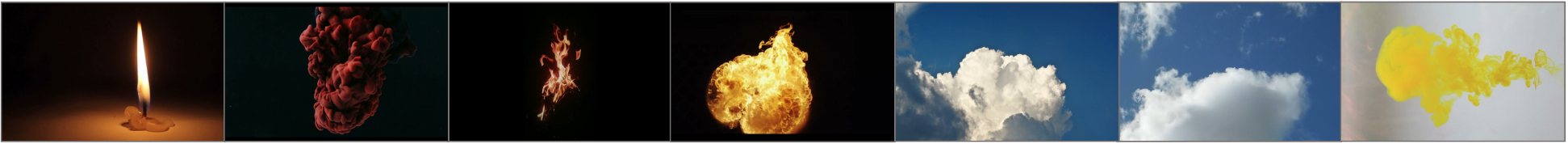}
\caption{Category GSF Examples.}
  \label{appendix_fig:GSF}
\end{figure}

\paragraph{GSF: Gas, Smoke, and Fire.}
Gas, smoke, and fire include clouds, smoke plumes, steam, flames, and other non-solid phenomena with evolving topology. 
Typical edits involve changing color, density, direction, expansion, turbulence, or flame/smoke intensity. 
The key physical criterion is coherent turbulent evolution: the edited region should evolve naturally over time rather than translating rigidly as a texture. 
A successful edit should preserve soft boundaries, diffusion-like motion, and plausible topological change. 
Common failures include frozen smoke, rigid patch motion, abrupt color flicker, inconsistent plume expansion, or frame-wise appearance changes that do not follow turbulent dynamics. 
We use GSF to evaluate whether models can handle non-rigid phenomena whose shapes are not fixed but continuously evolve as shown in Figure~\ref{appendix_fig:GSF}.

\begin{figure}[!htbp]
  \centering
  \includegraphics[width=\linewidth]{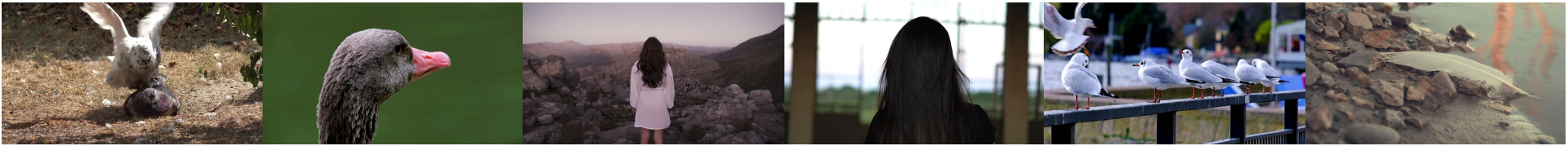}
\caption{Category HFF Examples.}
  \label{appendix_fig:HFF}
\end{figure}

\paragraph{HFF: Hair, Fur, and Feather.}
Hair, fur, and feather include fine fibrous or strand-like structures on animals, humans, and other textured subjects. 
Typical edits involve changing orientation, density, motion direction, fluffiness, waving strength, or local strand response. 
The key physical criterion is strand-level coherence: fine structures should remain locally consistent and should move coherently with the underlying body or external motion. 
Valid edits should preserve strand separation, density, and texture fidelity while avoiding temporal flicker, texture melting, or sudden disappearance of fine details. 
This category is challenging because global object appearance may remain plausible even when local fibers are temporally unstable. 
We use HFF to test whether editing methods can preserve high-frequency non-rigid structures under motion as shown in Figure~\ref{appendix_fig:HFF}.

\begin{figure}[!htbp]
  \centering
  \includegraphics[width=\linewidth]{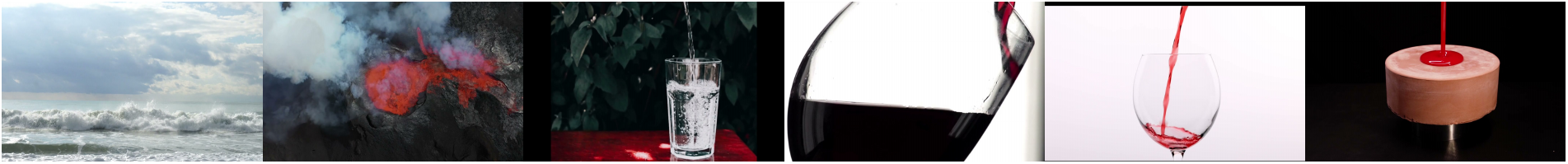}
\caption{Category LFS Examples.}
  \label{appendix_fig:LSF}
\end{figure}

\paragraph{LFS: Liquid Free Surfaces.}
Liquid free surfaces include water surfaces, droplets, ripples, waves, splashes, and other visible liquid-air interfaces. 
Typical edits involve changing ripple size, wave propagation, splash intensity, droplet interaction, or surface disturbance. 
The key physical criterion is fluid-surface coherence: ripples and waves should propagate smoothly and consistently over time. 
A valid edit should maintain plausible surface continuity, fluidity, and local motion propagation instead of producing static texture changes or rigid displacement. 
Common failures include frozen-flow artifacts, unrealistic ripple expansion, inconsistent wave phase, disappearing disturbances, or water behaving like a solid texture. 
We use LFS to evaluate whether models can edit liquid motion while preserving temporally coherent fluid dynamics as shown in Figure~\ref{appendix_fig:LSF}.

\paragraph{Category-specific evaluation.}
For all six categories, the target edit must be visually verifiable, localized to the intended editable region, and distinguishable from an appearance-only modification. 
During annotation, we avoid prompts that can be satisfied by simple color, texture, or style changes unless the task explicitly requires a material-regime change. 
The category label determines the physical criteria used in NRVE-Acc's material-conditioned plausibility query. 
For example, CTS emphasizes fold continuity and thin-shell topology, HFF emphasizes strand coherence, GSF emphasizes turbulent evolution, and LFS emphasizes ripple or wave propagation. 
This category-conditioned design allows the same evaluation protocol to diagnose different non-rigid regimes without collapsing them into a single generic motion score.

\section{Limitations}
\label{appendix:Limitations}
\paragraph{NRVBench.}
NRVBench is designed as a diagnostic benchmark for physics-aware non-rigid video editing rather than a general-purpose video generation dataset or a deployment-readiness test. 
Its coverage is limited to six curated physics-grounded non-rigid categories: articulated soft bodies, cloth and thin shells, deformable solid objects, gas/smoke/fire, hair/fur/feather, and liquid free surfaces. 
Although this taxonomy enables controlled and interpretable evaluation, it may not cover all object categories, material regimes, scene types, physical interactions, or real-world editing tasks.

\paragraph{NRVE-Acc.}
NRVE-Acc relies on VLM judges to evaluate instruction alignment, material-specific physical plausibility, and temporal consistency. 
Although we use multiple VLM judges and compare them with human annotations to reduce dependence on a single evaluator, absolute scores may still be affected by judge calibration, visual sensitivity, language interpretation, and model-specific biases. 
Therefore, NRVE-Acc is best interpreted through controlled within-protocol comparisons rather than as a universal measure of video editing quality.

\paragraph{Future Extensions.}
Moving forward, we plan to extend NRVBench along three directions. First, we will incorporate synthetic non-rigid scenes with controllable material parameters and ground-truth motion fields, enabling more direct validation of physical correctness. Second, we will expand expert pairwise preference annotations to better calibrate material-aware plausibility judgments beyond VLM-based scoring. Third, we will explore 3D- and motion-based measurements, such as reconstructed deformation trajectories and flow consistency, to complement the current perceptual diagnostic protocol. These extensions will further strengthen NRVBench as a bridge between perceptual evaluation and physically grounded video editing assessment.

\section{Experimental Setup}
\label{appendix:experiments_setup}

\subsection{Evaluation protocol.}
We evaluate all methods on NRVBench under a controlled inference-time protocol. 
Each method receives the same source video, target editing instruction, negative prompt, and editing mask when applicable. 
To ensure comparability, we use the same frame sampling strategy, video resolution, and evaluation split across all methods. 
The main benchmark contains 180 videos resampled to 60 frames, and the long-video subset contains 150-frame videos for analyzing extended temporal behavior. 
No method is fine-tuned on NRVBench, and all results are produced in a zero-shot or inference-time editing setting. 
This controlled setup is intended to isolate editing behavior from confounding factors such as benchmark-specific training, additional paired data, or task-specific adaptation.

\subsection{Baselines.}
We compare representative inference-time video editing methods, including TokenFlow~\cite{geyer2024tokenflow}, AnyV2V~\cite{ku2024anyv2v}, VidToMe~\cite{li2024vidtome}, Pyramid-Edit~\cite{Li2025fivebench}, and Wan-Edit~\cite{Li2025fivebench}. 
These methods cover different editing mechanisms, including inversion-based editing, token or feature propagation, token merging, and inversion-free editing. 
We further include VM-Edit, our simple training-free region-conditioned method, to contextualize the stability--plasticity trade-off in localized non-rigid editing. 
All baselines are evaluated with their official implementations or released checkpoints whenever available, using recommended hyperparameters unless otherwise specified. 
For methods that do not natively support mask-conditioned editing, we use their standard text-guided video editing pipeline and evaluate the resulting videos under the same metrics.

\paragraph{TokenFlow.} TokenFlow is a video editing method that does not require training or fine-tuning, based on a pre trained text to image diffusion model for text driven editing of input videos. The core idea is to explicitly propagate diffusion features that are consistent across frames in the feature space of the diffusion model, rather than directly training the video model, in order to obtain temporal consistent video editing results. This implementation is mainly based on Stable Diffusion 2.1, combined with DDIM inversion, Plug and Play feature injection, and TokenFlow attention block to maintain the spatial structure and motion dynamics of the input video.
In the experiments, the input video is first frame extracted and encoded into VAE latent space. Subsequently, DDIM inversion is used to save the noisy latents for each time step, and during the editing stage, denoising sampling is performed based on the target text prompt from the inverted latent trajectory. Save the editing results frame by frame to the \texttt{imd\_od} directory.
For TokenFlow, we use Stable Diffusion v2.1 as the base text-to-image diffusion model and run all experiments on CUDA with a random seed. Before editing, each input video is resized to 512 x 512 and converted into a sequence of frames. The negative prompt is set to “worst quality, low quality, blurry, absolute black, absolute white, low res, extra limbs, extra digits, misplaced objects, mutated anatomy, monochrome, horror”. The number of frames used for editing is selected from the available inverted latents and truncated when necessary so that it is divisible by the editing batch size.

\paragraph{AnyV2V.}
We evaluate AnyV2V using its I2VGen-XL based implementation. AnyV2V is a training-free video-to-video editing framework that reduces video editing to a first-frame image editing problem and then propagates the edited appearance to the full video with an image-to-video diffusion model. In our pipeline, the first frame of each source video is edited by an image editing model according to the target instruction. The source video is then inverted into the latent space of I2VGen-XL using DDIM inversion. During generation, the edited first frame serves as the visual condition, while the inverted source latents provide the motion and structural initialization. Plug-and-Play feature and attention injection are applied during denoising to preserve the temporal dynamics and layout of the original video while following the edited first-frame appearance and target text prompt.
For AnyV2V, we use the \texttt{ali-vilab/i2vgen-xl} checkpoint with half precision and run all experiments on CUDA with a random seed. All videos are processed at a spatial resolution of $512 \times 512$, with $32$ frames and a target frame rate of $8$ fps. The random latent mixing ratio is set to $0.0$, so the initialization uses only the inverted source latent. The editing prompt is the target prompt for each edit, and the negative prompt is ``worst quality, low quality, blurry, absolute black, absolute white, low res, extra limbs, extra digits, misplaced objects, mutated anatomy, monochrome, horror''. The final edited videos are saved as MP4, GIF, and individual PNG frames.

\begin{table}[t]
  \centering
  \scriptsize
  \setlength{\tabcolsep}{2.5pt}
  \renewcommand{\arraystretch}{0.85}
  \caption{Methods, publication venues, and base T2I/V models.}
  \label{appendix_table:method_pub_base_model}
  \resizebox{0.50\linewidth}{!}{%
    \begin{tabular}{@{}lcc@{}}
      \toprule
      \textbf{Method} & \textbf{Venue} & \textbf{Base Model} \\
      \midrule
      TokenFlow~\cite{geyer2024tokenflow} 
      & ICCV'23 & SD2.1 \\
      AnyV2V~\cite{ku2024anyv2v} 
      & TMLR'24 & I2VGen-XL \\
      VidToMe~\cite{li2024vidtome} 
      & CVPR'24 & SD1.5 \\
      Pyramid-Edit~\cite{Li2025fivebench} 
      & ICCV'25 & Pyramid-Flow \\
      Wan-Edit~\cite{Li2025fivebench} 
      & ICCV'25 & Wan2.1 \\
      \textbf{VM-Edit} 
      & -- & \textbf{Wan2.2} \\
      \bottomrule
    \end{tabular}%
  }
\end{table}

\paragraph{Pyramid-Edit.}
Pyramid-Edit is a diffusion-based video editing method built on the Pyramid-Flow architecture. In our setting, the method takes the source video frames, a source prompt, and a target editing prompt as input, and performs editing in the latent space of the Pyramid-Flow video model. The source video is first encoded by the causal video VAE into multi-scale latent representations. During sampling, Pyramid-Edit jointly evaluates source-conditioned and target-conditioned denoising predictions, and updates the target latent trajectory by transferring the edit direction while preserving the source video structure and motion. The method processes the video progressively in a pyramid manner, using previously reconstructed latent conditions to maintain temporal consistency across frames. The final edited latents are decoded back to RGB frames and saved as both individual frames and an edited video.
For Pyramid-Edit, we use the \texttt{pyramid\_flux} variant of the Pyramid-Flow model with the checkpoint \texttt{pyramid-flow-miniflux}. The model is run on CUDA using \texttt{bf16} precision, and VAE tiling is enabled to reduce memory usage. The source and target prompts are taken from the benchmark annotation, and the text encoder appends ``hyper quality, Ultra HD, 8K'' to non-empty prompts. The negative prompt is set to the benchmark-provided negative prompt, i.e., ``worst quality, low quality, blurry, absolute black, absolute white, low res, extra limbs, extra digits, misplaced objects, mutated anatomy, monochrome, horror''. The final edited video is exported as \texttt{edit.mp4} at $12$ fps.

\paragraph{Wan-Edit.}
Wan-Edit is a rectified-flow-based video editing method built on the Wan2.1-T2V-1.3B text-to-video model. Given a source video, a source prompt, and a target prompt, Wan-Edit first encodes the source video into the latent space of the Wan VAE. During generation, it computes denoising predictions under the source prompt, target prompt, and unconditional prompt. The source-conditioned prediction represents the original video dynamics and appearance, while the target-conditioned prediction introduces the desired edit. Wan-Edit updates the moving latent using the difference between target-guided and source-guided predictions, thereby preserving the source motion while steering the content toward the target prompt. The edited latent is decoded by the Wan VAE and saved as the final edited video.
For Wan-Edit, we use the \texttt{Wan2.1-T2V-1.3B} checkpoint with the \texttt{t2v-1.3B} task setting. The model runs on a single CUDA GPU with \texttt{bf16} model parameters. The source prompt is taken from the benchmark \texttt{source\_prompt}, and the target prompt is taken from \texttt{target\_prompt}. The method uses Wan's default negative prompt when no explicit negative prompt is provided. Model offloading is enabled by default in the single-GPU setting to reduce memory usage.

\paragraph{VidToMe.}
VidToMe is a zero-shot video editing method that improves temporal consistency by applying token merging inside the self-attention layers of a pretrained image diffusion model. In our implementation, the method is built on Stable Diffusion and first performs DDIM inversion on the source video to obtain an initial noisy latent representation. During editing, the source video is processed in short temporal chunks. Within each chunk, VidToMe merges similar self-attention tokens across frames to align temporally redundant visual information, and optionally carries global tokens across chunks to encourage long-range consistency. The edited video is generated from the inverted latent using the target text prompt, while depth ControlNet provides structural guidance from the source video frames. This enables the model to preserve the original scene layout and motion while applying the requested semantic edit.
For VidToMe, we use Stable Diffusion v1.5 with the checkpoint \texttt{runwayml/stable-diffusion-v1-5} and run inference on CUDA with \texttt{fp16} precision. The random seed is fixed to $123$, and xFormers memory-efficient attention is enabled when available. All videos are resized and center-cropped to $512 \times 512$. The negative prompt is taken from the benchmark entry, following the form ``worst quality, low quality, blurry, absolute black, absolute white, low res, extra limbs, extra digits, misplaced objects, mutated anatomy, monochrome, horror''. Batch-aligned matching is enabled. Frames are selected according to the configured frame range, and the generated output is saved as \texttt{output.mp4} at $30$ fps together with individual edited frames.

\subsection{VLM judges for NRVE-Acc.}
For NRVE-Acc, we evaluate edited videos with multiple VLM judges to avoid tying the metric to a single proprietary model. 
We use three open-source judges, Qwen2.5-VL-7B-Instruct, Qwen2.5-VL-32B-Instruct, and Qwen2.5-VL-72B-Instruct, together with two closed-source judges, Gemini-2.5 Pro and Kimi-2.5. 
Each judge is queried with the same structured prompts and diagnostic questions. 
NRVE-Acc decomposes evaluation into instruction alignment, physics plausibility, and temporal consistency. 
Instruction alignment is evaluated using multiple-choice diagnostic questions, physics plausibility is evaluated with material-conditioned criteria, and temporal consistency is evaluated using sampled frames together with optical-flow motion cues. 
We report both component scores and cross-judge statistics, including the mean and standard deviation of NRVE-Acc across judges.

\subsection{Scope of NRVE-Acc.}
NRVE-Acc should be interpreted as a perceptual diagnostic metric rather than a simulator-level physical verifier. Non-rigid dynamics such as cloth folding, fluid surfaces, smoke, fire, and soft-body motion are difficult to validate from monocular videos without ground-truth geometry, material parameters, or external force measurements. Our goal is therefore not to certify compliance with physical laws, but to diagnose visually observable failures that commonly arise in non-rigid video editing, such as broken topology, rubber-like deformation, flickering surfaces, object popping, and material-inconsistent motion. To reduce judge dependence, we use material-conditioned questions, mask-guided regions, flow-based temporal evidence, multiple VLM judges, and human-ranking validation. We view NRVE-Acc as a practical proxy for physics-aware perceptual evaluation, complementary to future simulation-based or sensor-grounded metrics.

\subsection{Common evaluation metrics.}
In addition to NRVE-Acc, we report a broad set of conventional video editing metrics to characterize standard editing behavior. 
For structure preservation, we report structure distance. 
For background preservation, we report PSNR, LPIPS, MSE, and SSIM between edited videos and the corresponding source videos. 
For text alignment, we report CLIP-S and CLIP-S$_{\mathrm{edit}}$. 
For perceptual image quality, we report NIQE. 
For motion quality, we report a motion score computed from video motion statistics, and for efficiency we report FPS. 
These common metrics measure appearance preservation, semantic alignment, perceptual quality, motion magnitude, and runtime efficiency, but they do not directly test whether the edited non-rigid deformation is physically plausible. 
Therefore, we use them as complementary diagnostics rather than as the primary evidence for non-rigid editing success.

\subsection{Implementation and compute.}
All experiments are conducted on a single NVIDIA H20 GPU with 96G memory per GPU. 
For each method, we generate videos using the same input frame length and resolution whenever supported by the method. 
Runtime is measured as the average video generation speed in frames per second, excluding metric computation and VLM-judge evaluation unless otherwise stated. 
For VLM-based evaluation, open-source judges are run locally with precision, e.g., bfloat16 inference, while closed-source judges are accessed through their official APIs. 
We fix random seeds whenever supported by the method and use deterministic evaluation scripts for all metric computations.

\paragraph{Metric normalization and reporting conventions.}
We use a single reporting convention for all tables and formulas. All frame-level metrics are first computed per frame and then averaged over frames and videos unless otherwise stated. For background-preservation metrics, PSNR is reported in dB, SSIM is reported as $100\!\times$SSIM, LPIPS is reported as $10^3\!\times$LPIPS, and MSE is reported as $10^4\!\times$MSE computed on RGB values normalized to $[0,1]$. Lower LPIPS and MSE indicate better preservation, while higher PSNR and SSIM indicate better preservation. For text alignment, CLIP-S denotes the average CLIP cosine similarity between edited frames and the target text prompt, multiplied by 100. CLIP-E denotes the directional CLIP editing score, computed as the cosine similarity between the visual edit direction and the textual edit direction, i.e., between $\phi_I(V_{\mathrm{edit}})-\phi_I(V_{\mathrm{src}})$ and $\phi_T(P_{\mathrm{tgt}})-\phi_T(P_{\mathrm{src}})$, also multiplied by 100. The motion score is computed from optical-flow-based motion statistics and should be interpreted as a motion-magnitude diagnostic rather than a standalone quality metric: a larger value indicates stronger apparent motion, but not necessarily more physically plausible or instruction-consistent motion. We therefore use motion score only together with NRVE-Acc and qualitative inspection.

For NRVE-Acc, instruction alignment is evaluated with multiple-choice questions and mapped to $\{1.0,0.5,0.0\}$ for correct, partially correct/ambiguous, and incorrect answers. Physics plausibility and temporal consistency are queried with rubric-based VLM prompts that require the judge to return a single integer rating from 1 to 5, where 1 denotes severe artifacts or incoherent motion and 5 denotes physically plausible or temporally coherent motion. We normalize these ratings by $(r-1)/4$ so that each component lies in $[0,1]$. If a VLM response contains multiple numbers, we parse the final explicitly stated rating; if the response refuses to answer, does not contain a valid option/rating, or violates the required output format after one deterministic retry, we assign the lowest score for that component. The final NRVE-Acc is computed per instance from the normalized component scores using the geometric mean in Eq.~(1), and only then averaged across instances; therefore the reported final NRVE-Acc is not the geometric mean of the displayed component averages.

\section{Diagnostic Analysis}
\label{appendix:Diagnostic Analysis}

\subsection{Negative Control.}
\begin{figure}[htbp]
    \centering
    \includegraphics[width=0.9\linewidth]{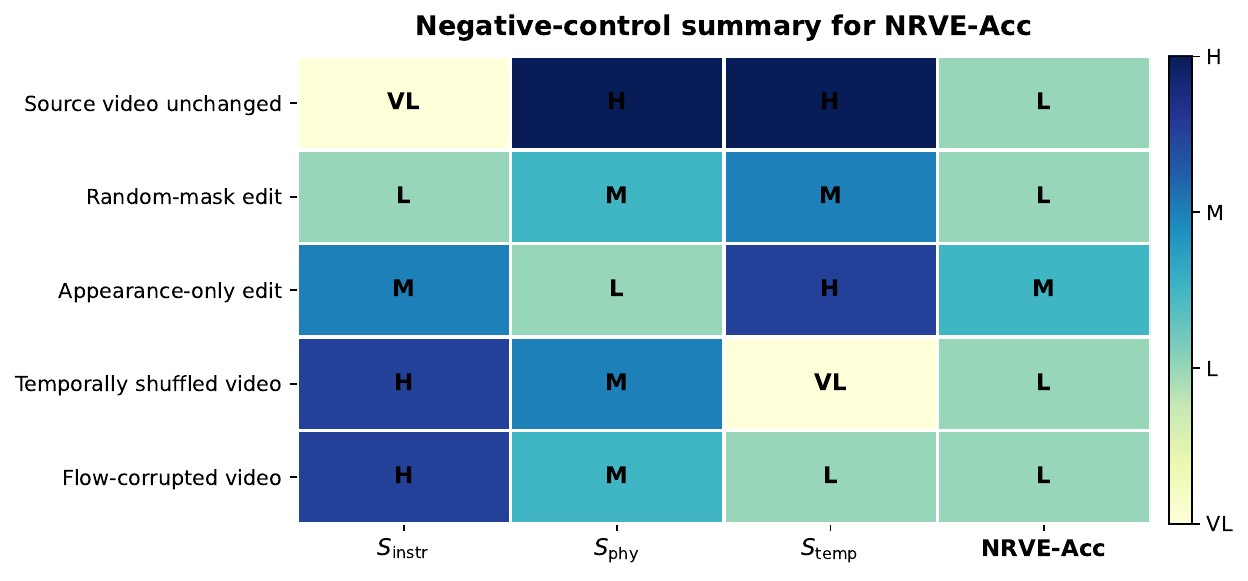}
    \caption{
    Negative-control summary for validating NRVE-Acc. 
    VL, L, M, and H denote very low, low, medium, and high expected scores, respectively. 
    Each control is designed to fail in a predictable component: unchanged source videos fail instruction alignment, random-mask edits fail localized editing correctness, appearance-only edits fail physical deformation, and temporally shuffled or flow-corrupted videos fail temporal consistency.
    }
    \label{appendix_fig:negative_control_heatmap}
\end{figure}

To further verify that NRVE-Acc measures physics-aware editing success rather than superficial visual similarity, we construct negative-control variants on a representative subset of NRVBench. Specifically, we sample five representative instances from each category and generate five negative controls for each sampled instance. Each control is designed to preserve or corrupt a specific subset of factors while keeping the source video, target instruction, and evaluation prompts unchanged. This allows us to test whether the three NRVE-Acc components respond to the intended failure mode: instruction alignment should decrease when the requested edit is absent, physics plausibility should decrease when the deformation violates the material regime, and temporal consistency should decrease when motion evidence is corrupted or temporally incoherent as shown in Figure~\ref{appendix_fig:negative_control_heatmap} and Table~\ref{appendix_table:negative_control_summary}.

\begin{table}[htbp]
\centering
\small
\setlength{\tabcolsep}{5pt}
\renewcommand{\arraystretch}{1.05}
\caption{
Summary of negative controls for validating NRVE-Acc. 
Each control targets a specific failure mode that should be penalized by the corresponding NRVE-Acc component.
}
\label{appendix_table:negative_control_summary}
\begin{tabular}{lp{9.2cm}}
\toprule
Negative control & Main failure mode \\
\midrule
Source video unchanged 
& No requested edit is performed; the video remains natural but instruction-inconsistent. \\

Random-mask edit 
& Changes occur in irrelevant regions or leak into the background, weakening localized editing correctness. \\

Appearance-only edit 
& Color, texture, or style changes are present, but the requested non-rigid deformation is absent or materially implausible. \\

Temporally shuffled video 
& Individual frames may look plausible, but frame order is corrupted, causing severe temporal incoherence. \\

Flow-corrupted video 
& Global frame order is preserved, but local motion cues are corrupted, causing flicker, popping, or discontinuous deformation. \\
\bottomrule
\end{tabular}
\end{table}

\textbf{Source-video unchanged.}
We first evaluate the original source video without applying any edit. This control typically preserves background appearance, visual quality, and natural source motion, and can therefore obtain favorable conventional preservation scores. However, it should receive a low instruction-alignment score whenever the requested target edit is not present. This control tests whether NRVE-Acc penalizes identity-preserving but non-editing outputs, which conventional metrics may incorrectly favor.
\textbf{Random-mask edit.}
We then replace the ground-truth editing mask with a randomly shifted or randomly sampled mask of comparable area before applying the editing pipeline. The resulting video may still contain visible changes, but these changes occur in irrelevant regions or leak into the background. This control tests whether the evaluation is sensitive to spatially localized editing correctness rather than only detecting the presence of any visual change.
\textbf{Appearance-only edit.}
For the appearance-only control, we apply color, texture, or style changes to the target region while suppressing changes in non-rigid motion. Such outputs may improve global text alignment, especially for prompts involving visible attributes, but they fail to realize the requested deformable dynamics. This control directly targets the gap between semantic alignment and physical editing success: a video may look textually related to the prompt while failing to modify cloth folding, fluid propagation, hair motion, articulated motion, or other non-rigid behavior.
\textbf{Temporally shuffled video.}
To isolate temporal reasoning, we construct a temporally shuffled control by permuting the frame order of an edited video while keeping all frames visually unchanged. Since individual frames remain plausible, frame-level appearance metrics and single-frame VLM judgments may remain high. However, the resulting motion contains discontinuities, reversals, or abrupt jumps.
\textbf{Flow-corrupted video.}
Finally, we construct a flow-corrupted control by injecting local motion inconsistencies into the edited video, such as frame-wise local warping, inconsistent object displacement, or optical-flow perturbations within the edited mask. Unlike full frame shuffling, this control preserves the global frame order and appearance but corrupts the local non-rigid motion evidence. It is designed to test whether the temporal branch detects subtle flicker, popping, discontinuous deformation, and motion-field inconsistency.

These controls are not intended to be competitive editing baselines. Instead, they serve as sanity checks for the evaluator: a valid physics-aware metric should not reward unchanged videos, spatially misplaced edits, appearance-only modifications, or temporally incoherent motion as successful non-rigid edits.

\subsection{Per-Dimension Agreement.}
\begin{figure*}[t]
    \centering
    \captionsetup[subfigure]{font=footnotesize}
    \setlength{\tabcolsep}{2pt}

    \begin{subfigure}[t]{0.47\textwidth}
        \centering
        \includegraphics[width=\linewidth]{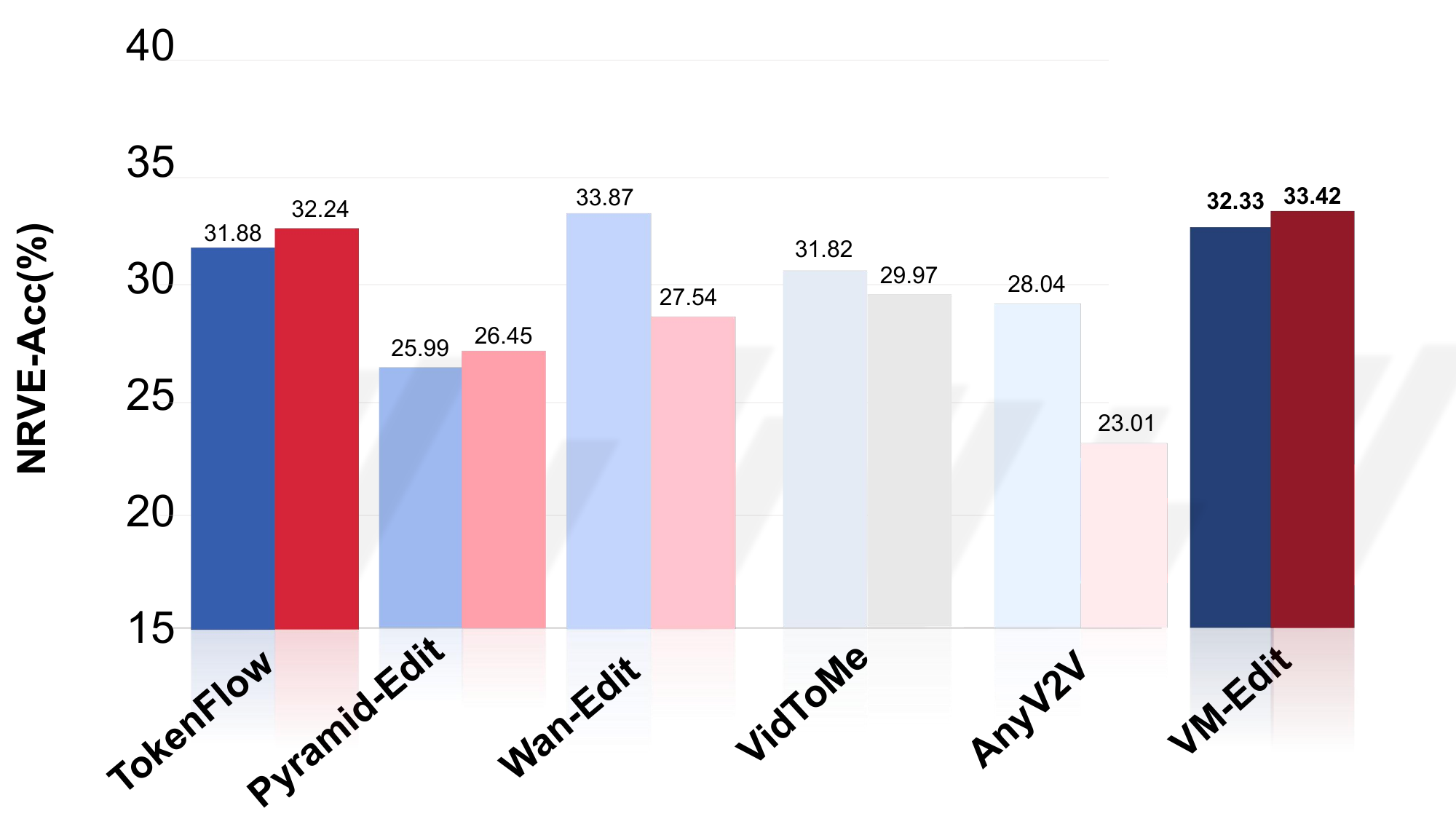}
        \caption{NRVE-Acc results and human evaluation(right) on ASB}
        \label{appendix_fig:appendix_ASB}
    \end{subfigure}
    \hfill
    \begin{subfigure}[t]{0.47\textwidth}
        \centering
        \includegraphics[width=\linewidth]{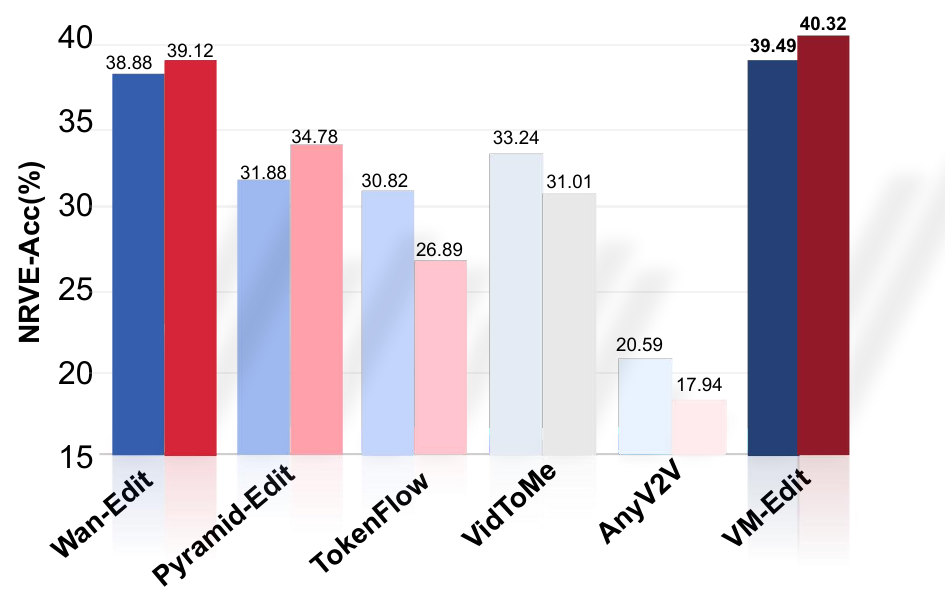}
        \caption{NRVE-Acc results and human evaluation(right) on CTS}
        \label{appendix_fig:appendix_CTS}
    \end{subfigure}

    \begin{subfigure}[t]{0.47\textwidth}
        \centering
        \includegraphics[width=\linewidth]{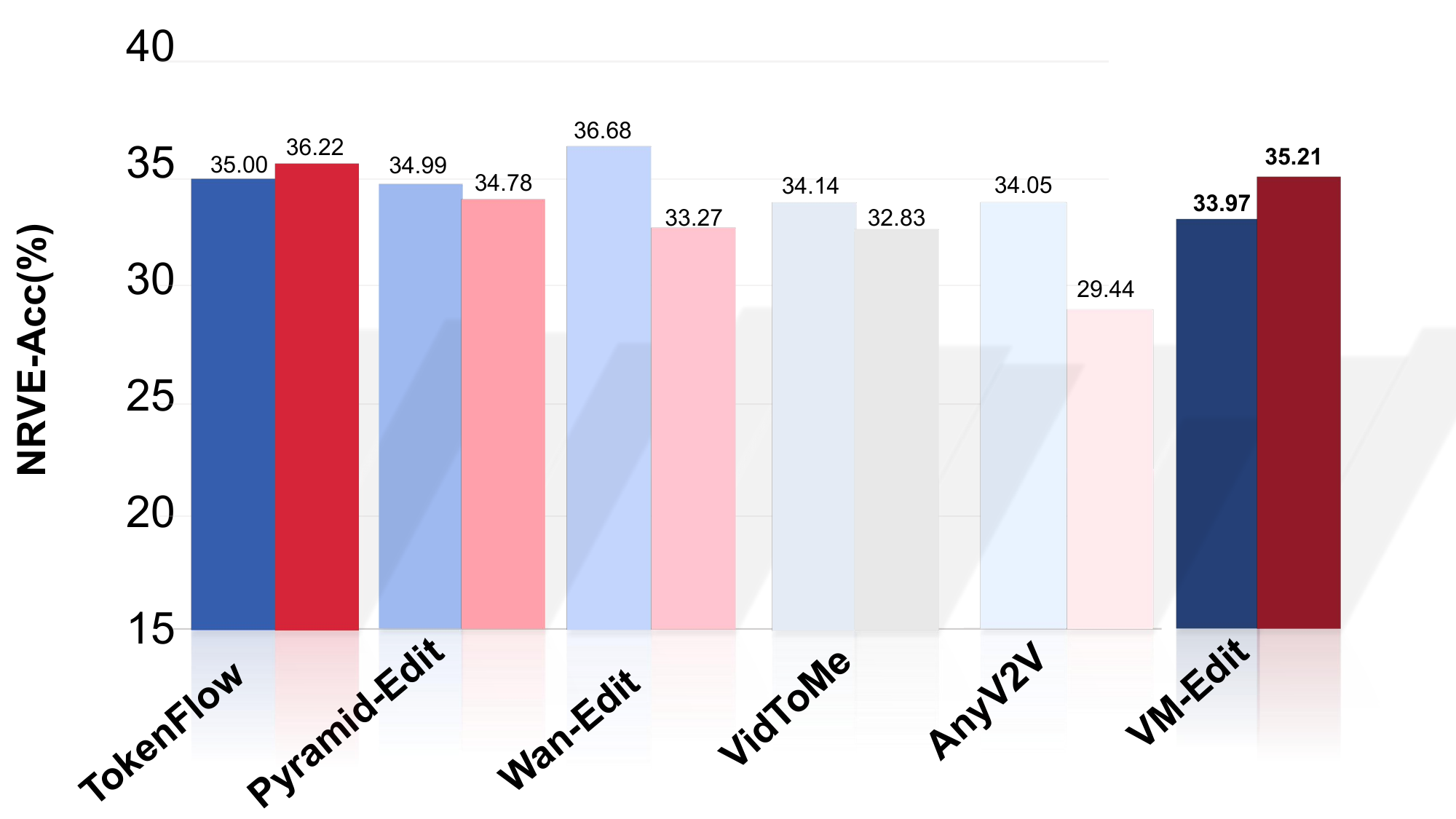}
        \caption{NRVE-Acc results and human evaluation(right) on DSO}
        \label{appendix_fig:appendix_DSO}
    \end{subfigure}
    \hfill
    \begin{subfigure}[t]{0.47\textwidth}
        \centering
        \includegraphics[width=\linewidth]{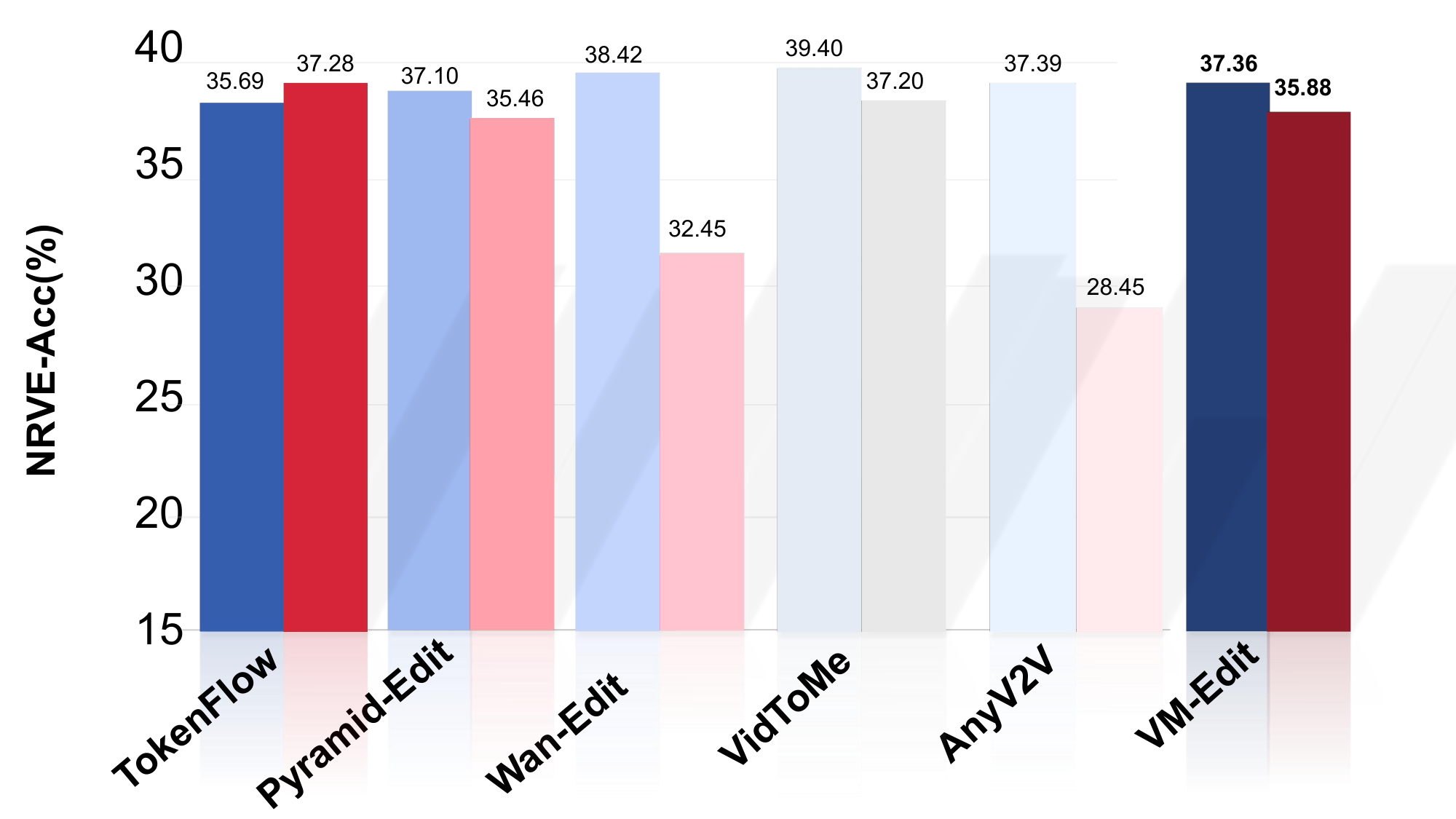}
        \caption{NRVE-Acc results and human evaluation(right) on GSF}
        \label{appendix_fig:appendix_GSF}
    \end{subfigure}

    \begin{subfigure}[t]{0.47\textwidth}
        \centering
        \includegraphics[width=\linewidth]{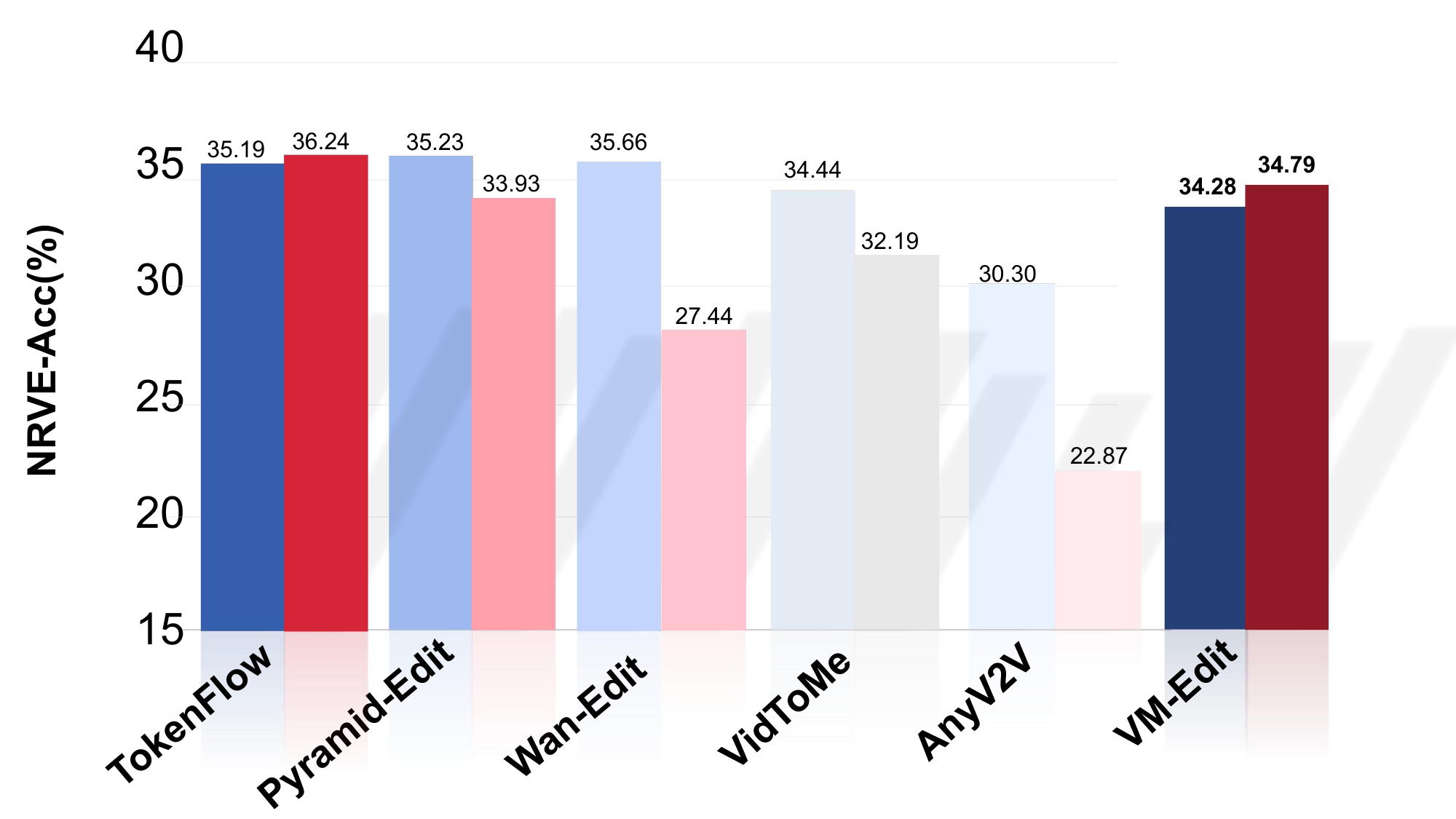}
        \caption{NRVE-Acc results and human evaluation(right) on HFF}
        \label{appendix_fig:appendix_HFF}
    \end{subfigure}
    \hfill
    \begin{subfigure}[t]{0.47\textwidth}
        \centering
        \includegraphics[width=\linewidth]{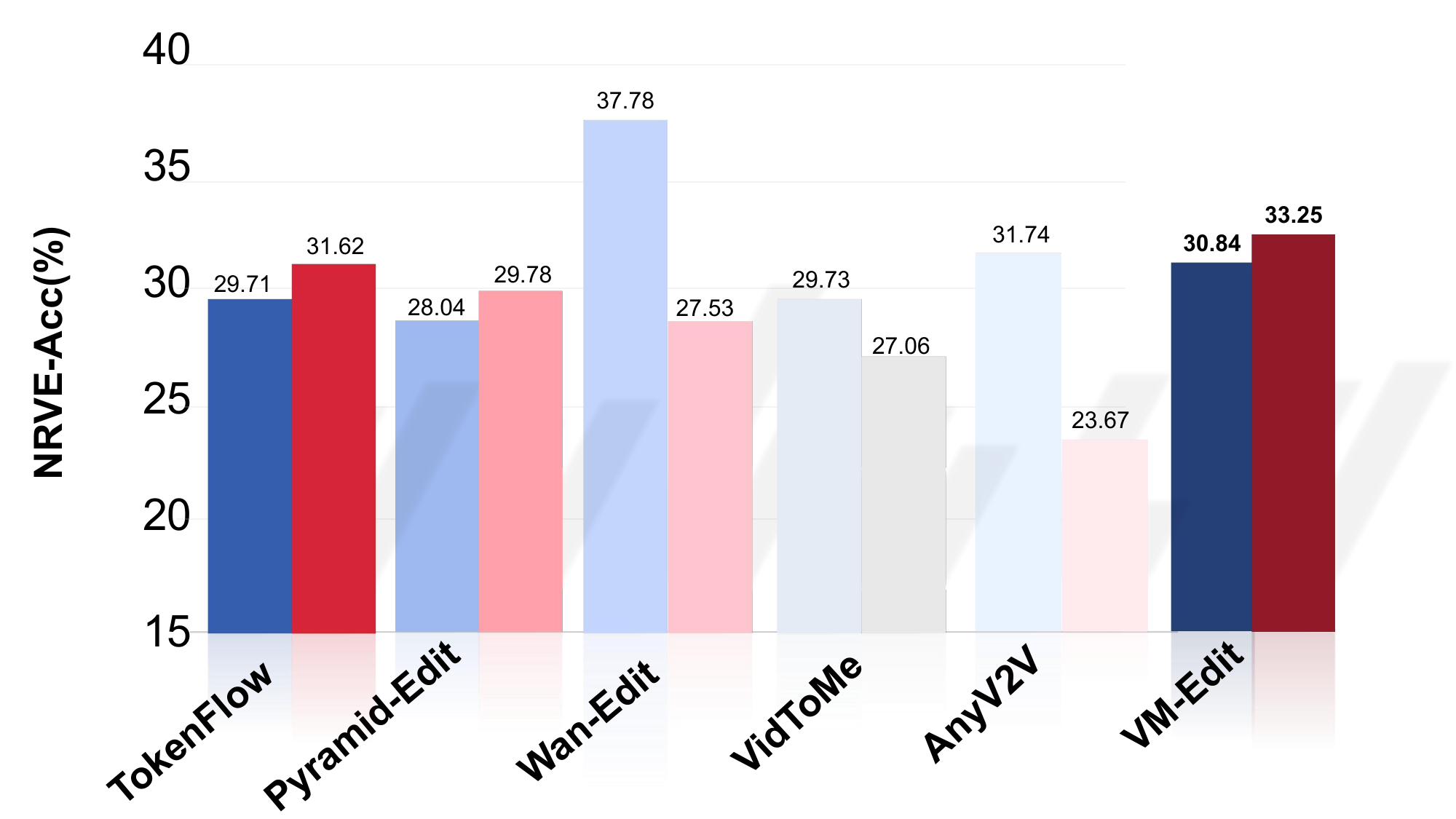}
        \caption{NRVE-Acc results and human evaluation(right) on LFS}
        \label{appendix_fig:appendix_LFS}
    \end{subfigure}

    \caption{
    Per-category comparison between NRVE-Acc results and human evaluation. 
    Each subfigure reports NRVE-Acc results on the left and human evaluation on the right.
    }
    \label{fig:appendix_all_categories}

\end{figure*}

To better understand the category-specific behavior of NRVE-Acc, we conduct a per-dimension human agreement analysis that differs from the aggregate human evaluation used for the overall Human NRVE-Acc. 
For each non-rigid category, we sample 10 representative instances and ask five human annotators to evaluate each edited video using the same scoring rubric as the VLM judges. 
Specifically, annotators independently score instruction alignment, physics plausibility, and temporal consistency, rather than providing only a single overall preference or summed score. 
For each instance, we first average the five annotator scores for each dimension, and then compute category-level means over the 10 sampled instances. 
This yields human reference scores for each category and each NRVE-Acc component. 
We compare these category-level human scores with the corresponding VLM-judge scores to assess whether NRVE-Acc captures the intended failure modes within each category. We compute the resulting category-level NRVE-Acc using the same aggregation rule as the main evaluation and compare it against the VLM-based NRVE-Acc as shown in Figures~\ref{appendix_fig:appendix_ASB},
\ref{appendix_fig:appendix_CTS},
\ref{appendix_fig:appendix_DSO},
\ref{appendix_fig:appendix_GSF},
\ref{appendix_fig:appendix_HFF},
and~\ref{appendix_fig:appendix_LFS}..

\subsection{VLM Failure Cases}
\label{appendix:vlm_failure_cases}

Although NRVE-Acc provides a structured and interpretable protocol for evaluating non-rigid video editing, VLM judges are not perfect physical reasoners. We therefore analyze representative failure cases where VLM judgments diverge from human annotations. This analysis is intended to clarify the limitations of VLM-based evaluation and to explain why we report multiple judges and human-alignment statistics rather than relying on a single proprietary evaluator.

\begin{figure}[t]
  \centering
  \includegraphics[width=\linewidth]{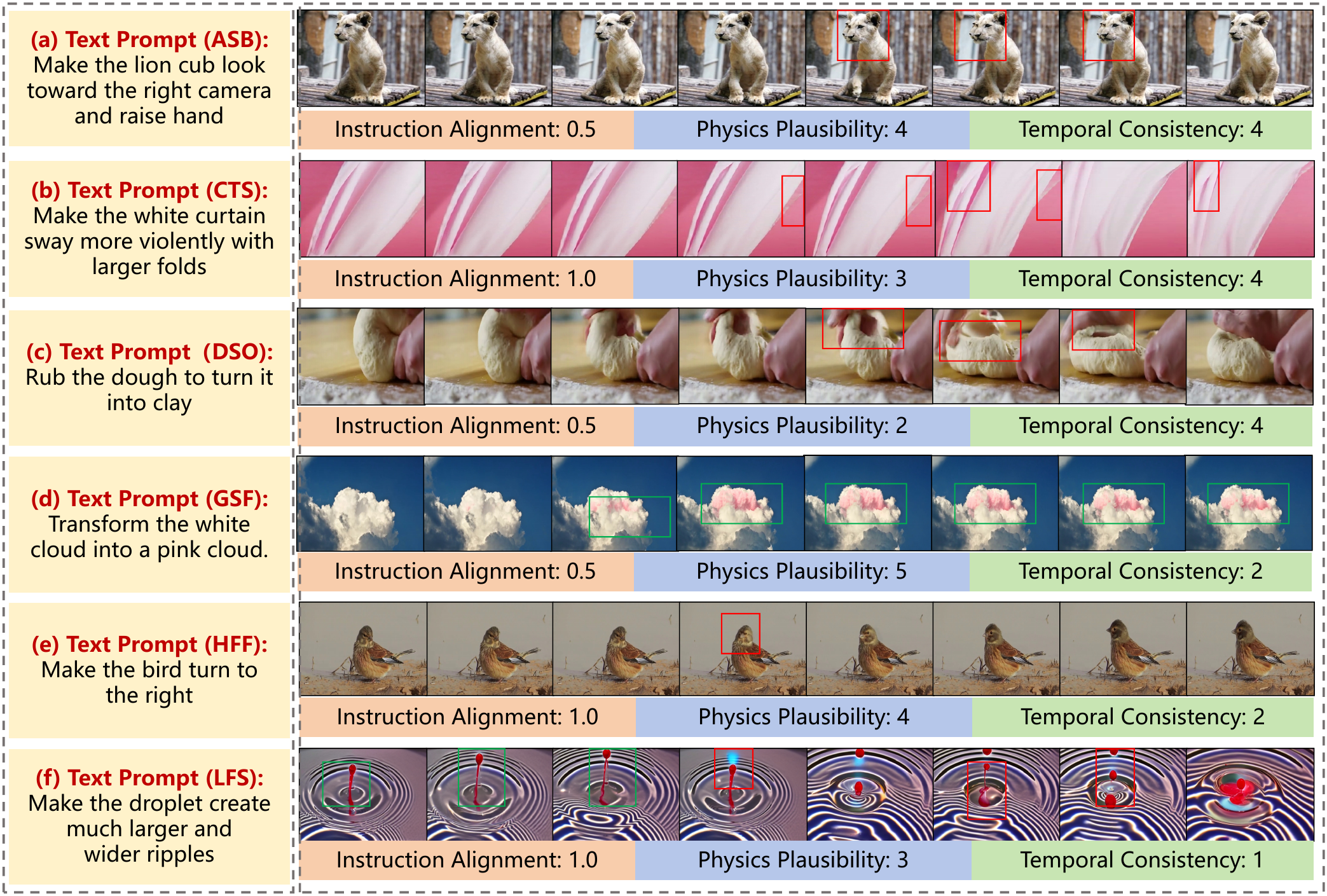}
\caption{
Case studies across the six NRVBench categories.
}
  \label{appnedix_fig:case_study}
\end{figure}

We observe three common types of VLM failure and we also give the case study for each category in Figure~\ref{fig:case_study}. First, VLMs may confuse appearance change with successful non-rigid editing. For example, when a model changes the color or texture of smoke, cloth, or liquid, the judge may assign a high instruction-alignment score even if the requested deformation is absent. Second, VLMs may over-rely on single-frame visual plausibility and under-penalize temporal artifacts such as flicker, popping, or discontinuous deformation. This is especially common in gas, smoke, fire, and liquid-surface cases, where individual frames can appear realistic while motion evolution is physically inconsistent. Third, VLMs may misread material-specific behavior, such as treating a cloth fold, a fluid ripple, or a deformable solid as generic motion rather than evaluating the corresponding material regime. 

These cases do not invalidate NRVE-Acc, but they motivate our judge-aware design. NRVE-Acc decomposes evaluation into instruction alignment, physics plausibility, and temporal consistency, rather than asking for a single holistic judgment. It also uses multiple open- and closed-source VLM judges and compares their rankings with human annotations. As shown in our human-evaluation analysis, judge-specific absolute scores can differ due to calibration, but relative method rankings remain broadly consistent with human judgments. Therefore, we interpret VLM scores primarily through within-judge comparisons, cross-judge statistics, and representative failure cases.

\section{VM-Edit}
\label{appendix:VM-Edit}

\begin{figure}[t]
  \centering
  \includegraphics[width=\columnwidth]{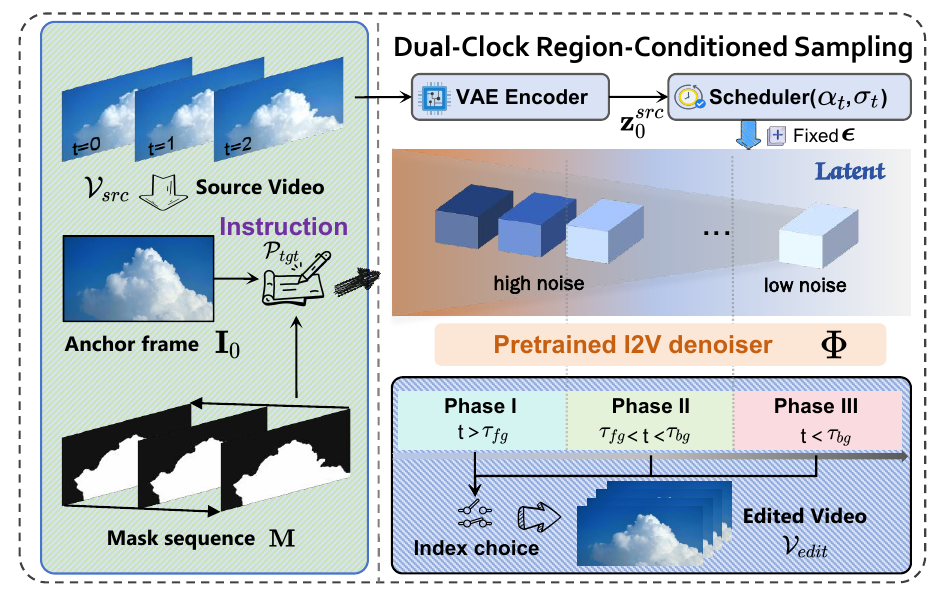}
  \caption{Overview of VM-Edit.}
  \label{appendix_fig:VM-Edit}
\end{figure}

\paragraph{Problem Formulation.}
We formulate non-rigid video editing as synthesizing an edited video sequence $\mathcal{V}_{edit}$ conditioned on a source video $\mathcal{V}_{src}=\{ \mathbf{I}^{src}_i\}_{i=0}^{T-1}$ and a target textual instruction $\mathcal{P}_{tgt}$ as shown in Figure~\ref{appendix_fig:VM-Edit}.
VM-Edit is built on a pre-trained Image-to-Video (I2V) diffusion transformer (Wan2.2~\cite{wang2025wan}) and operates in a strictly training-free manner by modifying only the sampling procedure.
Inference takes four standardized inputs:
(1) an Appearance Anchor $\mathbf{I}_0=\mathbf{I}^{src}_0$ (the first frame) to preserve lighting/texture;
(2) the Target Instruction $\mathcal{P}_{tgt}$ specifying the desired non-rigid edit;
(3) the Source Prior $\mathcal{V}_{src}$, which is encoded into a latent reference trajectory and injected via region-wise anchoring to provide motion and structure;
and
(4) a Spatial Layout mask sequence $\mathbf{M}$. We detail the mask details, systematic guide, and permissible editing strength in the appendices \emph{C}.

\paragraph{Latent Reference Construction.}
Let $\text{Enc}(\cdot)$ be the pre-trained video VAE encoder of the I2V backbone, producing a latent video tensor $\mathbf{z}_0$.
We encode the source video to obtain a source-consistent reference by applying the encoder to the source video, i.e., $\mathbf{z}^{src}_{0} = \text{Enc}(\mathcal{V}_{src})$.
To anchor regions at a matched noise level during sampling, we construct a noised source reference at each diffusion step $t$ using the same noise schedule as the backbone:
\begin{equation}
\mathbf{z}^{src}_t = \alpha_t\,\mathbf{z}^{src}_0 + \sigma_t\,\boldsymbol{\epsilon}, \qquad \boldsymbol{\epsilon}\sim\mathcal{N}(0,\mathbf{I}),
\label{eq:source_noising}
\end{equation}
where $(\alpha_t,\sigma_t)$ follow the model's scheduler and $\boldsymbol{\epsilon}$ is a standard Gaussian noise tensor with the same shape as $\mathbf{z}^{src}_0$, sampled once per video and reused for all $t$. For reproducibility, we sample $\boldsymbol{\epsilon}$ once per video and reuse it across timesteps to obtain a consistent reference trajectory across noise levels.

\paragraph{Dual-Clock Region-Conditioned Sampling.}
Let $\mathbf{z}_t$ denote the current latent at reverse diffusion step $t$ (larger $t$ indicates higher noise). Let $\Phi(\cdot)$ be one reverse-step update of the pre-trained denoiser/sampler, conditioned on $\mathcal{P}_{tgt}$ and the anchor frame $\mathbf{I}_0$:
\begin{equation}
\hat{\mathbf{z}}_{t-1} = \Phi(\mathbf{z}_t, t, \mathcal{P}_{tgt}, \mathbf{I}_0).
\label{eq:denoise_update}
\end{equation}
We enforce region-wise control by recomposing the next latent with a \emph{foreground edit} term and a \emph{background stabilization} term:
\begin{equation}
\mathbf{z}_{t-1} \leftarrow \mathbf{m}\odot \mathbf{z}^{fg}_{t-1} + (\mathbf{1}-\mathbf{m})\odot \mathbf{z}^{bg}_{t-1},
\label{eq:recompose}
\end{equation}
where $\odot$ is element-wise multiplication and $\mathbf{m}$ is downsampled to the latent resolution (details below).

\paragraph{Permissible Editing Strength via Two Timesteps.}
To explicitly allocate plasticity between regions, we use two timestep indices $\tau_{fg}$ and $\tau_{bg}$ (with $\tau_{fg}>\tau_{bg}$). Following an SDEdit-style initialization at $t=\tau_{fg}$ , we enforce a plasticity gap during the interval $\tau_{fg} \ge t > \tau_{bg}$. In this phase, the foreground evolves freely via the diffusion model to accommodate non-rigid deformations, while the background is explicitly anchored to the source latent trajectory to suppress temporal flicker. Once $t \le \tau_{bg}$, both regions are refined jointly to ensure boundary coherence.

\paragraph{Adaptive Proposal.}
Let $\mathbf{z}_t$ denote the current latent at reverse diffusion step $t$, where larger $t$ indicates higher noise. 
Let $\Phi(\cdot)$ be one reverse-step update of the pre-trained denoiser/sampler, conditioned on the target prompt $\mathcal{P}_{tgt}$ and the anchor frame $\mathbf{I}_0$:
\begin{equation}
\hat{\mathbf{z}}_{t-1} = \Phi(\mathbf{z}_t, t, \mathcal{P}_{tgt}, \mathbf{I}_0).
\label{eq:denoise_update}
\end{equation}

The scheduler computes two normalized signals: an edit-demand score $d_{\mathrm{edit}}$ and a coupling-risk score $d_{\mathrm{risk}}$. 
The edit-demand score measures how much freedom the foreground region needs for non-rigid deformation, while the coupling-risk score measures how likely independent foreground editing will introduce boundary artifacts or background instability.

We define the edit-demand score as
\begin{equation}
d_{\mathrm{edit}}
=
\mathrm{Clip}_{[0,1]}
\left(
\lambda_p D_{\mathrm{txt}}
+
\lambda_m D_{\mathrm{mot}}
+
\lambda_a D_{\mathrm{area}}
\right),
\label{eq:edit_demand}
\end{equation}
where $D_{\mathrm{txt}}$ is the normalized text distance between the source and target prompts, $D_{\mathrm{mot}}$ is the foreground motion magnitude estimated from the source video, and $D_{\mathrm{area}}$ is the relative mask area. 
Intuitively, stronger semantic change, larger foreground motion, or larger edited regions require a higher-noise starting point and thus more foreground plasticity.

We define the coupling-risk score as
\begin{equation}
d_{\mathrm{risk}}
=
\mathrm{Clip}_{[0,1]}
\left(
\lambda_b D_{\mathrm{bd}}
+
\lambda_g D_{\mathrm{bg}}
\right),
\label{eq:coupling_risk}
\end{equation}
where $D_{\mathrm{bd}}$ measures mask-boundary complexity and $D_{\mathrm{bg}}$ measures background motion magnitude. 
A higher coupling risk indicates that foreground and background should be jointly refined earlier to avoid boundary discontinuity, flicker, or texture leakage.

The foreground clock and the background clock are then determined automatically:
\begin{equation}
\tau_{fg}
=
\tau_{\min}
+
d_{\mathrm{edit}}
(\tau_{\max}-\tau_{\min}),
\label{eq:adaptive_tau_fg}
\end{equation}
\begin{equation}
\Delta_{\tau}
=
\Delta_{\min}
+
d_{\mathrm{edit}}(1-d_{\mathrm{risk}})
(\Delta_{\max}-\Delta_{\min}),
\qquad
\tau_{bg}
=
\max(\tau_{\min}, \tau_{fg}-\Delta_{\tau}).
\label{eq:adaptive_tau_bg}
\end{equation}
Thus, videos requiring stronger non-rigid deformation start from a higher-noise foreground state, while videos with complex boundaries or unstable backgrounds enter joint refinement earlier. 
This removes the need for per-video manual timestep selection and makes the sampling schedule dependent on the physical and structural properties of the input.

Given the adaptive clocks, VM-Edit recomposes the next latent with a foreground edit term and a background stabilization term:
\begin{equation}
\mathbf{z}_{t-1}
\leftarrow
\mathbf{m}\odot \mathbf{z}^{fg}_{t-1}
+
(\mathbf{1}-\mathbf{m})\odot \mathbf{z}^{bg}_{t-1},
\label{eq:recompose}
\end{equation}
where $\mathbf{m}$ is the editing mask downsampled to the latent resolution. 
For $\tau_{fg} \ge t > \tau_{bg}$, the foreground follows the target-conditioned denoising trajectory to allow non-rigid deformation, while the background is anchored to the source trajectory to suppress temporal flicker. 
For $t \le \tau_{bg}$, both regions are jointly refined to improve boundary coherence and global consistency. 

\section{Detailed Results Analysis}
\label{appendix:detailed_results_analysis}

\subsection{Common Metrics}
Table~\ref{appendix_table:benchmark_results} reports the conventional metric comparison on NRVBench. 
Overall, VM-Edit achieves the best performance on most quality-oriented metrics. 
Excluding the unedited reference source, VM-Edit ranks first on 8 out of 10 metrics on the long-video subset and 7 out of 10 metrics on the standard NRVBench setting. 
These results indicate that VM-Edit consistently improves structural fidelity, background preservation, and perceptual image quality, while maintaining competitive text alignment and runtime efficiency.

\paragraph{Standard NRVBench.}
On the standard NRVBench setting, VM-Edit shows even stronger improvements in structural and background-preservation metrics. 
VM-Edit reduces the structural distance from $17.66$ for Wan-Edit to $8.69$, corresponding to a $50.8\%$ relative reduction. 
For background preservation, VM-Edit again achieves the best results among all editing methods, with a PSNR of $35.79$, LPIPS of $47.89$, MSE of $4.74$, and SSIM of $95.72$. 
Compared with Wan-Edit, VM-Edit improves PSNR by $6.42$ dB, reduces LPIPS by $57.2\%$, reduces MSE by $72.7\%$, and improves SSIM by $3.57$. 
These results show that VM-Edit is particularly effective at suppressing unwanted changes to the source video while performing the requested edit.
For text alignment, VM-Edit remains competitive but does not achieve the best score on the standard benchmark. 
Pyramid-Edit obtains the highest CLIP score of $26.65$, while TokenFlow obtains the highest CLIP$_e$ score of $23.28$. 
VM-Edit achieves $26.15$ and $23.19$ on these two metrics, trailing the best methods by only $0.50$ and $0.09$, respectively. 
This indicates a mild trade-off between stronger visual preservation and text-alignment scores under conventional CLIP-based evaluation.
VM-Edit achieves the best NIQE score of $7.36$, slightly outperforming even the reference source score of $7.37$. 
This suggests that the generated videos maintain high perceptual quality and avoid severe artifacts. 
In contrast to the long-video subset, VM-Edit also obtains the best motion score among editing methods on standard NRVBench, reaching $60.94$, slightly higher than Wan-Edit's $60.65$. 
This demonstrates that VM-Edit can preserve or recover natural motion patterns in standard-length videos while maintaining superior visual fidelity.

\paragraph{Long-video subset.}
On the long-video subset, VM-Edit demonstrates strong robustness under extended temporal contexts. 
For structural consistency, VM-Edit obtains the lowest distance score of $16.45$, improving over Wan-Edit, the strongest method in this category, by $8.8\%$. 
The advantage becomes more pronounced in background preservation. 
Compared with Wan-Edit, VM-Edit improves PSNR from $32.36$ to $37.35$, reduces LPIPS from $96.76$ to $58.91$, reduces MSE from $7.99$ to $2.99$, and increases SSIM from $95.66$ to $97.36$. 
In relative terms, this corresponds to a $39.1\%$ reduction in LPIPS and a $62.6\%$ reduction in MSE, showing that VM-Edit better preserves non-edited regions over long videos.
VM-Edit also achieves the best text-alignment scores on the long-video subset, with CLIP and CLIP$_e$ scores of $27.87$ and $25.48$, respectively. 
Compared with Pyramid-Edit, the strongest method for text alignment, VM-Edit improves CLIP by $0.19$ and CLIP$_e$ by $0.49$. 
This suggests that the improved preservation ability of VM-Edit does not come at the cost of prompt adherence. 
For image quality assessment, VM-Edit obtains the lowest NIQE score of $9.04$, outperforming the next-best method, Pyramid-Edit, by $5.6\%$.
The main exception is the motion metric, where TokenFlow achieves the highest score of $62.70$, while VM-Edit obtains $43.69$. 
This gap suggests that VM-Edit is more conservative in preserving the source video structure and background, which may reduce the amount of apparent motion measured by conventional motion metrics. 
In terms of speed, VM-Edit runs at $2.34$ FPS. 
Although it is not the fastest method, it is faster than Wan-Edit and Pyramid-Edit, while providing substantially better preservation and perceptual quality.

\begin{table}[!htbp]
    \centering
    \fontsize{6.8}{7.2}\selectfont
    \setlength{\tabcolsep}{2.4pt}
    \renewcommand{\arraystretch}{0.96}
    \caption{
    Conventional metric comparison on NRVBench. 
    }
    \label{appendix_table:benchmark_results}
    \resizebox{\linewidth}{!}{
    \begin{tabular}{@{}lcccccccccc@{}}
        \toprule
        \multirow{2}{*}{Method}
        & \multicolumn{1}{c}{Struct.}
        & \multicolumn{4}{c}{Background Preservation}
        & \multicolumn{2}{c}{Text Align.}
        & \multicolumn{1}{c}{IQA}
        & \multicolumn{1}{c}{Motion}
        & \multicolumn{1}{c}{Speed} \\
        \cmidrule(lr){2-2}
        \cmidrule(lr){3-6}
        \cmidrule(lr){7-8}
        \cmidrule(lr){9-9}
        \cmidrule(lr){10-10}
        \cmidrule(lr){11-11}
        & Dist.$\downarrow$
        & PSNR$\uparrow$
        & LPIPS$\downarrow$
        & MSE$\downarrow$
        & SSIM$\uparrow$
        & CLIP$\uparrow$
        & CLIP$_e\uparrow$
        & NIQE$\downarrow$
        & Mot.$\uparrow$
        & FPS$\uparrow$ \\
        \midrule

        \multicolumn{11}{c}{\textbf{Long-video subset} 
        $(15 \times 3 \times 150\ \mathrm{frames})$} \\
        \midrule
        Wan-Edit~\citep{Li2025fivebench}      
        & 18.04 & 32.36 & 96.76 & 7.99 & 95.66 
        & 27.52 & 24.41 & 10.70 & 60.06 & 2.03 \\
        Pyramid-Edit~\citep{Li2025fivebench}  
        & 84.06 & 22.41 & 128.14 & 77.72 & 87.78 
        & 27.68 & 24.99 & 9.58 & 57.95 & 1.47 \\
        TokenFlow~\citep{geyer2024tokenflow}     
        & 50.52 & 24.19 & 98.68 & 53.96 & 88.38 
        & 26.83 & 24.69 & 9.96 & \textbf{62.70} & 2.96 \\
        VidToMe~\citep{li2024vidtome}       
        & 231.82 & 13.80 & 214.59 & 491.33 & 75.65 
        & 25.00 & 22.85 & 10.37 & 55.78 & \textbf{4.59} \\
        AnyV2V~\citep{ku2024anyv2v}
        & 182.00 & 13.61 & 354.90 & 786.42 & 72.44 
        & 23.31 & 21.97 & 11.63 & 54.63 & 3.55 \\
        \textbf{VM-Edit}
        & \textbf{16.45} & \textbf{37.35} & \textbf{58.91} & \textbf{2.99} & \textbf{97.36}
        & \textbf{27.87} & \textbf{25.48} & \textbf{9.04} & 43.69 & 2.34 \\
        \midrule

        \multicolumn{11}{c}{\textbf{Standard NRVBench} 
        $(180 \times 60\ \mathrm{frames})$} \\
        \midrule
        Reference source 
        & 0 & $\infty$ & 0 & 0 & 100 
        & 26.66 & 23.83 & 7.37 & 90.83 & / \\
        Wan-Edit~\citep{Li2025fivebench}      
        & 17.66 & 29.37 & 111.92 & 17.35 & 92.15 
        & 26.63 & 23.24 & 8.28 & 60.65 & 2.67 \\
        Pyramid-Edit~\citep{Li2025fivebench}   
        & 83.30 & 21.73 & 147.40 & 95.24 & 84.50 
        & \textbf{26.65} & 22.47 & 8.26 & 53.89 & 2.78 \\
        TokenFlow~\citep{geyer2024tokenflow}     
        & 111.93 & 21.10 & 134.31 & 118.23 & 75.30 
        & 26.47 & \textbf{23.28} & 7.98 & 58.49 & 3.11 \\
        VidToMe~\citep{li2024vidtome}       
        & 364.93 & 11.80 & 300.64 & 778.40 & 59.78 
        & 26.60 & 22.89 & 8.34 & 57.76 & \textbf{3.89} \\
        AnyV2V~\citep{ku2024anyv2v}        
        & 329.34 & 13.83 & 353.94 & 646.59 & 58.48 
        & 23.32 & 20.75 & 9.47 & 49.14 & 2.76 \\
        \textbf{VM-Edit}
        & \textbf{8.69} & \textbf{35.79} & \textbf{47.89} & \textbf{4.74} & \textbf{95.72}
        & 26.15 & 23.19 & \textbf{7.36} & \textbf{60.94} & 2.21 \\
        \bottomrule
    \end{tabular}
    }
\end{table}

\paragraph{Summary.}
The results in Table~\ref{appendix_table:benchmark_results} show that VM-Edit provides the strongest overall balance among structure preservation, background fidelity, perceptual quality, and text alignment. 
Its most significant gains appear in preservation-oriented metrics, including Dist., PSNR, LPIPS, MSE, and SSIM, where VM-Edit consistently outperforms prior methods across both long-video and standard settings. 
Although VM-Edit is not always the fastest method and shows a lower motion score on the long-video subset, its strong performance on the majority of quality metrics highlights its effectiveness for high-fidelity video editing, especially when preserving the source content is critical. The per-edit results reported in Tables~\ref{appendix_table:benchmark_results_edit1}--\ref{appendix_table:benchmark_results_edit6} further confirm that VM-Edit consistently delivers strong structural fidelity and background preservation across diverse editing instructions, while maintaining competitive text alignment, motion quality, and inference speed.

\begin{figure}[htbp]
  \centering
  \includegraphics[width=\columnwidth]{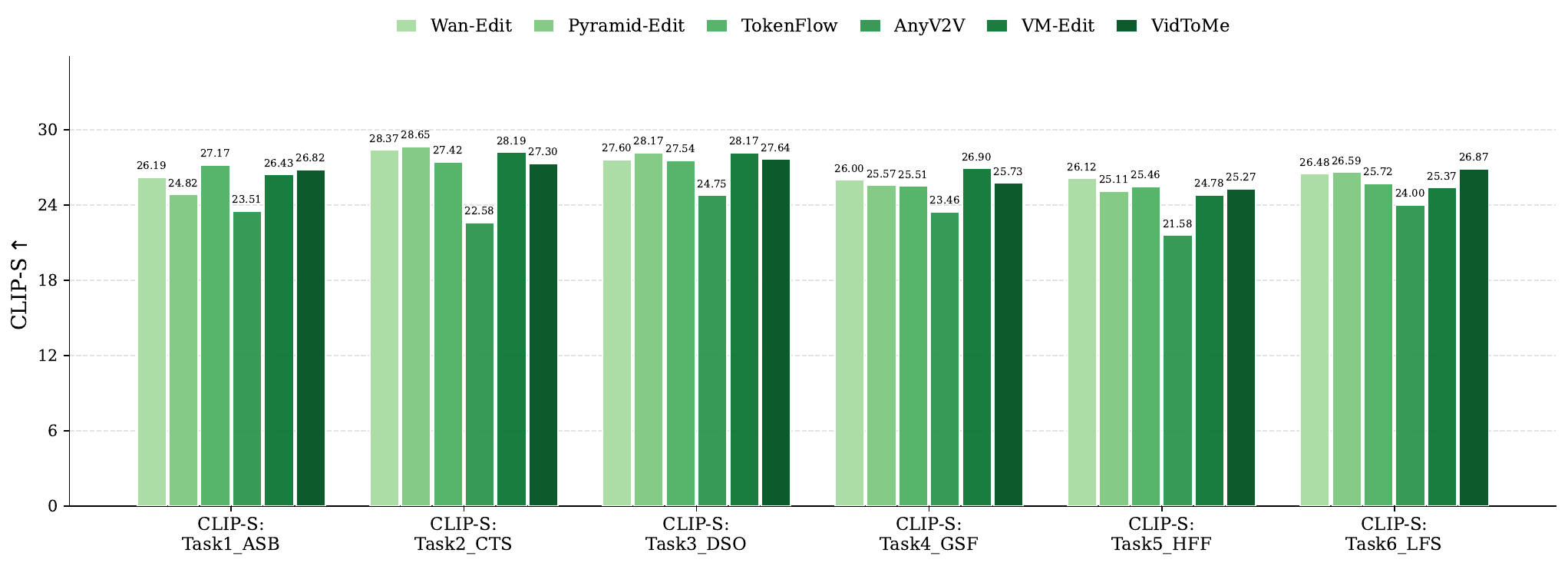}
  \caption{The text alignment CLIP-S data for the six models.}
  \label{appendix_fig:CLIP-S}
\end{figure}

\begin{figure}[htbp]
  \centering
  \includegraphics[width=\columnwidth]{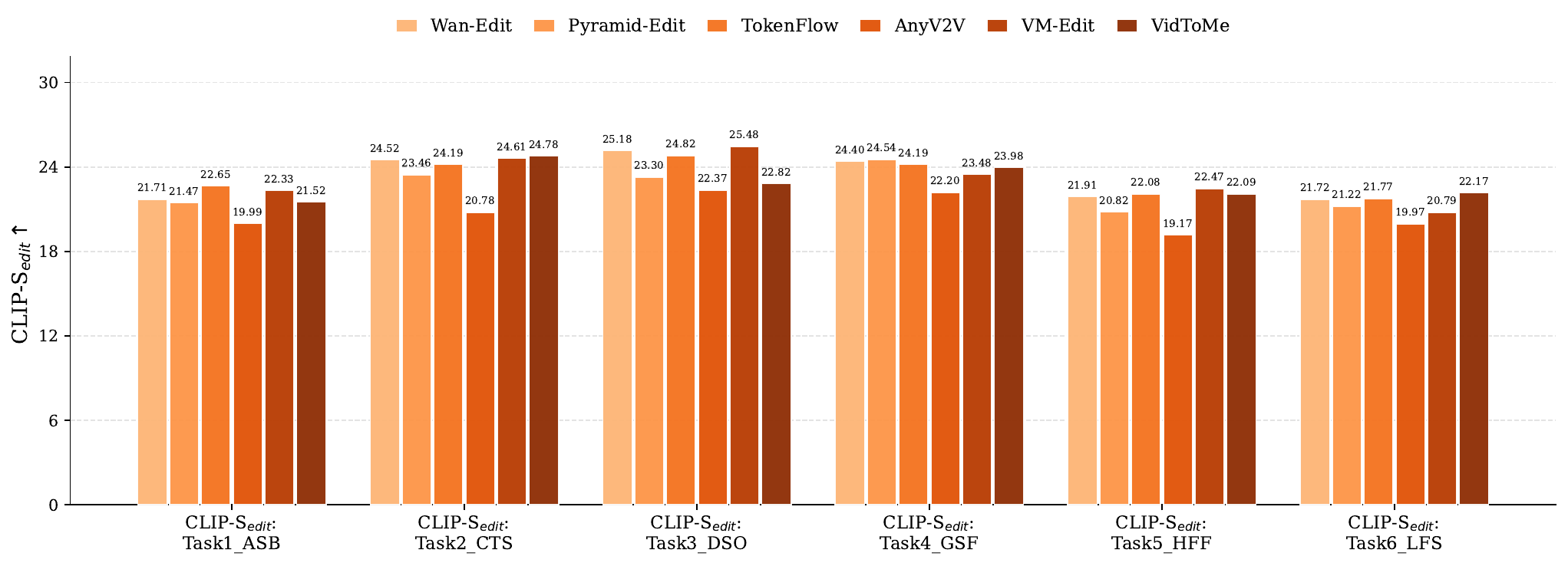}
  \caption{The text alignment CLIP-S-edit data for the six models.}
  \label{appendix_fig:CLIP-Se}
\end{figure}

\begin{figure}[htbp]
  \centering
  \includegraphics[width=\columnwidth]{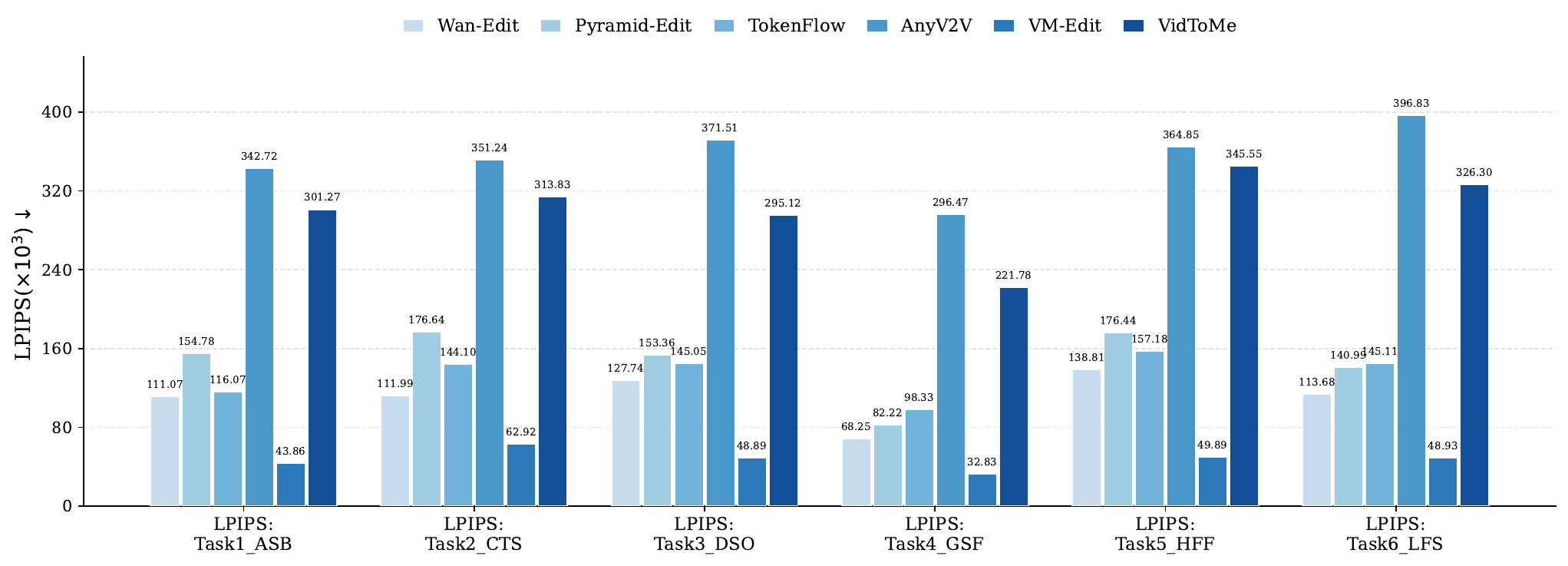}
  \caption{The LPIPS background preservation for the six models.}
  \label{appendix_fig:LPIPS}
\end{figure}

\begin{figure}[htbp]
  \centering
  \includegraphics[width=\columnwidth]{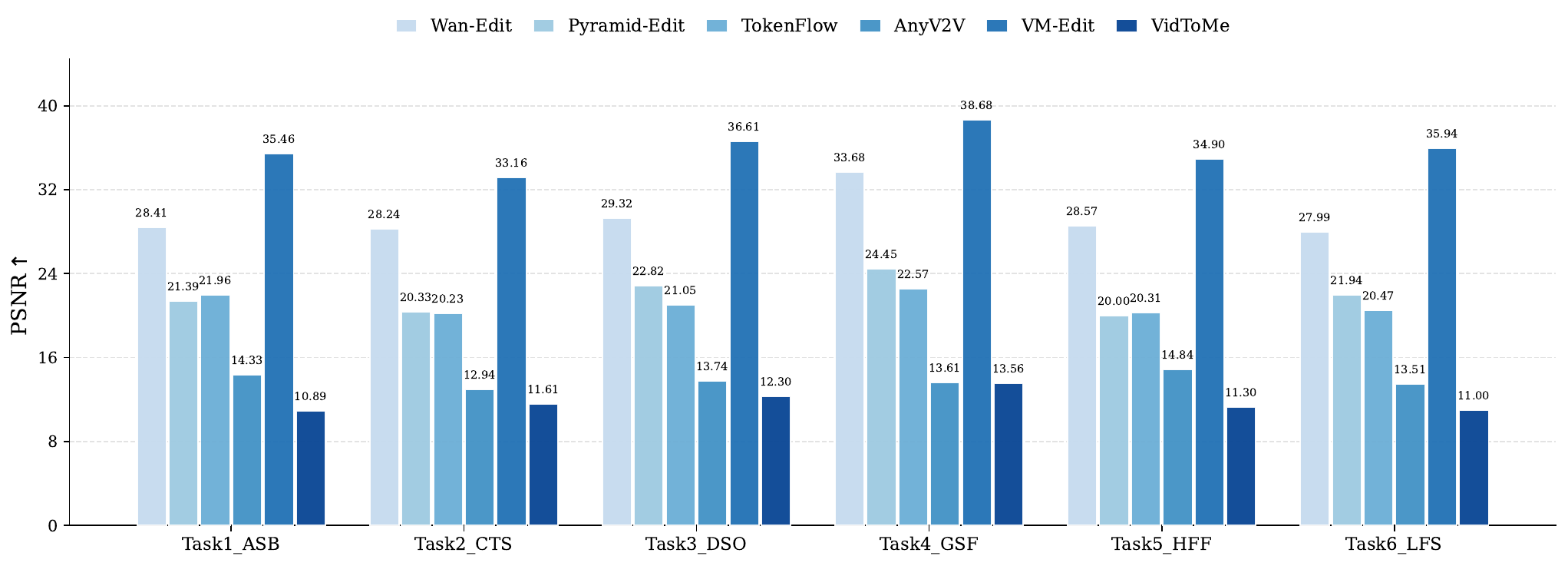}
  \caption{The PSNR background preservation for the six models.}
  \label{appendix_fig:PSNR}
\end{figure}

\begin{figure}[htbp]
  \centering
  \includegraphics[width=\columnwidth]{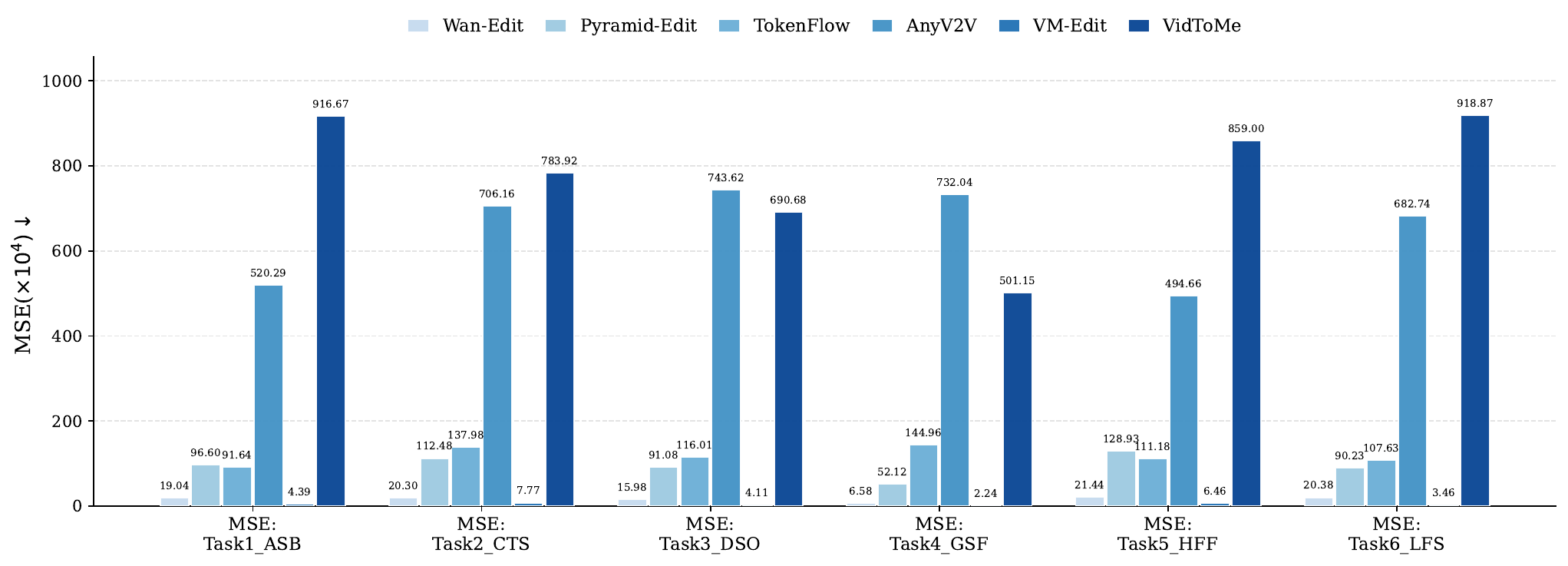}
  \caption{The MSE background preservation for the six models.}
  \label{appendix_fig:MSE}
\end{figure}

\begin{figure}[htbp]
  \centering
  \includegraphics[width=\columnwidth]{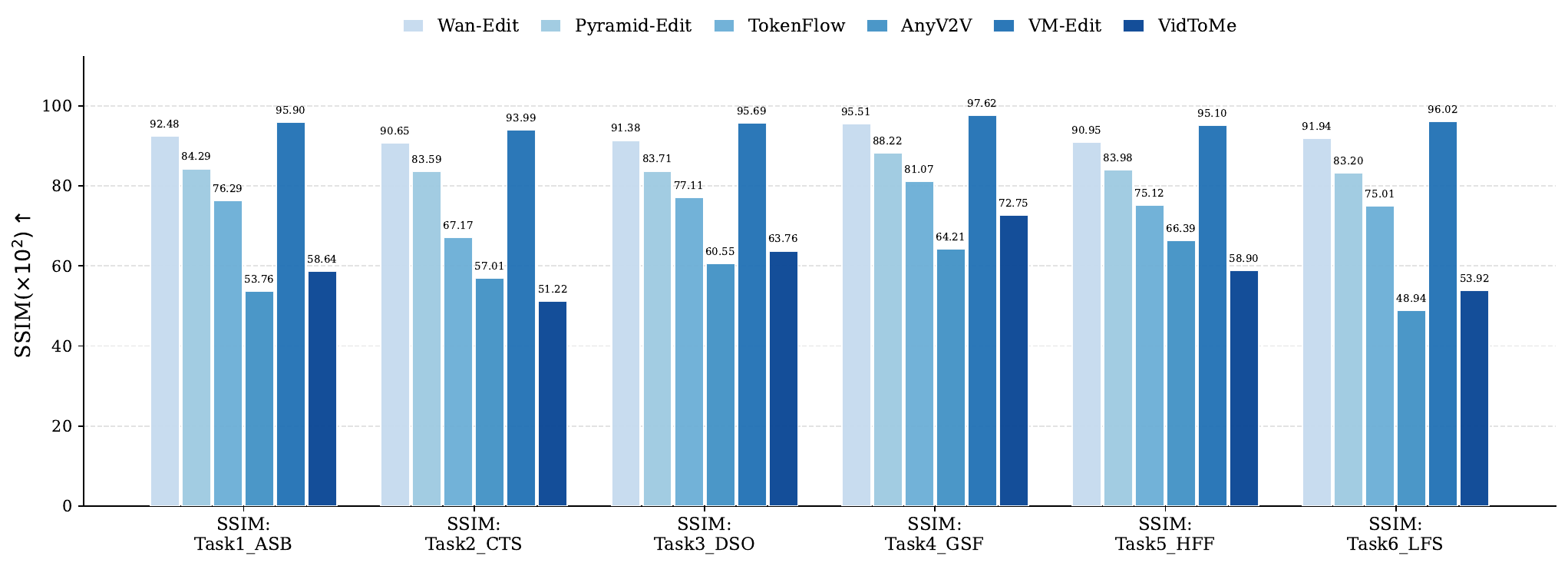}
  \caption{The SSIM background preservation for the six models.}
  \label{appendix_fig:SSIM}
\end{figure}

\begin{figure}[htbp]
  \centering
  \includegraphics[width=\columnwidth]{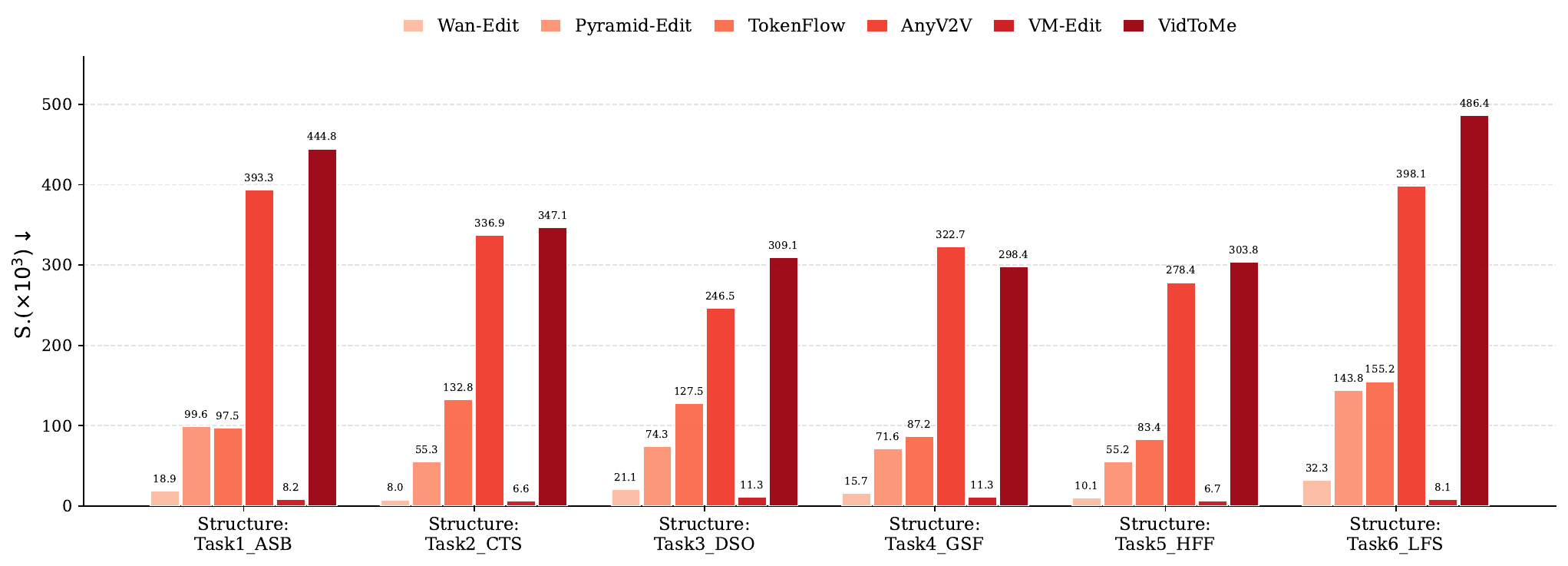}
  \caption{The Structure Distance for the six models.}
  \label{appendix_fig:structure_distance}
\end{figure}

\paragraph{Significance analysis.}
Since the table reports aggregate scores rather than per-sample distributions, we focus on consistency-based and practical significance rather than claiming significance from a specific hypothesis test. 
Across the six edit-specific tables, VM-Edit achieves the best background-preservation scores for PSNR, LPIPS, MSE, and SSIM in every edit setting on both the long-video subset and the standard NRVBench benchmark. 
This yields consistent wins across $6 \times 2$ evaluation settings for all four preservation metrics, demonstrating that the advantage of VM-Edit is stable across editing instructions and video lengths. 
Moreover, VM-Edit obtains the best structural distance in $11$ out of $12$ per-edit settings, showing that its improvements are not only concentrated in pixel-level preservation but also extend to geometric and structural consistency. 
In contrast, the gains in text-alignment, motion, and speed metrics are more mixed, which suggests a trade-off between aggressive editing dynamics and faithful preservation. 
Nevertheless, the large and repeated margins on preservation-oriented metrics provide strong evidence that VM-Edit delivers practically significant improvements for high-fidelity video editing. In the main paper, we report the Motion Fidelity results across the six non-rigid motion categories to highlight the temporal and dynamic consistency of different methods. 
For completeness, we additionally provide detailed results for the remaining conventional metrics in Appendix. 
Specifically, text-alignment metrics are shown in Figures~\ref{appendix_fig:CLIP-S}--\ref{appendix_fig:CLIP-Se}, background-preservation metrics are shown in Figures~\ref{appendix_fig:LPIPS}--\ref{appendix_fig:SSIM}, and structural consistency is shown in Figure~\ref{appendix_fig:structure_distance}. 
These supplementary figures provide a more comprehensive view of model performance beyond Motion Fidelity.

\begin{table}[!htbp]
    \centering
    \fontsize{6.8}{7.2}\selectfont
    \setlength{\tabcolsep}{2.4pt}
    \renewcommand{\arraystretch}{1.12}
    \caption{
    Conventional metric comparison on NRVBench for Edit 1.
    }
    \label{appendix_table:benchmark_results_edit1}
    \resizebox{\linewidth}{!}{
    \begin{tabular}{@{}lcccccccccc@{}}
        \toprule
        \multirow{2}{*}{Method}
        & \multicolumn{1}{c}{Struct.}
        & \multicolumn{4}{c}{Background Preservation}
        & \multicolumn{2}{c}{Text Align.}
        & \multicolumn{1}{c}{IQA}
        & \multicolumn{1}{c}{Motion}
        & \multicolumn{1}{c}{Speed} \\
        \cmidrule(lr){2-2}
        \cmidrule(lr){3-6}
        \cmidrule(lr){7-8}
        \cmidrule(lr){9-9}
        \cmidrule(lr){10-10}
        \cmidrule(lr){11-11}
        & Dist.$\downarrow$
        & PSNR$\uparrow$
        & LPIPS$\downarrow$
        & MSE$\downarrow$
        & SSIM$\uparrow$
        & CLIP$\uparrow$
        & CLIP$_e\uparrow$
        & NIQE$\downarrow$
        & Mot.$\uparrow$
        & FPS$\uparrow$ \\
        \midrule

        \multicolumn{11}{c}{\textbf{Long-video subset} 
        $(15 \times 3 \times 150\ \mathrm{frames})$} \\
        \midrule
        Wan-Edit~\citep{Li2025fivebench}      
        & 8.89 & 30.61 & 146.37 & 10.74 & 96.41 & 28.24 & 22.28 & 10.76 & 69.12 & 2.03 \\
        Pyramid-Edit~\citep{Li2025fivebench}      
        & 39.21 & 20.54 & 174.66 & 103.63 & 92.23 & 26.93 & 21.99 & 9.15 & 65.65 & 1.47 \\
        TokenFlow~\citep{geyer2024tokenflow}      
        & 12.26 & 24.87 & 124.88 & 36.64 & 95.37 & \textbf{31.30} & \textbf{26.11} & 9.21 & \textbf{72.51} & 2.96 \\
        VidToMe~\citep{li2024vidtome}      
        & 105.94 & 13.83 & 261.56 & 504.89 & 79.97 & 28.15 & 23.45 & 9.57 & 68.70 & \textbf{4.59} \\
        AnyV2V~\citep{ku2024anyv2v}      
        & 85.62 & 15.36 & 386.01 & 460.46 & 86.44 & 24.77 & 20.74 & 11.96 & 59.52 & 3.55 \\
        \textbf{VM-Edit}      
        & \textbf{6.31} & \textbf{38.01} & \textbf{82.53} & \textbf{2.11} & \textbf{98.02} & 30.87 & 25.67 & \textbf{8.84} & 57.41 & 2.34 \\
        \midrule

        \multicolumn{11}{c}{\textbf{Standard NRVBench} 
        $(180 \times 60\ \mathrm{frames})$} \\
        \midrule
        Reference source 
        & 0 & $\infty$ & 0 & 0 & 100 & 26.27 & 22.34 & 7.59 & 92.89 & / \\
        Wan-Edit~\citep{Li2025fivebench}      
        & 18.87 & 28.41 & 111.07 & 19.40 & 92.48 & 25.20 & 21.71 & 8.05 & 63.62 & 2.67 \\
        Pyramid-Edit~\citep{Li2025fivebench}      
        & 99.61 & 21.39 & 154.78 & 96.60 & 84.29 & 24.82 & 21.47 & 7.86 & 56.77 & 2.78 \\
        TokenFlow~\citep{geyer2024tokenflow}      
        & 97.53 & 21.96 & 116.07 & 91.64 & 76.29 & \textbf{27.17} & \textbf{22.65} & 8.32 & 60.43 & 3.11 \\
        VidToMe~\citep{li2024vidtome}      
        & 444.78 & 10.89 & 301.29 & 916.67 & 58.64 & 26.82 & 21.52 & \textbf{7.74} & 64.50 & \textbf{3.89} \\
        AnyV2V~\citep{ku2024anyv2v}      
        & 393.33 & 14.33 & 342.72 & 520.29 & 53.76 & 23.51 & 19.99 & 9.38 & 47.30 & 2.76 \\
        \textbf{VM-Edit}      
        & \textbf{8.22} & \textbf{35.46} & \textbf{43.86} & \textbf{4.39} & \textbf{95.90} & 26.43 & 22.32 & 8.32 & \textbf{65.47} & 2.21 \\
        \bottomrule
    \end{tabular}
    }
\end{table}

\begin{table}[!htbp]
    \centering
    \fontsize{6.8}{7.2}\selectfont
    \setlength{\tabcolsep}{2.4pt}
    \renewcommand{\arraystretch}{1.12}
    \caption{
    Conventional metric comparison on NRVBench for Edit 2.
    }
    \label{appendix_table:benchmark_results_edit2}
    \resizebox{\linewidth}{!}{
    \begin{tabular}{@{}lcccccccccc@{}}
        \toprule
        \multirow{2}{*}{Method}
        & \multicolumn{1}{c}{Struct.}
        & \multicolumn{4}{c}{Background Preservation}
        & \multicolumn{2}{c}{Text Align.}
        & \multicolumn{1}{c}{IQA}
        & \multicolumn{1}{c}{Motion}
        & \multicolumn{1}{c}{Speed} \\
        \cmidrule(lr){2-2}
        \cmidrule(lr){3-6}
        \cmidrule(lr){7-8}
        \cmidrule(lr){9-9}
        \cmidrule(lr){10-10}
        \cmidrule(lr){11-11}
        & Dist.$\downarrow$
        & PSNR$\uparrow$
        & LPIPS$\downarrow$
        & MSE$\downarrow$
        & SSIM$\uparrow$
        & CLIP$\uparrow$
        & CLIP$_e\uparrow$
        & NIQE$\downarrow$
        & Mot.$\uparrow$
        & FPS$\uparrow$ \\
        \midrule

        \multicolumn{11}{c}{\textbf{Long-video subset} 
        $(15 \times 3 \times 150\ \mathrm{frames})$} \\
        \midrule
        Wan-Edit~\citep{Li2025fivebench}      
        & 22.61 & 35.03 & 58.47 & 4.17 & 95.28 & 28.66 & 25.62 & 12.31 & 51.30 & 2.03 \\
        Pyramid-Edit~\citep{Li2025fivebench}      
        & 88.89 & 24.99 & 61.03 & 41.05 & 88.26 & \textbf{29.77} & 27.09 & 10.69 & 48.11 & 1.47 \\
        TokenFlow~\citep{geyer2024tokenflow}      
        & 28.77 & 25.47 & 69.01 & 36.70 & 91.87 & 28.65 & 25.43 & 12.22 & 53.40 & 2.96 \\
        VidToMe~\citep{li2024vidtome}      
        & 325.03 & 13.82 & 154.51 & 486.18 & 77.16 & 25.26 & 24.79 & 11.45 & 45.37 & \textbf{4.59} \\
        AnyV2V~\citep{ku2024anyv2v}      
        & 261.47 & 11.83 & 346.17 & 1375.32 & 61.78 & 23.84 & 23.23 & 12.50 & \textbf{54.72} & 3.55 \\
        \textbf{VM-Edit}      
        & \textbf{18.73} & \textbf{37.61} & \textbf{46.09} & \textbf{2.57} & \textbf{97.58} & 29.60 & \textbf{27.22} & \textbf{10.14} & 31.81 & 2.34 \\
        \midrule

        \multicolumn{11}{c}{\textbf{Standard NRVBench} 
        $(180 \times 60\ \mathrm{frames})$} \\
        \midrule
        Reference source 
        & 0 & $\infty$ & 0 & 0 & 100 & 27.53 & 24.20 & 7.08 & 90.66 & / \\
        Wan-Edit~\citep{Li2025fivebench}      
        & 7.98 & 28.24 & 111.99 & 20.30 & 90.65 & 28.38 & 24.52 & 7.80 & 57.84 & 2.67 \\
        Pyramid-Edit~\citep{Li2025fivebench}      
        & 55.28 & 20.33 & 176.64 & 112.48 & 83.59 & \textbf{28.65} & 23.46 & 7.84 & 54.07 & 2.78 \\
        TokenFlow~\citep{geyer2024tokenflow}      
        & 132.78 & 20.23 & 144.10 & 137.98 & 67.17 & 27.42 & 24.19 & \textbf{7.30} & 53.04 & 3.11 \\
        VidToMe~\citep{li2024vidtome}      
        & 347.05 & 11.61 & 313.86 & 783.92 & 51.22 & 27.29 & \textbf{24.78} & 7.35 & 53.30 & \textbf{3.89} \\
        AnyV2V~\citep{ku2024anyv2v}      
        & 336.93 & 12.94 & 351.24 & 706.16 & 57.01 & 22.58 & 20.78 & 8.61 & 45.17 & 2.76 \\
        \textbf{VM-Edit}      
        & \textbf{6.56} & \textbf{33.16} & \textbf{62.92} & \textbf{7.77} & \textbf{93.99} & 28.19 & 24.61 & 7.62 & \textbf{58.86} & 2.21 \\
        \bottomrule
    \end{tabular}
    }
\end{table}

\clearpage

\begin{table}[!htbp]
    \centering
    \fontsize{6.8}{7.2}\selectfont
    \setlength{\tabcolsep}{2.4pt}
    \renewcommand{\arraystretch}{1.12}
    \caption{
    Conventional metric comparison on NRVBench for Edit 3.
    }
    \label{appendix_table:benchmark_results_edit3}
    \resizebox{\linewidth}{!}{
    \begin{tabular}{@{}lcccccccccc@{}}
        \toprule
        \multirow{2}{*}{Method}
        & \multicolumn{1}{c}{Struct.}
        & \multicolumn{4}{c}{Background Preservation}
        & \multicolumn{2}{c}{Text Align.}
        & \multicolumn{1}{c}{IQA}
        & \multicolumn{1}{c}{Motion}
        & \multicolumn{1}{c}{Speed} \\
        \cmidrule(lr){2-2}
        \cmidrule(lr){3-6}
        \cmidrule(lr){7-8}
        \cmidrule(lr){9-9}
        \cmidrule(lr){10-10}
        \cmidrule(lr){11-11}
        & Dist.$\downarrow$
        & PSNR$\uparrow$
        & LPIPS$\downarrow$
        & MSE$\downarrow$
        & SSIM$\uparrow$
        & CLIP$\uparrow$
        & CLIP$_e\uparrow$
        & NIQE$\downarrow$
        & Mot.$\uparrow$
        & FPS$\uparrow$ \\
        \midrule

        \multicolumn{11}{c}{\textbf{Long-video subset} 
        $(15 \times 3 \times 150\ \mathrm{frames})$} \\
        \midrule
        Wan-Edit~\citep{Li2025fivebench}      
        & \textbf{21.52} & 31.66 & 85.90 & 9.45 & 94.92 & 27.50 & 26.06 & 8.90 & 59.27 & 2.03 \\
        Pyramid-Edit~\citep{Li2025fivebench}      
        & 102.28 & 21.96 & 130.00 & 75.94 & 86.19 & \textbf{28.64} & 25.98 & 8.68 & \textbf{62.56} & 1.47 \\
        TokenFlow~\citep{geyer2024tokenflow}      
        & 103.38 & 22.81 & 100.53 & 76.14 & 78.69 & 26.10 & 25.50 & 8.36 & 61.25 & 2.96 \\
        VidToMe~\citep{li2024vidtome}      
        & 216.59 & 13.96 & 214.01 & 456.33 & 71.71 & 24.12 & 22.40 & 10.30 & 53.96 & \textbf{4.59} \\
        AnyV2V~\citep{ku2024anyv2v}      
        & 150.63 & 16.07 & 289.82 & 647.30 & 68.10 & 23.68 & 21.26 & 9.51 & 55.59 & 3.55 \\
        \textbf{VM-Edit}      
        & 24.12 & \textbf{36.46} & \textbf{48.15} & \textbf{4.27} & \textbf{96.48} & 27.81 & \textbf{26.55} & \textbf{8.12} & 41.34 & 2.34 \\
        \midrule

        \multicolumn{11}{c}{\textbf{Standard NRVBench} 
        $(180 \times 60\ \mathrm{frames})$} \\
        \midrule
        Reference source 
        & 0 & $\infty$ & 0 & 0 & 100 & 28.17 & 26.62 & 7.37 & 91.02 & / \\
        Wan-Edit~\citep{Li2025fivebench}      
        & 21.07 & 29.32 & 127.74 & 15.98 & 91.38 & 27.60 & 25.18 & 8.67 & \textbf{67.65} & 2.67 \\
        Pyramid-Edit~\citep{Li2025fivebench}      
        & 74.32 & 22.28 & 153.36 & 91.08 & 83.71 & \textbf{28.17} & 23.30 & 8.72 & 56.07 & 2.78 \\
        TokenFlow~\citep{geyer2024tokenflow}      
        & 127.49 & 21.05 & 145.05 & 116.01 & 77.11 & 27.54 & 24.82 & \textbf{3.48} & 64.39 & 3.11 \\
        VidToMe~\citep{li2024vidtome}      
        & 309.10 & 12.30 & 295.14 & 690.68 & 63.76 & 27.64 & 22.82 & 9.16 & 63.23 & \textbf{3.89} \\
        AnyV2V~\citep{ku2024anyv2v}      
        & 246.52 & 13.74 & 371.51 & 743.62 & 60.55 & 24.76 & 22.37 & 10.00 & 55.56 & 2.76 \\
        \textbf{VM-Edit}      
        & \textbf{11.28} & \textbf{36.61} & \textbf{48.89} & \textbf{4.11} & \textbf{95.69} & 26.90 & \textbf{25.48} & 9.04 & 66.74 & 2.21 \\
        \bottomrule
    \end{tabular}
    }
\end{table}

\begin{table}[!htbp]
    \centering
    \fontsize{6.8}{7.2}\selectfont
    \setlength{\tabcolsep}{2.4pt}
    \renewcommand{\arraystretch}{1.12}
    \caption{
    Conventional metric comparison on NRVBench for Edit 4.
    }
    \label{appendix_table:benchmark_results_edit4}
    \resizebox{\linewidth}{!}{
    \begin{tabular}{@{}lcccccccccc@{}}
        \toprule
        \multirow{2}{*}{Method}
        & \multicolumn{1}{c}{Struct.}
        & \multicolumn{4}{c}{Background Preservation}
        & \multicolumn{2}{c}{Text Align.}
        & \multicolumn{1}{c}{IQA}
        & \multicolumn{1}{c}{Motion}
        & \multicolumn{1}{c}{Speed} \\
        \cmidrule(lr){2-2}
        \cmidrule(lr){3-6}
        \cmidrule(lr){7-8}
        \cmidrule(lr){9-9}
        \cmidrule(lr){10-10}
        \cmidrule(lr){11-11}
        & Dist.$\downarrow$
        & PSNR$\uparrow$
        & LPIPS$\downarrow$
        & MSE$\downarrow$
        & SSIM$\uparrow$
        & CLIP$\uparrow$
        & CLIP$_e\uparrow$
        & NIQE$\downarrow$
        & Mot.$\uparrow$
        & FPS$\uparrow$ \\
        \midrule

        \multicolumn{11}{c}{\textbf{Long-video subset} 
        $(15 \times 3 \times 150\ \mathrm{frames})$} \\
        \midrule
        Wan-Edit~\citep{Li2025fivebench}      
        & 8.53 & 30.82 & 144.85 & 10.24 & 96.48 & \textbf{27.94} & 22.56 & 10.78 & 70.10 & 2.03 \\
        Pyramid-Edit~\citep{Li2025fivebench}      
        & 48.21 & 19.93 & 183.59 & 124.47 & 91.99 & 26.43 & 23.71 & 9.22 & 62.70 & 1.47 \\
        TokenFlow~\citep{geyer2024tokenflow}      
        & 13.71 & 24.68 & 125.25 & 38.41 & 95.08 & 27.39 & \textbf{24.14} & 9.08 & \textbf{72.00} & 2.96 \\
        VidToMe~\citep{li2024vidtome}      
        & 165.43 & 13.19 & 275.92 & 555.32 & 77.62 & 26.22 & 22.17 & 9.78 & 68.27 & \textbf{4.59} \\
        AnyV2V~\citep{ku2024anyv2v}      
        & 113.80 & 14.13 & 401.80 & 559.53 & 86.81 & 20.67 & 21.28 & 13.36 & 52.01 & 3.55 \\
        \textbf{VM-Edit}      
        & \textbf{6.43} & \textbf{38.15} & \textbf{82.44} & \textbf{2.07} & \textbf{98.02} & 27.68 & 23.59 & \textbf{8.86} & 58.03 & 2.34 \\
        \midrule

        \multicolumn{11}{c}{\textbf{Standard NRVBench} 
        $(180 \times 60\ \mathrm{frames})$} \\
        \midrule
        Reference source 
        & 0 & $\infty$ & 0 & 0 & 100 & 25.74 & 24.80 & 7.82 & 89.60 & / \\
        Wan-Edit~\citep{Li2025fivebench}      
        & 15.71 & 33.68 & 68.25 & 6.58 & 95.51 & 26.00 & 24.40 & 8.01 & \textbf{64.38} & 2.67 \\
        Pyramid-Edit~\citep{Li2025fivebench}      
        & 71.59 & 24.45 & 82.22 & 52.12 & 88.22 & \textbf{26.57} & \textbf{24.54} & 8.66 & 57.76 & 2.78 \\
        TokenFlow~\citep{geyer2024tokenflow}      
        & 87.22 & 22.57 & 98.33 & 144.96 & 81.07 & 25.51 & 24.19 & \textbf{0.00} & 64.26 & 3.11 \\
        VidToMe~\citep{li2024vidtome}      
        & 298.47 & 13.56 & 221.80 & 501.15 & 72.25 & 25.73 & 23.98 & 9.30 & 60.13 & \textbf{3.89} \\
        AnyV2V~\citep{ku2024anyv2v}      
        & 322.69 & 13.61 & 296.47 & 732.04 & 64.21 & 23.46 & 22.20 & 9.94 & 58.06 & 2.76 \\
        \textbf{VM-Edit}      
        & \textbf{11.32} & \textbf{38.68} & \textbf{32.83} & \textbf{2.24} & \textbf{97.62} & 24.78 & 23.48 & 8.84 & 62.59 & 2.21 \\
        \bottomrule
    \end{tabular}
    }
\end{table}

\clearpage
\begin{table}[!htbp]
    \centering
    \fontsize{6.8}{7.2}\selectfont
    \setlength{\tabcolsep}{2.4pt}
    \renewcommand{\arraystretch}{1.12}
    \caption{
    Conventional metric comparison on NRVBench for Edit 5.
    }
    \label{appendix_table:benchmark_results_edit5}
    \resizebox{\linewidth}{!}{
    \begin{tabular}{@{}lcccccccccc@{}}
        \toprule
        \multirow{2}{*}{Method}
        & \multicolumn{1}{c}{Struct.}
        & \multicolumn{4}{c}{Background Preservation}
        & \multicolumn{2}{c}{Text Align.}
        & \multicolumn{1}{c}{IQA}
        & \multicolumn{1}{c}{Motion}
        & \multicolumn{1}{c}{Speed} \\
        \cmidrule(lr){2-2}
        \cmidrule(lr){3-6}
        \cmidrule(lr){7-8}
        \cmidrule(lr){9-9}
        \cmidrule(lr){10-10}
        \cmidrule(lr){11-11}
        & Dist.$\downarrow$
        & PSNR$\uparrow$
        & LPIPS$\downarrow$
        & MSE$\downarrow$
        & SSIM$\uparrow$
        & CLIP$\uparrow$
        & CLIP$_e\uparrow$
        & NIQE$\downarrow$
        & Mot.$\uparrow$
        & FPS$\uparrow$ \\
        \midrule

        \multicolumn{11}{c}{\textbf{Long-video subset} 
        $(15 \times 3 \times 150\ \mathrm{frames})$} \\
        \midrule
        Wan-Edit~\citep{Li2025fivebench}      
        & 19.00 & 35.59 & 58.57 & 3.59 & 95.64 & 27.66 & 24.59 & 12.33 & 49.19 & 2.03 \\
        Pyramid-Edit~\citep{Li2025fivebench}      
        & 115.45 & 24.65 & 72.86 & 43.76 & 83.22 & \textbf{28.34} & \textbf{26.48} & 10.65 & 49.82 & 1.47 \\
        TokenFlow~\citep{geyer2024tokenflow}      
        & 28.16 & 25.29 & 65.77 & 36.02 & 91.77 & 26.32 & 24.64 & 12.41 & \textbf{53.24} & 2.96 \\
        VidToMe~\citep{li2024vidtome}      
        & 337.01 & 13.75 & 159.21 & 489.05 & 76.62 & 24.60 & 24.08 & 11.18 & 45.11 & \textbf{4.59} \\
        AnyV2V~\citep{ku2024anyv2v}      
        & 275.29 & 11.46 & 359.75 & 1093.65 & 60.99 & 23.11 & 22.82 & 12.83 & 49.32 & 3.55 \\
        \textbf{VM-Edit}      
        & \textbf{18.95} & \textbf{37.28} & \textbf{46.21} & \textbf{2.95} & \textbf{97.56} & 27.37 & 25.72 & \textbf{10.26} & 33.08 & 2.34 \\
        \midrule

        \multicolumn{11}{c}{\textbf{Standard NRVBench} 
        $(180 \times 60\ \mathrm{frames})$} \\
        \midrule
        Reference source 
        & 0 & $\infty$ & 0 & 0 & 100 & 25.79 & 23.00 & 6.36 & 90.25 & / \\
        Wan-Edit~\citep{Li2025fivebench}      
        & 10.07 & 28.57 & 138.81 & 21.44 & 90.95 & \textbf{26.12} & 21.91 & 8.20 & \textbf{60.82} & 2.67 \\
        Pyramid-Edit~\citep{Li2025fivebench}      
        & 55.22 & 20.00 & 176.44 & 128.93 & 83.98 & 25.11 & 20.82 & 7.70 & 52.36 & 2.78 \\
        TokenFlow~\citep{geyer2024tokenflow}      
        & 83.39 & 20.31 & 157.18 & 111.18 & 75.12 & 25.46 & 22.08 & \textbf{0.00} & 57.63 & 3.11 \\
        VidToMe~\citep{li2024vidtome}      
        & 303.81 & 11.44 & 345.57 & 859.00 & 58.90 & 25.27 & 22.09 & 7.51 & 57.25 & \textbf{3.89} \\
        AnyV2V~\citep{ku2024anyv2v}      
        & 278.45 & 14.84 & 364.85 & 494.66 & 66.39 & 21.59 & 19.17 & 8.91 & 45.93 & 2.76 \\
        \textbf{VM-Edit}      
        & \textbf{6.69} & \textbf{34.90} & \textbf{49.89} & \textbf{6.46} & \textbf{95.10} & 25.38 & \textbf{22.47} & 7.64 & 60.75 & 2.21 \\
        \bottomrule
    \end{tabular}
    }
\end{table}

\begin{table}[!htbp]
    \centering
    \fontsize{6.8}{7.2}\selectfont
    \setlength{\tabcolsep}{2.4pt}
    \renewcommand{\arraystretch}{1.12}
    \caption{
    Conventional metric comparison on NRVBench for Edit 6.
    }
    \label{appendix_table:benchmark_results_edit6}
    \resizebox{\linewidth}{!}{
    \begin{tabular}{@{}lcccccccccc@{}}
        \toprule
        \multirow{2}{*}{Method}
        & \multicolumn{1}{c}{Struct.}
        & \multicolumn{4}{c}{Background Preservation}
        & \multicolumn{2}{c}{Text Align.}
        & \multicolumn{1}{c}{IQA}
        & \multicolumn{1}{c}{Motion}
        & \multicolumn{1}{c}{Speed} \\
        \cmidrule(lr){2-2}
        \cmidrule(lr){3-6}
        \cmidrule(lr){7-8}
        \cmidrule(lr){9-9}
        \cmidrule(lr){10-10}
        \cmidrule(lr){11-11}
        & Dist.$\downarrow$
        & PSNR$\uparrow$
        & LPIPS$\downarrow$
        & MSE$\downarrow$
        & SSIM$\uparrow$
        & CLIP$\uparrow$
        & CLIP$_e\uparrow$
        & NIQE$\downarrow$
        & Mot.$\uparrow$
        & FPS$\uparrow$ \\
        \midrule

        \multicolumn{11}{c}{\textbf{Long-video subset} 
        $(15 \times 3 \times 150\ \mathrm{frames})$} \\
        \midrule
        Wan-Edit~\citep{Li2025fivebench}      
        & 34.51 & 30.91 & 86.98 & 9.61 & 94.83 & 27.76 & \textbf{26.73} & 9.02 & 55.67 & 2.03 \\
        Pyramid-Edit~\citep{Li2025fivebench}      
        & 95.85 & 22.95 & 130.91 & 65.85 & 87.01 & \textbf{29.20} & 26.42 & 9.46 & 60.01 & 1.47 \\
        TokenFlow~\citep{geyer2024tokenflow}      
        & 107.30 & 22.71 & 101.05 & 84.82 & 79.01 & 27.26 & 26.08 & 8.86 & \textbf{60.75} & 2.96 \\
        VidToMe~\citep{li2024vidtome}      
        & 217.20 & 14.18 & 213.60 & 446.19 & 72.63 & 24.75 & 22.34 & 9.90 & 54.28 & \textbf{4.59} \\
        AnyV2V~\citep{ku2024anyv2v}      
        & 181.93 & 13.46 & 370.03 & 828.64 & 65.77 & 25.10 & 24.08 & 10.75 & 55.48 & 3.55 \\
        \textbf{VM-Edit}      
        & \textbf{23.95} & \textbf{36.48} & \textbf{47.95} & \textbf{4.13} & \textbf{96.47} & 27.49 & 26.68 & \textbf{8.10} & 40.51 & 2.34 \\
        \midrule

        \multicolumn{11}{c}{\textbf{Standard NRVBench} 
        $(180 \times 60\ \mathrm{frames})$} \\
        \midrule
        Reference source 
        & 0 & $\infty$ & 0 & 0 & 100 & 26.23 & 21.94 & 7.95 & 90.36 & / \\
        Wan-Edit~\citep{Li2025fivebench}      
        & 32.27 & 27.99 & 113.68 & 20.38 & 91.94 & 26.47 & 21.72 & 8.93 & 49.58 & 2.67 \\
        Pyramid-Edit~\citep{Li2025fivebench}      
        & 143.76 & 21.94 & 140.99 & 90.23 & 83.20 & 26.59 & 21.22 & 8.81 & 46.32 & 2.78 \\
        TokenFlow~\citep{geyer2024tokenflow}      
        & 155.19 & 20.47 & 145.11 & 107.63 & 75.01 & 25.73 & 21.77 & \textbf{0.00} & 51.22 & 3.11 \\
        VidToMe~\citep{li2024vidtome}      
        & 486.37 & 11.00 & 326.29 & 918.97 & 53.92 & \textbf{26.87} & \textbf{22.17} & 8.96 & 48.76 & \textbf{3.89} \\
        AnyV2V~\citep{ku2024anyv2v}      
        & 398.12 & 13.51 & 396.83 & 682.74 & 48.94 & 24.00 & 19.97 & 9.95 & 42.83 & 2.76 \\
        \textbf{VM-Edit}      
        & \textbf{8.09} & \textbf{35.94} & \textbf{48.93} & \textbf{3.46} & \textbf{96.02} & 25.26 & 20.79 & 9.15 & \textbf{51.25} & 2.21 \\
        \bottomrule
    \end{tabular}
    }
\end{table}

\clearpage
\subsection{NRVE-Acc Analysis}
\label{appendix_sec:nrve_acc_analysis}

Table~\ref{tab:nrve_multijudge_results} reports the quantitative comparison on NRVBench using NRVE-Acc. 
Different from conventional low-level metrics, NRVE-Acc evaluates video editing quality from three semantic dimensions: physical plausibility, temporal consistency, and instruction following. 
The final NRVE-Acc is computed per instance before averaging, and the cross-judge mean and standard deviation further reflect the robustness of each method under different VLM judge calibrations.

\paragraph{Component-level analysis.}
VM-Edit achieves the best physical plausibility score, reaching $S_{\mathrm{phy}}=71.31$, outperforming Wan-Edit, the strongest method on this component, by $1.44$ points. 
This suggests that VM-Edit produces edits that are more consistent with physical realism and scene-level visual plausibility. 
For temporal consistency, VM-Edit obtains $S_{\mathrm{temp}}=42.80$, ranking second among all methods and trailing Wan-Edit by $1.55$ points. 
This indicates that VM-Edit maintains competitive temporal coherence while emphasizing stronger structural and background preservation. 
For instruction following, VM-Edit obtains $S_{\mathrm{instr}}=62.57$, which is slightly higher than TokenFlow and AnyV2V, but lower than Pyramid-Edit and Wan-Edit. 
This result suggests that while VM-Edit remains competitive in following editing instructions, some baselines may produce more aggressive semantic modifications that are favored by instruction-alignment judges.

\paragraph{Judge-level analysis.}
Across the five VLM judges, VM-Edit shows consistently strong performance. 
It achieves a cross-judge mean NRVE-Acc of $30.89$, ranking second overall and only $0.62$ points lower than Wan-Edit. 
Compared with TokenFlow, Pyramid-Edit, VidToMe, and AnyV2V, VM-Edit improves the cross-judge mean by $0.99$, $2.46$, $2.11$, and $4.91$ points, respectively. 
Notably, VM-Edit ranks within the top three under all five judges, demonstrating that its advantage is not tied to a specific judge family or calibration preference. 
For example, VM-Edit obtains the second-best scores under Qwen-7B, Gemini, and Kimi, and remains highly competitive under Qwen-32B and Qwen-72B.

\paragraph{Cross-judge robustness.}
A key advantage of VM-Edit lies in its cross-judge stability. 
VM-Edit achieves the lowest standard deviation across judges, with $\mathrm{Std}=3.58$, compared with $4.06$ for Wan-Edit, $4.07$ for TokenFlow, $4.56$ for VidToMe, $6.29$ for AnyV2V, and $6.93$ for Pyramid-Edit. 
This corresponds to an $11.8\%$ reduction in judge sensitivity compared with Wan-Edit and a $48.3\%$ reduction compared with Pyramid-Edit. 
Therefore, although Wan-Edit obtains a slightly higher cross-judge mean, VM-Edit provides the most stable evaluation results across different VLM judges. 
This is important because VLM-based evaluation can be sensitive to judge calibration, and a lower cross-judge variance indicates more reliable and judge-agnostic editing quality.

\begin{table}[htbp]
    \centering
    \small
    \setlength{\tabcolsep}{3.2pt}
    \renewcommand{\arraystretch}{1.10}
    \caption{
    Quantitative comparison on NRVBench using NRVE-Acc.
    The final NRVE-Acc is computed per instance before averaging(not equal to the geometric mean of the displayed component averages).
    Mean and Std summarize cross-judge performance and sensitivity to judge calibration.
    }
    \label{appendix_table:nrve_multijudge_results}
    \resizebox{\linewidth}{!}{
    \begin{tabular}{lcccccccccc}
        \toprule
        \multirow{2}{*}{Method} 
        & \multicolumn{3}{c}{Component Avg.} 
        & \multicolumn{5}{c}{VLM Judges} 
        & \multicolumn{2}{c}{Cross-judge} \\
        \cmidrule(lr){2-4}
        \cmidrule(lr){5-9}
        \cmidrule(lr){10-11}
        & $S_{\mathrm{phy}}\uparrow$ 
        & $S_{\mathrm{temp}}\uparrow$ 
        & $S_{\mathrm{instr}}\uparrow$ 
        & Qwen-7B 
        & Qwen-32B 
        & Qwen-72B 
        & Gemini 
        & Kimi 
        & Mean$\uparrow$ 
        & Std$\downarrow$ \\
        \midrule
        TokenFlow~\citep{geyer2024tokenflow}     
        & 68.16 & 41.69 & 62.33 
        & 33.05 & 34.77 & 29.85 
        & 22.95 & 28.88 
        & 29.90 & 4.07 \\

        Pyramid-Edit~\citep{Li2025fivebench}  
        & 59.39 & 41.40 & 67.25 
        & 32.20 & 38.02 & 29.87 
        & 17.66 & 24.42 
        & 28.43 & 6.93 \\

        Wan-Edit~\citep{Li2025fivebench}      
        & 69.87 & 44.35 & 64.25 
        & 36.88 & 34.94 & 31.82 
        & 26.53 & 27.40 
        & 31.51 & 4.06 \\

        VidToMe~\citep{li2024vidtome}       
        & 69.64 & 38.98 & 61.46 
        & 30.24 & 34.03 & 31.41 
        & 20.66 & 27.56 
        & 28.78 & 4.56 \\

        AnyV2V~\citep{ku2024anyv2v}        
        & 59.24 & 39.28 & 62.05 
        & 30.35 & 33.88 & 26.00 
        & 15.33 & 24.33 
        & 25.98 & 6.29 \\

        \textbf{Ours} 
        & 71.31 & 42.80 & 62.57 
        & 34.71 & 34.91 & 30.53 
        & 25.55 & 28.76 
        & 30.89 & 3.58 \\
        \bottomrule
    \end{tabular}
    }
\end{table}

\paragraph{Summary.}
Overall, the NRVE-Acc results show that VM-Edit offers a strong balance between semantic editing quality and evaluation robustness. 
It achieves the best physical plausibility score, competitive temporal consistency and instruction-following scores, the second-best cross-judge mean, and the lowest cross-judge standard deviation. 
Together with the conventional metric results, these findings indicate that VM-Edit not only preserves visual structure and background content effectively, but also produces semantically plausible edits that are consistently preferred across diverse VLM judges. As shown in Table~\ref{tab:nrve_gemini25pro_per_edit}, Gemini-2.5-Pro further validates the physical plausibility of VM-Edit. 
Although our method is not designed to explicitly optimize NRVE-Acc-style physical metrics, VM-Edit achieves the best average $S_{\mathrm{phy}}$, with the best or tied-best physical plausibility on five out of six edit types. 
This result highlights the potential of VM-Edit in preserving physically consistent video dynamics. Detailed per-edit results under different VLM evaluators are provided in Tables~\ref{tab:nrve_gemini25pro_per_edit}--\ref{tab:nrve_kimi25_per_edit}.

\begin{table*}[!htbp]
    \centering
    \caption{Per-edit quantitative comparison on NRVBench using Gemini-2.5-Pro as the VLM evaluator. Scores are reported as percentages ($\times 100$). \textbf{Bold} and \uline{underline} indicate the best and second-best results in each column, respectively.}
    \label{tab:nrve_gemini25pro_per_edit}
    \renewcommand{\arraystretch}{1.22}
    \setlength{\tabcolsep}{4.5pt}
    \vspace{0.5em}
\begin{minipage}{0.49\textwidth}
    \centering
    \small
    \textbf{Edit 1}\\[2pt]
    \resizebox{\linewidth}{!}{
    \begin{tabular}{lcccc}
        \toprule
        Method & $S_{phy}\uparrow$ & $S_{temp}\uparrow$ & $S_{instr}\uparrow$ & \textbf{NRVE-Acc}$\uparrow$ \\
        \midrule
        TokenFlow & 52.00 & 28.00 & 46.67 & \uline{19.56} \\
        Pyramid-Edit & 22.67 & 20.67 & \uline{51.67} & 14.03 \\
        Wan-Edit & \uline{53.33} & 27.33 & \textbf{53.33} & \textbf{20.36} \\
        VidToMe & 44.00 & \uline{32.00} & 46.67 & 19.14 \\
        AnyV2V & 21.38 & 26.67 & 40.00 & 15.33 \\
        \textbf{Ours} & \textbf{73.33} & \textbf{32.67} & 43.33 & 18.55 \\
        \bottomrule
    \end{tabular}
    }
\end{minipage}
\vspace{\fill}
\begin{minipage}{0.49\textwidth}
    \centering
    \small
    \textbf{Edit 2}\\[2pt]
    \resizebox{\linewidth}{!}{
    \begin{tabular}{lcccc}
        \toprule
        Method & $S_{phy}\uparrow$ & $S_{temp}\uparrow$ & $S_{instr}\uparrow$ & \textbf{NRVE-Acc}$\uparrow$ \\
        \midrule
        TokenFlow & \uline{72.67} & \uline{36.00} & 45.00 & 26.36 \\
        Pyramid-Edit & 29.33 & 30.67 & 28.33 & 8.92 \\
        Wan-Edit & \textbf{85.33} & 32.67 & \uline{56.67} & \uline{31.09} \\
        VidToMe & 61.33 & 32.67 & 40.00 & 18.04 \\
        AnyV2V & 44.14 & 32.00 & 21.67 & 7.77 \\
        \textbf{Ours} & \textbf{85.33} & \textbf{36.67} & \textbf{58.33} & \textbf{31.46} \\
        \bottomrule
    \end{tabular}
    }
\end{minipage}

\begin{minipage}{0.49\textwidth}
    \centering
    \small
    \textbf{Edit 3}\\[2pt]
    \resizebox{\linewidth}{!}{
    \begin{tabular}{lcccc}
        \toprule
        Method & $S_{phy}\uparrow$ & $S_{temp}\uparrow$ & $S_{instr}\uparrow$ & \textbf{NRVE-Acc}$\uparrow$ \\
        \midrule
        TokenFlow & 58.00 & \uline{43.33} & 51.67 & 22.62 \\
        Pyramid-Edit & \textbf{70.67} & 26.00 & 46.67 & 17.64 \\
        Wan-Edit & 62.67 & \textbf{44.67} & \uline{55.00} & \textbf{24.82} \\
        VidToMe & 52.00 & 25.33 & 50.00 & 20.39 \\
        AnyV2V & 52.00 & 28.67 & 40.00 & 18.02 \\
        \textbf{Ours} & \uline{66.00} & 38.67 & \textbf{60.00} & \uline{23.16} \\
        \bottomrule
    \end{tabular}
    }
\end{minipage}
\hfill
\begin{minipage}{0.49\textwidth}
    \centering
    \small
    \textbf{Edit 4}\\[2pt]
    \resizebox{\linewidth}{!}{
    \begin{tabular}{lcccc}
        \toprule
        Method & $S_{phy}\uparrow$ & $S_{temp}\uparrow$ & $S_{instr}\uparrow$ & \textbf{NRVE-Acc}$\uparrow$ \\
        \midrule
        TokenFlow & 56.67 & \uline{54.00} & 51.67 & 26.58 \\
        Pyramid-Edit & \uline{66.00} & 36.00 & \uline{63.33} & 24.74 \\
        Wan-Edit & 60.00 & \textbf{60.67} & 61.67 & \uline{30.54} \\
        VidToMe & 58.67 & 46.00 & \textbf{70.00} & 22.98 \\
        AnyV2V & 38.67 & \uline{54.00} & 55.00 & 20.97 \\
        \textbf{Ours} & \textbf{68.00} & 49.33 & 58.33 & \textbf{32.08} \\
        \bottomrule
    \end{tabular}
    }
\end{minipage}
\vspace{\fill}

\begin{minipage}{0.49\textwidth}
    \centering
    \small
    \textbf{Edit 5}\\[2pt]
    \resizebox{\linewidth}{!}{
    \begin{tabular}{lcccc}
        \toprule
        Method & $S_{phy}\uparrow$ & $S_{temp}\uparrow$ & $S_{instr}\uparrow$ & \textbf{NRVE-Acc}$\uparrow$ \\
        \midrule
        TokenFlow & 49.33 & \uline{37.33} & 48.33 & 20.21 \\
        Pyramid-Edit & 32.67 & 28.67 & \textbf{71.67} & 17.97 \\
        Wan-Edit & \uline{61.33} & 36.00 & \uline{65.00} & \textbf{26.48} \\
        VidToMe & 30.67 & 28.00 & 55.00 & 18.73 \\
        AnyV2V & 32.67 & 30.00 & 45.00 & 12.24 \\
        \textbf{Ours} & \textbf{62.67} & \textbf{38.67} & 60.00 & \uline{23.34} \\
        \bottomrule
    \end{tabular}
    }
\end{minipage}
\hfill
\begin{minipage}{0.49\textwidth}
    \centering
    \small
    \textbf{Edit 6}\\[2pt]
    \resizebox{\linewidth}{!}{
    \begin{tabular}{lcccc}
        \toprule
        Method & $S_{phy}\uparrow$ & $S_{temp}\uparrow$ & $S_{instr}\uparrow$ & \textbf{NRVE-Acc}$\uparrow$ \\
        \midrule
        TokenFlow & 56.67 & \textbf{31.33} & 61.67 & 22.36 \\
        Pyramid-Edit & 47.33 & 22.00 & \textbf{68.33} & 22.66 \\
        Wan-Edit & \uline{63.33} & \uline{30.67} & \uline{66.67} & \textbf{25.90} \\
        VidToMe & 60.67 & 27.33 & 61.67 & 24.68 \\
        AnyV2V & 42.67 & 22.67 & 56.67 & 17.66 \\
        \textbf{Ours} & \textbf{73.33} & 24.00 & 60.00 & \uline{24.71} \\
        \bottomrule
    \end{tabular}
    }
\end{minipage}
\vspace{\fill}
\end{table*}

\clearpage

\begin{table*}[!htbp]
    \centering
    \caption{Per-edit quantitative comparison on NRVBench using Qwen2.5-VL-72B as the VLM evaluator. Scores are reported as percentages ($\times 100$). \textbf{Bold} and \uline{underline} indicate the best and second-best results in each column, respectively.}
    \label{tab:nrve_qwen72b_per_edit}
    \renewcommand{\arraystretch}{1.22}
\begin{minipage}{0.49\textwidth}
    \centering
    \small
    \textbf{Edit 1}\\[2pt]
    \resizebox{\linewidth}{!}{
    \begin{tabular}{lcccc}
        \toprule
        Method & $S_{phy}\uparrow$ & $S_{temp}\uparrow$ & $S_{instr}\uparrow$ & \textbf{NRVE-Acc}$\uparrow$ \\
        \midrule
        TokenFlow & 84.67 & 36.67 & \textbf{55.00} & \textbf{30.87} \\
        Pyramid-Edit & 44.00 & \textbf{39.33} & \textbf{55.00} & 22.19 \\
        Wan-Edit & 78.00 & \textbf{39.33} & 51.67 & 28.26 \\
        VidToMe & \uline{85.33} & \uline{38.67} & 51.67 & \uline{30.68} \\
        AnyV2V & 49.33 & 34.67 & \uline{53.33} & 23.73 \\
        \textbf{Ours} & \textbf{89.33} & 37.33 & 51.67 & 29.03 \\
        \bottomrule
    \end{tabular}
    }
\end{minipage}
\hfill
\begin{minipage}{0.49\textwidth}
    \centering
    \small
    \textbf{Edit 2}\\[2pt]
    \resizebox{\linewidth}{!}{
    \begin{tabular}{lcccc}
        \toprule
        Method & $S_{phy}\uparrow$ & $S_{temp}\uparrow$ & $S_{instr}\uparrow$ & \textbf{NRVE-Acc}$\uparrow$ \\
        \midrule
        TokenFlow & \uline{79.33} & \textbf{40.00} & 50.00 & 31.61 \\
        Pyramid-Edit & 66.00 & \textbf{40.00} & \textbf{60.00} & 30.53 \\
        Wan-Edit & \textbf{80.00} & \textbf{40.00} & \uline{58.33} & \textbf{33.09} \\
        VidToMe & \uline{79.33} & \uline{39.33} & 53.33 & 31.97 \\
        AnyV2V & 70.67 & \textbf{40.00} & 43.33 & 23.95 \\
        \textbf{Ours} & \uline{79.33} & \textbf{40.00} & 56.67 & \uline{32.75} \\
        \bottomrule
    \end{tabular}
    }
\end{minipage}

\begin{minipage}{0.49\textwidth}
    \centering
    \small
    \textbf{Edit 3}\\[2pt]
    \resizebox{\linewidth}{!}{
    \begin{tabular}{lcccc}
        \toprule
        Method & $S_{phy}\uparrow$ & $S_{temp}\uparrow$ & $S_{instr}\uparrow$ & \textbf{NRVE-Acc}$\uparrow$ \\
        \midrule
        TokenFlow & 69.33 & \uline{38.00} & 55.00 & 27.67 \\
        Pyramid-Edit & 66.00 & 37.33 & \textbf{70.00} & \uline{31.13} \\
        Wan-Edit & \textbf{74.67} & \textbf{38.67} & \uline{60.00} & \textbf{31.79} \\
        VidToMe & 70.67 & \textbf{38.67} & 58.33 & 29.25 \\
        AnyV2V & 68.67 & 37.33 & 55.00 & 23.40 \\
        \textbf{Ours} & \uline{72.00} & 37.33 & 58.33 & 30.49 \\
        \bottomrule
    \end{tabular}
    }
\end{minipage}
\hfill
\begin{minipage}{0.49\textwidth}
    \centering
    \small
    \textbf{Edit 4}\\[2pt]
    \resizebox{\linewidth}{!}{
    \begin{tabular}{lcccc}
        \toprule
        Method & $S_{phy}\uparrow$ & $S_{temp}\uparrow$ & $S_{instr}\uparrow$ & \textbf{NRVE-Acc}$\uparrow$ \\
        \midrule
        TokenFlow & 74.67 & 36.67 & 63.33 & 30.05 \\
        Pyramid-Edit & \uline{75.33} & 36.67 & \uline{70.00} & \uline{32.38} \\
        Wan-Edit & 72.00 & 38.67 & 65.00 & 31.71 \\
        VidToMe & \textbf{76.67} & \uline{39.33} & \textbf{71.67} & \textbf{33.96} \\
        AnyV2V & 71.33 & 38.00 & \uline{70.00} & 30.74 \\
        \textbf{Ours} & 70.97 & \textbf{39.35} & 61.29 & 31.21 \\
        \bottomrule
    \end{tabular}
    }
\end{minipage}

\begin{minipage}{0.49\textwidth}
    \centering
    \small
    \textbf{Edit 5}\\[2pt]
    \resizebox{\linewidth}{!}{
    \begin{tabular}{lcccc}
        \toprule
        Method & $S_{phy}\uparrow$ & $S_{temp}\uparrow$ & $S_{instr}\uparrow$ & \textbf{NRVE-Acc}$\uparrow$ \\
        \midrule
        TokenFlow & 76.67 & 37.33 & 61.67 & 30.17 \\
        Pyramid-Edit & 73.33 & \textbf{40.00} & \textbf{78.33} & 33.12 \\
        Wan-Edit & \uline{79.09} & \textbf{40.00} & 63.16 & \textbf{35.46} \\
        VidToMe & \textbf{80.00} & \uline{38.67} & 68.33 & \uline{33.44} \\
        AnyV2V & 68.67 & 34.00 & \uline{71.67} & 28.03 \\
        \textbf{Ours} & 75.33 & 38.00 & 60.00 & 30.40 \\
        \bottomrule
    \end{tabular}
    }
\end{minipage}
\hfill
\begin{minipage}{0.49\textwidth}
    \centering
    \small
    \textbf{Edit 6}\\[2pt]
    \resizebox{\linewidth}{!}{
    \begin{tabular}{lcccc}
        \toprule
        Method & $S_{phy}\uparrow$ & $S_{temp}\uparrow$ & $S_{instr}\uparrow$ & \textbf{NRVE-Acc}$\uparrow$ \\
        \midrule
        TokenFlow & 70.67 & \uline{38.67} & 56.67 & 28.74 \\
        Pyramid-Edit & 54.29 & 37.33 & 55.00 & 15.06 \\
        Wan-Edit & \textbf{75.65} & \textbf{39.09} & \textbf{61.76} & \textbf{30.58} \\
        VidToMe & \uline{72.00} & \uline{38.67} & \uline{58.33} & 29.18 \\
        AnyV2V & 64.67 & 37.33 & \uline{58.33} & 26.12 \\
        \textbf{Ours} & 71.72 & 37.93 & 56.90 & \uline{29.31} \\
        \bottomrule
    \end{tabular}
    }
\end{minipage}
\end{table*}

\begin{table*}[!htbp]
    \centering
    \caption{Per-edit quantitative comparison on NRVBench using Qwen2.5-VL-7B as the VLM evaluator. Scores are reported as percentages ($\times 100$). \textbf{Bold} and \uline{underline} indicate the best and second-best results in each column, respectively.}
    \label{tab:nrve_qwen7b_per_edit}
    \renewcommand{\arraystretch}{1.22}
\begin{minipage}{0.49\textwidth}
    \centering
    \small
    \textbf{Edit 1}\\[2pt]
    \resizebox{\linewidth}{!}{
    \begin{tabular}{lcccc}
        \toprule
        Method & $S_{phy}\uparrow$ & $S_{temp}\uparrow$ & $S_{instr}\uparrow$ & \textbf{NRVE-Acc}$\uparrow$ \\
        \midrule
        TokenFlow & 70.00 & 43.33 & \uline{58.33} & 31.88 \\
        Pyramid-Edit & 56.00 & 36.67 & \textbf{63.33} & 25.99 \\
        Wan-Edit & 72.67 & \textbf{55.33} & \uline{58.33} & \textbf{33.87} \\
        VidToMe & \textbf{84.00} & 34.00 & \uline{58.33} & 29.09 \\
        AnyV2V & 68.67 & 40.00 & \uline{58.33} & 28.04 \\
        \textbf{Ours} & \uline{75.33} & \uline{45.33} & \uline{58.33} & \uline{32.33} \\
        \bottomrule
    \end{tabular}
    }
\end{minipage}
\hfill
\begin{minipage}{0.49\textwidth}
    \centering
    \small
    \textbf{Edit 2}\\[2pt]
    \resizebox{\linewidth}{!}{
    \begin{tabular}{lcccc}
        \toprule
        Method & $S_{phy}\uparrow$ & $S_{temp}\uparrow$ & $S_{instr}\uparrow$ & \textbf{NRVE-Acc}$\uparrow$ \\
        \midrule
        TokenFlow & \uline{76.67} & 46.67 & 60.00 & 30.82 \\
        Pyramid-Edit & 65.33 & \uline{52.67} & 66.67 & 31.88 \\
        Wan-Edit & 75.33 & \textbf{57.33} & \uline{71.67} & \uline{38.88} \\
        VidToMe & \textbf{92.67} & 38.00 & 63.33 & 33.47 \\
        AnyV2V & \uline{76.67} & 42.00 & 46.67 & 20.59 \\
        \textbf{Ours} & \uline{76.67} & \textbf{57.33} & \textbf{73.33} & \textbf{39.49} \\
        \bottomrule
    \end{tabular}
    }
\end{minipage}

\begin{minipage}{0.49\textwidth}
    \centering
    \small
    \textbf{Edit 3}\\[2pt]
    \resizebox{\linewidth}{!}{
    \begin{tabular}{lcccc}
        \toprule
        Method & $S_{phy}\uparrow$ & $S_{temp}\uparrow$ & $S_{instr}\uparrow$ & \textbf{NRVE-Acc}$\uparrow$ \\
        \midrule
        TokenFlow & 66.00 & \uline{52.67} & 75.00 & \uline{35.00} \\
        Pyramid-Edit & 65.33 & 42.00 & \textbf{85.00} & 34.99 \\
        Wan-Edit & \uline{67.33} & \textbf{54.67} & 71.67 & \textbf{36.68} \\
        VidToMe & \textbf{74.55} & 36.36 & 59.09 & 28.16 \\
        AnyV2V & \uline{67.33} & 47.33 & \uline{80.00} & 34.05 \\
        \textbf{Ours} & 62.67 & 49.33 & 71.67 & 33.97 \\
        \bottomrule
    \end{tabular}
    }
\end{minipage}
\hfill
\begin{minipage}{0.49\textwidth}
    \centering
    \small
    \textbf{Edit 4}\\[2pt]
    \resizebox{\linewidth}{!}{
    \begin{tabular}{lcccc}
        \toprule
        Method & $S_{phy}\uparrow$ & $S_{temp}\uparrow$ & $S_{instr}\uparrow$ & \textbf{NRVE-Acc}$\uparrow$ \\
        \midrule
        TokenFlow & \textbf{73.33} & 48.67 & 76.67 & 35.69 \\
        Pyramid-Edit & \uline{72.67} & 47.33 & \uline{85.00} & 37.10 \\
        Wan-Edit & 70.67 & \textbf{56.00} & 78.33 & \textbf{38.42} \\
        VidToMe & / & / & / & / \\
        AnyV2V & \uline{72.67} & 46.67 & \textbf{88.33} & \uline{37.39} \\
        \textbf{Ours} & 69.33 & \uline{54.00} & 78.33 & 37.36 \\
        \bottomrule
    \end{tabular}
    }
\end{minipage}

\begin{minipage}{0.49\textwidth}
    \centering
    \small
    \textbf{Edit 5}\\[2pt]
    \resizebox{\linewidth}{!}{
    \begin{tabular}{lcccc}
        \toprule
        Method & $S_{phy}\uparrow$ & $S_{temp}\uparrow$ & $S_{instr}\uparrow$ & \textbf{NRVE-Acc}$\uparrow$ \\
        \midrule
        TokenFlow & \uline{75.33} & 45.33 & \uline{73.33} & 35.19 \\
        Pyramid-Edit & 70.00 & 48.67 & \textbf{83.33} & \uline{35.23} \\
        Wan-Edit & \uline{75.33} & \uline{51.33} & 71.67 & \textbf{35.66} \\
        VidToMe & / & / & / & / \\
        AnyV2V & \textbf{76.00} & 42.67 & 71.67 & 30.30 \\
        \textbf{Ours} & \uline{75.33} & \textbf{53.33} & 65.00 & 34.28 \\
        \bottomrule
    \end{tabular}
    }
\end{minipage}
\hfill
\begin{minipage}{0.49\textwidth}
    \centering
    \small
    \textbf{Edit 6}\\[2pt]
    \resizebox{\linewidth}{!}{
    \begin{tabular}{lcccc}
        \toprule
        Method & $S_{phy}\uparrow$ & $S_{temp}\uparrow$ & $S_{instr}\uparrow$ & \textbf{NRVE-Acc}$\uparrow$ \\
        \midrule
        TokenFlow & \uline{76.67} & 38.00 & 61.67 & 29.71 \\
        Pyramid-Edit & 60.67 & \uline{40.00} & \uline{68.33} & 28.04 \\
        Wan-Edit & \textbf{78.00} & \textbf{52.67} & \textbf{70.00} & \textbf{37.78} \\
        VidToMe & / & / & / & / \\
        AnyV2V & \uline{76.67} & \uline{40.00} & \textbf{70.00} & \uline{31.74} \\
        \textbf{Ours} & 69.33 & 37.33 & 66.67 & 30.84 \\
        \bottomrule
    \end{tabular}
    }
\end{minipage}
\end{table*}

\begin{table*}[!htbp]
    \centering
    \caption{Per-edit quantitative comparison on NRVBench using Qwen2.5-VL-32B as the VLM evaluator. Scores are reported as percentages ($\times 100$). \textbf{Bold} and \uline{underline} indicate the best and second-best results in each column, respectively.}
    \label{tab:nrve_qwen32b_per_edit}
    \renewcommand{\arraystretch}{1.22}
\begin{minipage}{0.49\textwidth}
    \centering
    \small
    \textbf{Edit 1}\\[2pt]
    \resizebox{\linewidth}{!}{
    \begin{tabular}{lcccc}
        \toprule
        Method & $S_{phy}\uparrow$ & $S_{temp}\uparrow$ & $S_{instr}\uparrow$ & \textbf{NRVE-Acc}$\uparrow$ \\
        \midrule
        TokenFlow & \uline{78.00} & 43.33 & 58.33 & \uline{32.98} \\
        Pyramid-Edit & 62.00 & \textbf{62.67} & \textbf{70.00} & \textbf{35.06} \\
        Wan-Edit & \textbf{80.00} & 43.33 & 53.33 & 30.56 \\
        VidToMe & 76.67 & 49.33 & 55.00 & 32.92 \\
        AnyV2V & 61.33 & 44.00 & \uline{63.33} & 28.60 \\
        \textbf{Ours} & \textbf{80.00} & \uline{54.67} & 53.33 & 31.41 \\
        \bottomrule
    \end{tabular}
    }
\end{minipage}
\hfill
\begin{minipage}{0.49\textwidth}
    \centering
    \small
    \textbf{Edit 2}\\[2pt]
    \resizebox{\linewidth}{!}{
    \begin{tabular}{lcccc}
        \toprule
        Method & $S_{phy}\uparrow$ & $S_{temp}\uparrow$ & $S_{instr}\uparrow$ & \textbf{NRVE-Acc}$\uparrow$ \\
        \midrule
        TokenFlow & \textbf{80.00} & 41.33 & 66.67 & 32.74 \\
        Pyramid-Edit & \textbf{80.00} & \textbf{66.67} & \uline{70.00} & \textbf{40.89} \\
        Wan-Edit & \textbf{80.00} & 42.00 & 68.33 & 33.99 \\
        VidToMe & \textbf{80.00} & \uline{48.67} & 60.00 & 33.58 \\
        AnyV2V & \textbf{80.00} & 42.67 & \textbf{73.33} & 34.08 \\
        \textbf{Ours} & \textbf{80.00} & 48.00 & \uline{70.00} & \uline{36.36} \\
        \bottomrule
    \end{tabular}
    }
\end{minipage}

\begin{minipage}{0.49\textwidth}
    \centering
    \small
    \textbf{Edit 3}\\[2pt]
    \resizebox{\linewidth}{!}{
    \begin{tabular}{lcccc}
        \toprule
        Method & $S_{phy}\uparrow$ & $S_{temp}\uparrow$ & $S_{instr}\uparrow$ & \textbf{NRVE-Acc}$\uparrow$ \\
        \midrule
        TokenFlow & \textbf{75.33} & \uline{53.33} & 68.33 & \uline{36.44} \\
        Pyramid-Edit & 67.33 & \textbf{60.00} & \textbf{78.33} & \textbf{38.73} \\
        Wan-Edit & 72.67 & 52.67 & 66.67 & 35.48 \\
        VidToMe & \uline{74.00} & 48.00 & \uline{71.67} & 35.81 \\
        AnyV2V & 73.33 & 46.00 & \textbf{78.33} & 34.19 \\
        \textbf{Ours} & 73.33 & 49.33 & 70.00 & 35.65 \\
        \bottomrule
    \end{tabular}
    }
\end{minipage}
\hfill
\begin{minipage}{0.49\textwidth}
    \centering
    \small
    \textbf{Edit 4}\\[2pt]
    \resizebox{\linewidth}{!}{
    \begin{tabular}{lcccc}
        \toprule
        Method & $S_{phy}\uparrow$ & $S_{temp}\uparrow$ & $S_{instr}\uparrow$ & \textbf{NRVE-Acc}$\uparrow$ \\
        \midrule
        TokenFlow & 77.33 & 48.00 & 75.00 & 36.81 \\
        Pyramid-Edit & \uline{78.00} & 47.33 & 80.00 & 37.99 \\
        Wan-Edit & 76.67 & \uline{52.00} & 73.33 & 37.82 \\
        VidToMe & \textbf{80.00} & 49.33 & \uline{81.67} & \textbf{39.20} \\
        AnyV2V & 76.67 & 46.67 & \textbf{83.33} & \uline{38.18} \\
        \textbf{Ours} & 75.33 & \textbf{54.67} & 71.67 & 37.90 \\
        \bottomrule
    \end{tabular}
    }
\end{minipage}

\begin{minipage}{0.49\textwidth}
    \centering
    \small
    \textbf{Edit 5}\\[2pt]
    \resizebox{\linewidth}{!}{
    \begin{tabular}{lcccc}
        \toprule
        Method & $S_{phy}\uparrow$ & $S_{temp}\uparrow$ & $S_{instr}\uparrow$ & \textbf{NRVE-Acc}$\uparrow$ \\
        \midrule
        TokenFlow & \uline{78.67} & 53.33 & 70.00 & 35.69 \\
        Pyramid-Edit & 77.33 & \textbf{66.67} & \textbf{86.67} & \textbf{43.06} \\
        Wan-Edit & \textbf{80.00} & \uline{60.00} & 73.33 & \uline{39.66} \\
        VidToMe & 77.33 & 51.33 & 65.00 & 32.43 \\
        AnyV2V & 74.00 & 46.67 & \uline{78.33} & 34.46 \\
        \textbf{Ours} & \textbf{80.00} & 54.00 & 70.00 & 36.92 \\
        \bottomrule
    \end{tabular}
    }
\end{minipage}
\hfill
\begin{minipage}{0.49\textwidth}
    \centering
    \small
    \textbf{Edit 6}\\[2pt]
    \resizebox{\linewidth}{!}{
    \begin{tabular}{lcccc}
        \toprule
        Method & $S_{phy}\uparrow$ & $S_{temp}\uparrow$ & $S_{instr}\uparrow$ & \textbf{NRVE-Acc}$\uparrow$ \\
        \midrule
        TokenFlow & 75.33 & 45.33 & \textbf{70.00} & \textbf{33.93} \\
        Pyramid-Edit & 69.33 & \textbf{58.00} & \uline{66.67} & 32.35 \\
        Wan-Edit & 76.00 & \uline{57.33} & 56.67 & 32.14 \\
        VidToMe & \textbf{78.67} & 47.33 & 60.00 & 30.24 \\
        AnyV2V & \uline{78.00} & 47.33 & \uline{66.67} & \uline{33.76} \\
        \textbf{Ours} & 70.00 & 48.67 & 61.67 & 31.19 \\
        \bottomrule
    \end{tabular}
    }
\end{minipage}
\end{table*}

\begin{table*}[!htbp]
    \centering
    \caption{Per-edit quantitative comparison on NRVBench using Kimi-2.5 as the VLM evaluator. Scores are reported as percentages ($\times 100$). \textbf{Bold} and \uline{underline} indicate the best and second-best results in each column, respectively.}
    \label{tab:nrve_kimi25_per_edit}
    \renewcommand{\arraystretch}{1.22}
\begin{minipage}{0.49\textwidth}
    \centering
    \footnotesize
    \textbf{Edit 1}\\[2pt]
    \resizebox{\linewidth}{!}{
    \begin{tabular}{lcccc}
        \toprule
        Method & $S_{phy}\uparrow$ & $S_{temp}\uparrow$ & $S_{instr}\uparrow$ & \textbf{NRVE-Acc}$\uparrow$ \\
        \midrule
        TokenFlow & 40.00 & 38.67 & \uline{60.71} & 21.96 \\
        Pyramid-Edit & 27.33 & 35.33 & \uline{60.71} & 21.75 \\
        Wan-Edit & 47.33 & 38.67 & 51.79 & 24.47 \\
        VidToMe & \uline{48.67} & \textbf{42.00} & \textbf{62.07} & \uline{25.76} \\
        AnyV2V & 30.56 & 35.00 & 58.93 & 19.79 \\
        \textbf{Ours} & \textbf{62.00} & \uline{39.33} & 57.41 & \textbf{27.75} \\
        \bottomrule
    \end{tabular}
    }
\end{minipage}
\hfill
\begin{minipage}{0.49\textwidth}
    \centering
    \small
    \textbf{Edit 2}\\[2pt]
    \resizebox{\linewidth}{!}{
    \begin{tabular}{lcccc}
        \toprule
        Method & $S_{phy}\uparrow$ & $S_{temp}\uparrow$ & $S_{instr}\uparrow$ & \textbf{NRVE-Acc}$\uparrow$ \\
        \midrule
        TokenFlow & \textbf{72.67} & 37.33 & \textbf{66.67} & \textbf{31.46} \\
        Pyramid-Edit & 48.00 & \uline{38.67} & 48.15 & 18.42 \\
        Wan-Edit & \uline{64.67} & \uline{38.67} & 58.62 & 27.02 \\
        VidToMe & 59.33 & \textbf{39.33} & \uline{64.81} & \uline{30.69} \\
        AnyV2V & 44.31 & 36.86 & 41.11 & 16.61 \\
        \textbf{Ours} & 58.67 & 32.00 & 59.26 & 27.54 \\
        \bottomrule
    \end{tabular}
    }
\end{minipage}

\begin{minipage}{0.49\textwidth}
    \centering
    \small
    \textbf{Edit 3}\\[2pt]
    \resizebox{\linewidth}{!}{
    \begin{tabular}{lcccc}
        \toprule
        Method & $S_{phy}\uparrow$ & $S_{temp}\uparrow$ & $S_{instr}\uparrow$ & \textbf{NRVE-Acc}$\uparrow$ \\
        \midrule
        TokenFlow & 57.33 & \uline{39.33} & \uline{70.00} & \textbf{30.22} \\
        Pyramid-Edit & \textbf{62.67} & 36.67 & 63.79 & 23.35 \\
        Wan-Edit & 44.67 & 38.00 & 63.79 & 28.17 \\
        VidToMe & 56.00 & 37.33 & 61.54 & 25.89 \\
        AnyV2V & 52.67 & \textbf{40.00} & \textbf{72.41} & 27.73 \\
        \textbf{Ours} & \uline{62.00} & 37.33 & 63.79 & \uline{29.28} \\
        \bottomrule
    \end{tabular}
    }
\end{minipage}
\hfill
\begin{minipage}{0.49\textwidth}
    \centering
    \small
    \textbf{Edit 4}\\[2pt]
    \resizebox{\linewidth}{!}{
    \begin{tabular}{lcccc}
        \toprule
        Method & $S_{phy}\uparrow$ & $S_{temp}\uparrow$ & $S_{instr}\uparrow$ & \textbf{NRVE-Acc}$\uparrow$ \\
        \midrule
        TokenFlow & 62.67 & \uline{43.33} & 65.00 & \textbf{35.14} \\
        Pyramid-Edit & \uline{68.00} & 35.33 & 71.67 & 29.79 \\
        Wan-Edit & \textbf{71.33} & 39.33 & \uline{74.14} & 29.36 \\
        VidToMe & 61.33 & \textbf{44.67} & \textbf{77.59} & 29.13 \\
        AnyV2V & 50.67 & \textbf{44.67} & \textbf{77.59} & 30.49 \\
        \textbf{Ours} & 64.00 & 42.67 & 68.97 & \uline{30.95} \\
        \bottomrule
    \end{tabular}
    }
\end{minipage}

\begin{minipage}{0.49\textwidth}
    \centering
    \small
    \textbf{Edit 5}\\[2pt]
    \resizebox{\linewidth}{!}{
    \begin{tabular}{lcccc}
        \toprule
        Method & $S_{phy}\uparrow$ & $S_{temp}\uparrow$ & $S_{instr}\uparrow$ & \textbf{NRVE-Acc}$\uparrow$ \\
        \midrule
        TokenFlow & 57.33 & \textbf{39.33} & \uline{75.86} & 26.05 \\
        Pyramid-Edit & 50.67 & 35.33 & 68.33 & 26.70 \\
        Wan-Edit & 62.00 & \uline{38.67} & \textbf{76.79} & \textbf{28.58} \\
        VidToMe & \textbf{65.33} & 34.67 & 67.86 & \uline{28.28} \\
        AnyV2V & 39.33 & \uline{38.67} & 72.41 & 25.36 \\
        \textbf{Ours} & \uline{64.67} & 36.00 & 68.97 & 28.24 \\
        \bottomrule
    \end{tabular}
    }
\end{minipage}
\hfill
\begin{minipage}{0.49\textwidth}
    \centering
    \small
    \textbf{Edit 6}\\[2pt]
    \resizebox{\linewidth}{!}{
    \begin{tabular}{lcccc}
        \toprule
        Method & $S_{phy}\uparrow$ & $S_{temp}\uparrow$ & $S_{instr}\uparrow$ & \textbf{NRVE-Acc}$\uparrow$ \\
        \midrule
        TokenFlow & 51.33 & \uline{36.00} & \textbf{71.67} & \uline{28.43} \\
        Pyramid-Edit & 53.33 & \uline{36.00} & \uline{71.43} & 26.48 \\
        Wan-Edit & \uline{56.00} & 34.67 & 70.69 & 26.82 \\
        VidToMe & \textbf{58.00} & \uline{36.00} & 70.00 & 25.62 \\
        AnyV2V & 53.33 & \uline{36.00} & 70.69 & 26.02 \\
        \textbf{Ours} & 53.33 & \textbf{40.67} & 63.79 & \textbf{28.78} \\
        \bottomrule
    \end{tabular}
    }
\end{minipage}
\end{table*}

\end{document}